\documentclass[logo, address]{trfr}
\usepackage[numbers,sort]{natbib}

\usepackage[utf8]{inputenc}
\usepackage{url}
\usepackage{float}
\usepackage{amsfonts}
\usepackage{amsmath}
\usepackage{amssymb}
\usepackage{nicefrac}
\usepackage[table]{xcolor}
\usepackage{amsthm}

\usepackage{comment}
\usepackage{xurl}
\newtcolorbox{promptbox}[1]{
  enhanced,
  breakable,
  colback=gray!5,
  colframe=black!60,
  fonttitle=\small\bfseries,
  coltitle=black,
  colbacktitle=gray!20,
  title=#1,
  sharp corners,
  boxrule=0.4pt,
  left=6pt, right=6pt, top=4pt, bottom=4pt,
  before skip=8pt, after skip=8pt,
}

\newcommand{\cellbase}[1]{\cellcolor[RGB]{220,220,220}\textbf{#1}}

\title{Test-Time Scaling in the Wild: Why Exploitation, Not Exploration,
       Is the Bottleneck}
\shorttitle{Test-Time Scaling in the Wild}

\abstract{
Test-time scaling (TTS) improves language model outputs by spending additional inference compute — generating multiple candidates, searching over partial sequences, or iteratively refining drafts. These techniques yield large gains on mathematics and code, but have been developed and stress-tested almost exclusively on tasks where verification is straightforward. We conduct the first compute-normalised comparison of five TTS families across five open-ended generation benchmarks spanning medicine, law, finance, general chat, and creative writing — grounded in a unified framework that decomposes the effectiveness of each method's token budget into exploration and exploitation. The answer depends on which side of that decomposition you examine. Scaling exploration works: the best candidate in the pool improves steadily with compute across all settings. What breaks is exploitation — the step that converts a rich candidate pool into a final output. With state-of-the-art generators,
reward models correlate at only $\hat{\rho}_v \approx 0.12$ with true quality, rendering selection near-random regardless of budget. Tree search amplifies this failure through diversity collapse. Refinement helps on one of five benchmarks; its apparent gains elsewhere are confounded. Only synthesis across candidates (Fusion) consistently improves over
single-sample baselines, yet still recovers only ${\sim}40\%$ of
available quality.
The candidate pool is not the bottleneck — choosing from it is.
}

\author[1,*]{Davide Romano}
\author[1,*]{Kanak Raj}
\author[1]{Jerrod Parker}
\author[1]{Daniele Giofr\`e}

\affiliation[1]{Thomson Reuters}

\contribution[*]{Equal contribution}

\correspondence{
\{davide.romano2,kanak.raj,jerrod.parker,daniele.giofre\}@thomsonreuters.com}

\begin{document}

\maketitle


\section{Introduction}
\label{sec:introduction}

Test-time scaling (TTS) improves language model performance without
updating weights, by investing additional compute at inference time.
Methods such as Best-of-$N$ sampling with reward models~\cite{snell2024scaling, beeching2024scalingtesttimecompute}, tree search~\cite{wu2024inference, puri2025rollout}, sequential refinement~\cite{madaan2023self} and extended thinking ~\cite{muennighoff2025s1} achieve large accuracy gains on mathematical reasoning, code generation, and multiple-choice tasks~\cite{skywork2024reward}.

TTS methods are split by how they exploit additional compute: some select among candidates with an external verifier---typically a reward model or process supervisor---while others critique or synthesise without one. On the benchmarks where TTS was developed---competition mathematics, program synthesis, multiple-choice QA ~\cite{Li2025STT,zhu2026codescalerscalingcodellm,song2025cttscollectivetesttimescaling}---verifier-based methods are high-performing because the verifier itself works: outcomes are binary, labelled supervision is abundant, and reward models trained on it rank candidates reliably. Open-ended generation breaks this premise. Quality on medical consultations, legal analyses, and creative writing is graded against nuanced rubrics rather than a single correct answer~\cite{shi2026plawbenchrubricbasedbenchmarkevaluating,
  elangovan2025development, sharma2025researchrubricsbenchmarkpromptsrubrics}, and supervision at comparable scale and reliability does not exist. The best off-the-shelf Reward Models (RMs) collapse to near random on open-ended tasks. Algorithms that depend on the verifier inherit this failure; methods that bypass it---synthesis across candidates or self-critique by the generator---fare unevenly across benchmarks and model families.

 We provide the first systematic evaluation of TTS on open-ended generation, benchmarking five TTS families---Best-of-$N$, Beam Search, Particle Filtering, Refinement, and Fusion---across five open-ended generation benchmarks at 
  matched compute. 


The central finding is that \textbf{exploitation---not exploration---is the bottleneck}. Oracle quality---the true quality of the best candidate in the pool, corrected for judge-noise inflation---rises with compute, just as on deterministic tasks. This surface similarity conceals a critical difference: realised quality stagnates or regresses for verifier-based methods, and even the strongest method Fusion captures only ${\sim}40\%$ of available  headroom---the gap between a single-sample baseline and the oracle. None of the three scaling axes we test---compute, reward model size, or generator size---closes this gap. The candidate pool contains high-quality answers; converting them into a strong final output is the open problem.

Beyond this central finding, we contribute:
\begin{enumerate}
    

    \item \textbf{Bias-corrected oracle estimator for TTS evaluation.}
    The naive oracle---the maximum judge score over a candidate pool---systematically overstates true pool quality, because taking the maximum of noisy scores selects partly on noise. This inflation grows with pool size and judge noise. We derive a closed-form per-entry bias-corrected estimator and estimate per-benchmark judge variance empirically from repeated scoring; the
  resulting gap between naive and corrected oracle is material on every benchmark.

    \item \textbf{Verifier correlation predicts and explains selection
    failure.} 
    We show analytically that for BoN, headroom capture approximately equals the verifier's correlation with ground truth (Eq.~\ref{eq:headroom}). Empirically, the Spearman correlation $\hat{\rho}_v$ between RM scores and true quality collapses to ${\sim}0.12$
  on open-ended generation; both tested Outcome Reward Models fail similarly, indicating a structural rather than model-specific failure. 

    \item \textbf{Per-method diagnosis of exploitation failure across all TTS families.} We provide the first head-to-head comparison of generative exploitation strategies (Fusion, Sequential Refinement) against selection-based methods at matched compute on open-ended tasks. The comparison reveals that no exploitation strategy scales reliably: Beam Search and Particle Filtering perform strictly worse than parallel Best-of-N due to RM-guided diversity collapse (40--60\% of independent-sampling diversity), and Sequential Refinement (SR) yields genuine improvement on only one of five benchmarks. Fusion is the sole method that improves over the single-sample baseline on every benchmark for Qwen3.5---yet yields mixed results on OLMo3, indicating that synthesis quality is a capability not all model families possess.

\end{enumerate}

\section{Related Work}
\label{sec:related}

\paragraph{Test-time scaling methods and verifiers.}
TTS methods fall approximately into five families: parallel, search,
refinement based, fusion based, and extended thinking.
\textit{Parallel methods} generate $N$ independent candidates and select via a
verifier. \textit{Search algorithms} structure exploration over
partial sequences via a Process Reward Model (PRM), from deterministic beam
search~\cite{beeching2024scalingtesttimecompute,wu2024inference} to stochastic
variants~\cite{puri2025rollout}. \textit{Sequential refinement} (or feedback)
iteratively critiques and rewrites a single
candidate~\cite{madaan2023self, madaan2025rethinking}. \textit{Fusion}
synthesises across multiple
candidates~\cite{khairi2025making}. Extended thinking
allocates the full budget to a single reasoning
trajectory~\cite{muennighoff2025s1}. Finally, many hybrid methods exist that combine strategies ~\cite{saadfalcon2025archonarchitecturesearchframework, inoue2025widerdeeperscalingllm,chang2025steplevelverifierguidedhybridtesttime} or improve weaknesses of the basic methods~\cite{snell2024scaling,dalal2026testtimecomputehurtoverestimation,liu20251bllmsurpass405b,zhang2025surveytesttimescalinglarge}.

The success of the first two families depends on the external verifier. ORMs evaluate a (query, completion) pair with a single score, and PRMs~\cite{lightman2023letsverifystepstep} are trained to score intermediate Chain of Thought (CoT) reasoning steps.
Existing ORMs and PRMs are trained predominantly on preference data or math-heavy datasets and one of the most common evaluation methods is Best-of-N sampling ~\cite{skywork2024reward, rewardbench2}.
Currently, there is no systematic comparison of these different families, and no framework to characterize how they trade off exploration (generating candidates) against exploitation (scoring, selecting, or synthesising a final output) under a shared token budget, or how that tradeoff degrades when verifier quality drops.

\paragraph{Evaluation beyond deterministic tasks.}
The evaluation of TTS methods is mostly performed on math-oriented tasks, while recently moving to multi-domain benchmarks, but still limited to deterministic and verifiable tasks~\cite{snell2024scaling,wu2024inference,zhang2025surveytesttimescalinglarge}.
Our work fills the remaining gap: the first systematic, compute-normalised comparison of the main TTS families on open-ended generation.

\section{Theoretical Framework}
\label{sec:theory}

We organise TTS methods along a single axis: how each method partitions a
fixed token budget between \emph{exploration} and
\emph{exploitation}  (Figure~\ref{fig:method_schema}).
This decomposition yields two implications that structure the analysis in
Section~\ref{sec:analysis}.

\subsection{Budget Decomposition and Quality Metrics}
\label{subsec:framework}

Let $\pi_\theta$ denote the generator, $\mathcal{V}$ a verifier assigning
a scalar quality score to a candidate $y$ given prompt $x$, and
$\mathcal{E}$ the exploitation step mapping a candidate pool $\mathcal{P}$
to a final response $\hat{y} \sim \mathcal{E}(\mathcal{P})$. We distinguish
three verifier types: the \textbf{ground-truth verifier} $\mathcal{V}^*$;
\textbf{external verifiers} $(\mathcal{V}_{\neq\theta})$, i.e.\ ORMs or PRMs,
with negligible cost relative to generation; and \textbf{self-verifiers}
$(\mathcal{V}_{=\theta})$, which use $\pi_\theta$ to generate new tokens as
part of exploitation (critiques in SR, synthesised responses in
Fusion).

Every TTS method partitions a fixed budget $T$ into exploration tokens $T_e$
and exploitation tokens $T_x$ ($T = T_e + T_x$); per-method accounting is in
Table~\ref{tab:token_accounting}.

\begin{table}[t]
\centering
\small
\caption{Token cost accounting per TTS family under budget $T$.
$\bar{\ell}$: mean generation length; $\bar{\ell}_c$: mean critique length;
$\bar{\ell}_f$: mean fusion output length; $K$: refinement steps; $N$:
candidates; $B$: beam width; $W$: branching factor; $d$: depth.}
\label{tab:token_accounting}
\resizebox{0.70\textwidth}{!}{%
\begin{tabular}{lllll}
\toprule
\textbf{Method} & $T_x$ & $T_e$ & $N$ & \textbf{Exploitation type} \\
\midrule
BoN + ORM      & $\approx 0$     & $\approx T$
               & $\lfloor T/\bar{\ell}\rfloor$ & External scoring \\
Tree search    & $\approx 0$     & $\approx T$
               & $B \times W^d$ (collapses)    & External scoring \\
SR    & $K\bar{\ell}_c$ & $T - K\bar{\ell}_c$
               & $K = T/(\bar{\ell}+\bar{\ell}_c)$ & Generative (sequential) \\
Fusion         & $\bar{\ell}_f$  & $T - \bar{\ell}_f$
               & $N - 1$         & Generative (synthesis) \\
Budget Forcing & $0$             & $T$
               & $1$             & None (single trajectory) \\
\bottomrule
\end{tabular}%
}
\end{table}

\paragraph{Quality metrics.}
Given $x$, let $\mu = \mathbb{E}_{y\sim\pi_{\theta}(\cdot|x)}[\mathcal{V}^*(y)]$
denote single-sample expected quality. The \textbf{oracle quality}
$Q^{*}(T) = \mathbb{E}_{\mathcal{P}}[\max_{y \in \mathcal{P}} \mathcal{V}^{*}(y)]$
is the expected score of the best candidate in the pool. The
\textbf{realised quality}
$Q(T) = \mathbb{E}_{\mathcal{P}}[\mathcal{V}^{*}(\mathcal{E}(\mathcal{P}))]$
is the gold score of the algorithm's actual output. These decompose as:
\begin{equation}
    Q(T) - \mu \;=\;
    \underbrace{(Q^{*}(T) - \mu)}_{\text{exploration headroom}}
    \;+\;
    \underbrace{(Q(T) - Q^{*}(T))}_{\text{net exploitation effect}},
    \label{eq:decomposition}
\end{equation}
where the first term measures the quality available in the pool and the
second measures how effectively the method converts that pool into a final
output. 
$Q^{*}(T)$ is defined as the maximum over the \emph{full} candidate pool
(including synthesised or refined outputs), so the net exploitation effect
is at most zero for every method by construction.

We operationalise exploitation quality through \textbf{headroom capture}:
\begin{equation}
    h \;=\; \frac{Q(T) - \mu}{Q^{*}(T) - \mu},
    \label{eq:headroom}
\end{equation}
the fraction of exploration headroom that the method realises. $h = 1$
means oracle quality is returned; $h = 0$ means no improvement over a
single sample; $h < 0$ means test-time compute hurts performance.

\paragraph{Oracle estimation.} 
A noisy judge $J(y)=\mathcal{V}^*(y)+\epsilon_J$, $\epsilon_J\sim\mathcal{N}(0,\sigma_J^2)$ inflates $\max_i J(y_i)$ above $Q^*(T)$. Assuming an i.i.d. Gaussian pool, we derive (Appendix~\ref{app:oracle_bias}) the entry-wise correction $\hat{O}^{\mathrm{ideal}}=\max_i J(y_i)-a_N[\sqrt{\hat\tau^2+\sigma_J^2}-\hat\tau]$, where $\hat\tau^2=\max(0,s_X^2-\sigma_J^2)$, $s_X^2$ is the empirical candidate variance, and $a_N=\mathbb{E}[\max_{i\le N}Z_i]$ for $Z_i\stackrel{\mathrm{i.i.d}}{\sim}\mathcal{N}(0,1)$ --- all oracle figures use $\hat{O}^{\mathrm{ideal}}$. 


\subsection{Implication 1: The Verifier Bottleneck}
\label{subsec:verifier}

For RM-based methods, headroom capture reduces to a single measurable
quantity. Under i.i.d.\ sampling, oracle quality grows as:
\begin{equation}
    Q^{*}(N) \;\approx\; \mu \;+\;
    \sigma\,\Phi^{-1}\!\!\left(\frac{N}{N+1}\right),
    \label{eq:order_stat}
\end{equation}
at rate $O(\sigma\sqrt{\log N})$. When the external verifier correlates imperfectly with true quality at rate
$\rho_v = \mathrm{Corr}[\mathcal{V}(y), \mathcal{V}^*(y)]$, dividing the
expected BoN quality by oracle headroom yields (derivation in
Appendix~\ref{app:bon-noisy}):
\begin{equation}
    h^{\mathrm{BoN}} \;\approx\; \rho_v.
    \label{eq:headroom_rho}
\end{equation}
Headroom capture for BoN \emph{is} the verifier correlation. When
$\rho_v = 0$, additional compute yields no benefit; when $\rho_v < 0$,
more compute actively harms performance.


\paragraph{Robustness to judge noise.} 
Equation~\ref{eq:headroom_rho} is robust to judge noise $\epsilon_J$ by construction:
as proven in Appendix~\ref{app:sec:rho_bias}, noise enters both $\hat{h}$ and $\hat{\rho}_v$ through the same
attenuation factor, so the two quantities cancel and the identity
$h \approx \rho_v$ holds at the level of raw measured scores without
any correction. Practitioners can therefore use $\hat{\rho}_v$ as a
direct predictor of correct headroom capture. 

\subsection{Implication 2: Oracle Ceilings Differ by Method}
\label{subsec:oracle}
For BoN the candidate pool is $N$ i.i.d.\ samples, so $Q^{*}_{\mathrm{BoN}}(T)$ is the standard order statistic over $N$ draws (Eq.~\ref{eq:order_stat}) --- monotonic in $T$ with diminishing returns. Fusion displaces one i.i.d. candidate with a synthesised output, giving $Q^{*}_{\mathrm{Fusion}}(T) \geq Q^{*}_{\mathrm{BoN}}(T)$ whenever the synthesis matches or exceeds the displaced candidate. SR's pool spans all intermediate and final drafts, so $Q^{*}_{\mathrm{SR}}(T)$ rises above the first-draft baseline only when conditioning on prior drafts genuinely improves quality. Per-method ceilings are studied empirically in Section~\ref{sec:analysis}.


  

\section{Experimental Setup}
\label{sec:setup}

\paragraph{Benchmarks.}
We evaluate on five open-ended generation benchmarks spanning medicine, law,
finance, general chat, and creative writing (Table~\ref{tab:benchmarks}).
None admits exact-match or binary verification. Full details in
Appendix~\ref{app:benchmarks}.

\begin{table}[h]
\centering
\small
\caption{Benchmark characteristics.}
\label{tab:benchmarks}
\resizebox{\textwidth}{!}{%
\begin{tabular}{llcllcc}
\toprule
\textbf{Benchmark} & \textbf{Domain} & \textbf{$n$} &
\textbf{Evaluation} & \textbf{Source} & \textbf{Scoring} &
\textbf{Avg source items} \\
\midrule
LEXam~\cite{fan2025lexam}
  & Law      & 516  & Reference-guided & Professor solutions
  & Holistic 0--1        & 1    \\
HealthBench~\cite{openai2025healthbench}
  & Medicine & 5000 & Rubric-only      & Physician rubrics
  & Binary per criterion & 11.4 \\
PRBench~\cite{scaleai2025prbench}
  & Finance, Law & 1650 & Rubric-only  & Expert rubrics
  & Binary per criterion & 17.7 \\
WildBench~\cite{lin2024wildbench}
  & Chat/General & 1024 & Rubric-only  & LLM-generated checklist
  & Holistic 1--10       & $\sim$11 \\
WritingBench~\cite{wu2025writingbench}
  & Writing  & 555  & Rubric-only      & LLM-generated rubric
  & 1--10 per criterion  & 5    \\
\bottomrule
\end{tabular}%
}
\end{table}

\paragraph{TTS Methods.}

\begin{figure}[t]
    \centering
    \includegraphics[width=\textwidth]{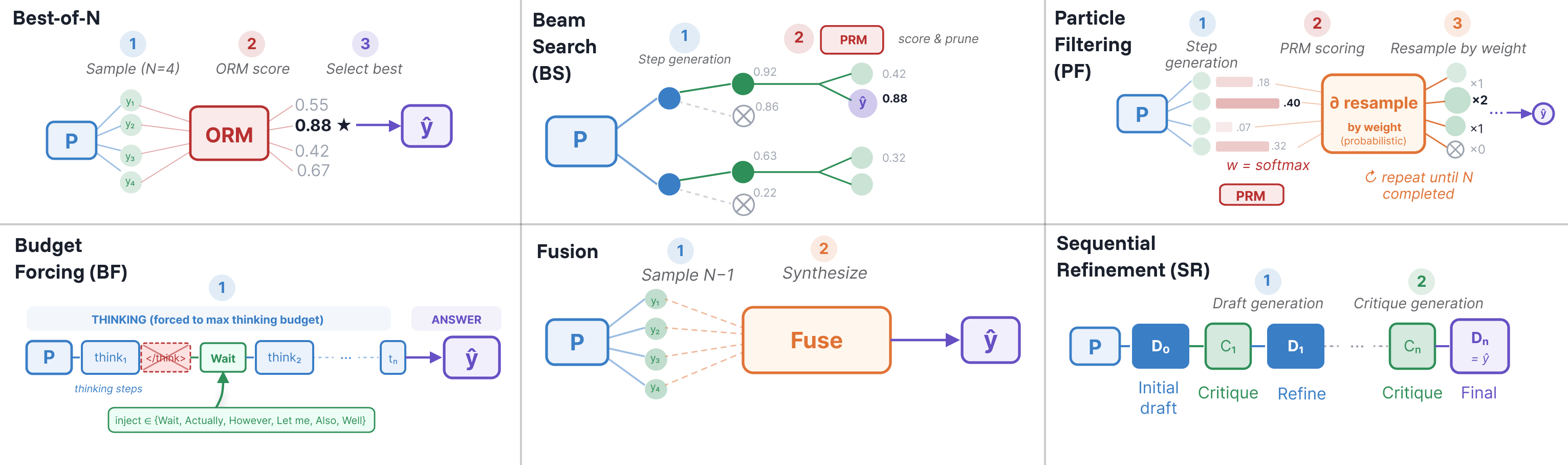}
    \caption{Schema for the six TTS methods in this paper. PF is the only method reproduced as it is, the other methods are variations from previous papers}
    \label{fig:method_schema}
\end{figure}

We evaluate one representative per TTS family (Figure~\ref{fig:method_schema}), with two tree search
variants to disentangle deterministic pruning from stochastic diversity:
\textbf{Best-of-$\boldsymbol{N}$} (BoN), parallel sampling, external ORM;
\textbf{Beam Search} (BS), deterministic tree search, PRM-guided;
\textbf{Particle Filter} (PF) ~\cite{puri2025rollout}, stochastic resampling, PRM-guided;
\textbf{Sequential Refinement} (SR), sequential refinement, self-verifier;
\textbf{Budget Forcing} (BF), extended thinking, single trajectory; and
\textbf{Fusion}, generative synthesis over $N$ candidates.
More details on the algorithms, hyperparameters and prompt templates in Appendix~\ref{app:hyperparams}.

\paragraph{Models.}
Generators: \textbf{OLMo3-7B-Think} and \textbf{OLMo3.1-32B-Think}~\cite{olmo2025olmo3},
and the \textbf{Qwen3.5} family (9B, 35B-A3B)~\cite{qwen35blog}.
ORMs for BoN: \textbf{Skywork-Reward-V2-Llama-3.1-8B}~\cite{skywork2024reward}
and \textbf{Llama-3.1-70B-Instruct-RM-RB2} ~\cite{rewardbench2}.
PRM for tree search: \textbf{VersaPRM-8B}~\cite{versaprm2025}. ORMs were selected based on their performance on RewardBench 2 ~\cite{rewardbench2} while VersaPRM was selected for its unique multi-domain training.

\paragraph{Compute Normalisation.}
We normalise by \emph{total generator output tokens} across all generative
steps, excluding the cost of discriminative reward models whose cost is negligible compared to autoregressive decoding. We define four compute levels:
\textbf{Low}, \textbf{Mid}, \textbf{High}, and \textbf{XHigh}, which match the average BoN output at $N \in \{2, 4, 8, 16\}$ within 25\%
tolerance (Table~\ref{tab:compute_bands}, Appendix~\ref{app:compute}).
Total compute across all experiments amounted to approximately $23{,}800$ GPU-hours on B200 GPUs (Table~\ref{tab:gpu-hours}, Appendix~\ref{app:hyperparams}).



\paragraph{Evaluation.}
Evaluating 27B generated tokens with each benchmark's native judge is cost-prohibitive; we instead use \textbf{Qwen3.5-397B-A17B}~\cite{qwen35blog} as our judge across all benchmarks (Appendix~\ref{app:hyperparams}). We validated the agreement against each benchmark's native judge on
stratified 5--15\% samples (Appendix~\ref{app:judge_agreement}). On
HealthBench, our
judge achieves Macro~F1 of 0.679 with human annotations, above the mean inter-annotator
agreement; on PRBench, criterion-level $\kappa = 0.679$, while WildBench
and WritingBench show respectively QWK 0.564 and 0.408 per-pair agreement. The lower WritingBench agreement reflects the benchmark's known verbosity bias rather than judge miscalibration (Section ~\ref{sec:generative_analysis}):
on identical BoN samples, our judge and the authors' official 7B Critic model~\cite{wu2025writingbench} yield the same headroom capture (0.19 vs.\ 0.19, more in Appendix ~\ref{app:verbosity}).  

\section{Results}
\label{sec:main_results}

Table~\ref{tab:main_qwen35b} reports compute-normalised results for
Qwen3.5-35B-A3B across all methods and benchmarks. Results for Qwen3.5-9B
and OLMo3 are in Appendices~\ref{app:qwen_results} and~\ref{app:olmo}.

\begin{table}[t]
\centering
\setlength{\tabcolsep}{2.5pt}
\tiny
\caption{Realised quality for \textbf{Qwen3.5-35B-A3B} across five
benchmarks and four compute levels. Cell colour encodes delta from the
single-sample baseline (BoN@1):
\colorbox[RGB]{56,142,60}{\textcolor{white}{green}} = improvement,
\colorbox[RGB]{200,60,60}{\textcolor{white}{red}} = regression.
Bold = gain $\geq 0.03$ over baseline.}
\label{tab:main_qwen35b}
\resizebox{\textwidth}{!}{%
\begin{tabular}{l ccccc ccccc ccccc}
\toprule
  & \multicolumn{5}{c}{\textbf{HealthBench}} & \multicolumn{5}{c}{\textbf{PRBench}} & \multicolumn{5}{c}{\textbf{LEXam}} \\
  \cmidrule(lr){2-6} \cmidrule(lr){7-11} \cmidrule(lr){12-16}
  & \footnotesize N=1 & \footnotesize Low & \footnotesize Mid & \footnotesize High & \footnotesize XHigh & \footnotesize N=1 & \footnotesize Low & \footnotesize Mid & \footnotesize High & \footnotesize XHigh & \footnotesize N=1 & \footnotesize Low & \footnotesize Mid & \footnotesize High & \footnotesize XHigh \\
\midrule
  BoN+Skywork & \cellbase{0.536} & \cellcolor[RGB]{252,253,252}0.538 & \cellcolor[RGB]{250,252,250}0.539 & \cellcolor[RGB]{249,252,250}0.539 & \cellcolor[RGB]{249,252,250}0.539 & \cellbase{0.292} & \cellcolor[RGB]{252,253,252}0.293 & \cellcolor[RGB]{252,253,252}0.294 & \cellcolor[RGB]{252,254,252}0.293 & \cellcolor[RGB]{255,254,254}0.292 & \cellbase{0.525} & \cellcolor[RGB]{252,253,252}0.527 & \cellcolor[RGB]{247,250,247}0.529 & \cellcolor[RGB]{243,248,244}0.531 & \cellcolor[RGB]{244,249,244}0.530 \\
  BoN+Llama70B & \cellbase{0.536} & \cellcolor[RGB]{249,251,249}0.539 & \cellcolor[RGB]{246,250,246}0.541 & \cellcolor[RGB]{244,249,245}0.541 & \cellcolor[RGB]{244,249,245}0.541 & \cellbase{0.292} & \cellcolor[RGB]{252,254,252}0.293 & \cellcolor[RGB]{249,252,250}0.295 & \cellcolor[RGB]{247,250,247}0.296 & \cellcolor[RGB]{241,247,242}0.298 & \cellbase{0.525} & \cellcolor[RGB]{248,251,248}0.528 & \cellcolor[RGB]{241,247,242}0.532 & \cellcolor[RGB]{238,246,239}0.533 & \cellcolor[RGB]{236,244,236}0.534 \\
  Fusion & \cellcolor{white}--- & \cellcolor{white}--- & \cellcolor[RGB]{173,209,175}\textbf{0.574} & \cellcolor[RGB]{145,193,148}\textbf{0.587} & \cellcolor[RGB]{144,192,147}\textbf{0.588} & \cellcolor{white}--- & \cellcolor{white}--- & \cellcolor[RGB]{207,228,208}0.315 & \cellcolor[RGB]{186,216,188}\textbf{0.324} & \cellcolor[RGB]{172,208,174}\textbf{0.331} & \cellcolor{white}--- & \cellcolor{white}--- & \cellcolor[RGB]{210,230,211}0.546 & \cellcolor[RGB]{209,229,210}0.546 & \cellcolor[RGB]{209,229,210}0.547 \\
  Beam Search & \cellcolor{white}--- & \cellcolor{white}--- & \cellcolor[RGB]{254,249,249}0.534 & \cellcolor[RGB]{250,232,232}0.526 & \cellcolor[RGB]{253,245,245}0.532 & \cellcolor{white}--- & \cellcolor{white}--- & \cellcolor[RGB]{253,247,247}0.289 & \cellcolor[RGB]{252,239,239}0.285 & \cellcolor[RGB]{252,240,240}0.285 & \cellcolor{white}--- & \cellcolor{white}--- & \cellcolor[RGB]{250,234,234}0.516 & \cellcolor[RGB]{254,255,254}0.526 & \cellcolor[RGB]{255,254,254}0.525 \\
  Particle Filter & \cellcolor{white}--- & \cellcolor{white}--- & \cellcolor[RGB]{253,244,244}0.532 & \cellcolor[RGB]{253,244,244}0.532 & \cellcolor[RGB]{252,240,240}0.530 & \cellcolor{white}--- & \cellcolor{white}--- & \cellcolor[RGB]{252,241,241}0.286 & \cellcolor[RGB]{250,233,233}0.282 & \cellcolor[RGB]{250,231,231}0.282 & \cellcolor{white}--- & \cellcolor{white}--- & \cellcolor[RGB]{242,248,243}0.531 & \cellcolor[RGB]{251,253,251}0.527 & \cellcolor[RGB]{254,255,254}0.526 \\
  SR & \cellcolor{white}--- & \cellcolor[RGB]{247,219,219}0.521 & \cellcolor[RGB]{245,210,210}0.517 & \cellcolor[RGB]{245,209,209}0.516 & \cellcolor[RGB]{244,202,202}0.513 & \cellcolor{white}--- & \cellcolor[RGB]{234,243,235}0.302 & \cellcolor[RGB]{214,232,215}0.311 & \cellcolor[RGB]{180,213,182}\textbf{0.327} & \cellcolor[RGB]{148,194,150}\textbf{0.342} & \cellcolor{white}--- & \cellcolor[RGB]{245,209,209}0.505 & \cellcolor[RGB]{244,206,206}0.504 & \cellcolor[RGB]{236,169,169}0.488 & \cellcolor[RGB]{232,152,152}0.480 \\
  Budget Forcing & \cellcolor{white}--- & \cellcolor[RGB]{249,252,249}0.539 & \cellcolor[RGB]{250,252,250}0.539 & \cellcolor{white}--- & \cellcolor{white}--- & \cellcolor{white}--- & \cellcolor[RGB]{247,251,247}0.296 & \cellcolor[RGB]{249,252,249}0.295 & \cellcolor{white}--- & \cellcolor{white}--- & \cellcolor{white}--- & \cellcolor[RGB]{254,254,254}0.526 & \cellcolor[RGB]{254,249,249}0.523 & \cellcolor{white}--- & \cellcolor{white}--- \\
\bottomrule
\end{tabular}%
}
\vspace{0.4em}
\resizebox{\textwidth}{!}{%
\begin{tabular}{l ccccc ccccc ccccc}
\toprule
  & \multicolumn{5}{c}{\textbf{WildBench}} & \multicolumn{5}{c}{\textbf{WritingBench}} & \multicolumn{5}{c}{\textbf{Overall}} \\
  \cmidrule(lr){2-6} \cmidrule(lr){7-11} \cmidrule(lr){12-16}
  & \footnotesize N=1 & \footnotesize Low & \footnotesize Mid & \footnotesize High & \footnotesize XHigh & \footnotesize N=1 & \footnotesize Low & \footnotesize Mid & \footnotesize High & \footnotesize XHigh & \footnotesize N=1 & \footnotesize Low & \footnotesize Mid & \footnotesize High & \footnotesize XHigh \\
\midrule
  BoN+Skywork & \cellbase{0.812} & \cellcolor[RGB]{228,240,228}0.825 & \cellcolor[RGB]{209,229,210}0.834 & \cellcolor[RGB]{197,222,199}0.839 & \cellcolor[RGB]{190,218,191}\textbf{0.843} & \cellbase{0.702} & \cellcolor[RGB]{243,248,243}0.708 & \cellcolor[RGB]{234,243,234}0.712 & \cellcolor[RGB]{227,239,227}0.715 & \cellcolor[RGB]{221,236,221}0.718 & \cellbase{0.574} & \cellcolor[RGB]{245,249,245}0.578 & \cellcolor[RGB]{238,246,239}0.581 & \cellcolor[RGB]{234,243,234}0.584 & \cellcolor[RGB]{232,242,233}0.584 \\
  BoN+Llama70B & \cellbase{0.812} & \cellcolor[RGB]{239,246,239}0.820 & \cellcolor[RGB]{228,240,229}0.825 & \cellcolor[RGB]{220,235,221}0.828 & \cellcolor[RGB]{217,233,217}0.830 & \cellbase{0.702} & \cellcolor[RGB]{245,249,245}0.707 & \cellcolor[RGB]{238,245,238}0.710 & \cellcolor[RGB]{231,241,231}0.714 & \cellcolor[RGB]{223,237,224}0.717 & \cellbase{0.574} & \cellcolor[RGB]{247,250,247}0.578 & \cellcolor[RGB]{240,247,241}0.580 & \cellcolor[RGB]{236,244,237}0.582 & \cellcolor[RGB]{232,242,233}0.584 \\
  Fusion & \cellcolor{white}--- & \cellcolor{white}--- & \cellcolor[RGB]{187,216,188}\textbf{0.844} & \cellcolor[RGB]{158,200,161}\textbf{0.857} & \cellcolor[RGB]{145,193,147}\textbf{0.864} & \cellcolor{white}--- & \cellcolor{white}--- & \cellcolor[RGB]{227,239,228}0.715 & \cellcolor[RGB]{234,243,234}0.712 & \cellcolor[RGB]{217,233,217}0.720 & \cellcolor{white}--- & \cellcolor{white}--- & \cellcolor[RGB]{201,224,202}0.599 & \cellcolor[RGB]{187,216,188}\textbf{0.606} & \cellcolor[RGB]{177,211,179}\textbf{0.610} \\
  Beam Search & \cellcolor{white}--- & \cellcolor{white}--- & \cellcolor[RGB]{238,246,239}0.820 & \cellcolor[RGB]{254,252,252}0.811 & \cellcolor[RGB]{248,251,248}0.816 & \cellcolor{white}--- & \cellcolor{white}--- & \cellcolor[RGB]{251,239,239}0.695 & \cellcolor[RGB]{252,241,241}0.696 & \cellcolor[RGB]{251,237,237}0.694 & \cellcolor{white}--- & \cellcolor{white}--- & \cellcolor[RGB]{254,248,248}0.571 & \cellcolor[RGB]{253,244,244}0.569 & \cellcolor[RGB]{253,248,248}0.571 \\
  Particle Filter & \cellcolor{white}--- & \cellcolor{white}--- & \cellcolor[RGB]{250,232,232}0.802 & \cellcolor[RGB]{250,230,230}0.801 & \cellcolor[RGB]{247,219,219}0.797 & \cellcolor{white}--- & \cellcolor{white}--- & \cellcolor[RGB]{251,237,237}0.694 & \cellcolor[RGB]{251,237,237}0.694 & \cellcolor[RGB]{248,225,225}0.689 & \cellcolor{white}--- & \cellcolor{white}--- & \cellcolor[RGB]{253,244,244}0.569 & \cellcolor[RGB]{252,241,241}0.567 & \cellcolor[RGB]{250,234,234}0.565 \\
  SR & \cellcolor{white}--- & \cellcolor[RGB]{228,240,229}0.825 & \cellcolor[RGB]{217,234,218}0.830 & \cellcolor[RGB]{216,233,217}0.831 & \cellcolor[RGB]{204,226,205}0.836 & \cellcolor{white}--- & \cellcolor[RGB]{179,212,180}\textbf{0.738} & \cellcolor[RGB]{152,197,154}\textbf{0.750} & \cellcolor[RGB]{126,182,129}\textbf{0.770} & \cellcolor[RGB]{126,182,129}\textbf{0.775} & \cellcolor{white}--- & \cellcolor[RGB]{246,250,246}0.578 & \cellcolor[RGB]{236,244,237}0.583 & \cellcolor[RGB]{228,240,229}0.586 & \cellcolor[RGB]{221,236,222}0.589 \\
  Budget Forcing & \cellcolor{white}--- & \cellcolor[RGB]{230,241,231}0.824 & \cellcolor[RGB]{212,231,213}0.832 & \cellcolor{white}--- & \cellcolor{white}--- & \cellcolor{white}--- & \cellcolor[RGB]{240,184,184}0.672 & \cellcolor[RGB]{236,169,169}0.665 & \cellcolor{white}--- & \cellcolor{white}--- & \cellcolor{white}--- & \cellcolor[RGB]{254,249,249}0.571 & \cellcolor[RGB]{254,248,248}0.571 & \cellcolor{white}--- & \cellcolor{white}--- \\
\bottomrule
\end{tabular}%
}
\end{table}

\paragraph{The defining pattern is stagnation.} BoN with either ORM yields negligible gains across an 8× budget increase. Tree search never improves over the single-sample baseline and actively degrades on several benchmarks. Budget Forcing is flat or regresses even at low compute — these reasoning models exhaust useful thinking within modest token budgets, and we do not scale it further. Yet oracle quality rises steadily with compute on every benchmark (Section~\ref{sec:analysis}): better candidates exist but no selection or search method identifies them.

\paragraph{Both ORMs fail almost identically.} Skywork-Reward-V2 and Llama-3.1-70B-RM produce near-identical realised scores across all benchmarks (overall at XHigh: both 0.584), pointing to a structural rather than model-specific failure quantified in Section~\ref{sec:rm_analysis}.

\paragraph{Fusion is the only method that consistently scales.} It improves over the single-sample baseline on every benchmark, reaching 0.610 overall at XHigh compute (vs.\ 0.584 for the best RM-based method and 0.574 baseline), a pattern that holds across model scales (Section ~\ref{sec:scaling_analysis}).

\paragraph{Sequential Refinement is unreliable.} It achieves the largest gain of any method on WritingBench (+7.3pp) compared to single inference baseline, and outperforms Fusion on PRBench (0.342 vs.\ 0.331), but regresses on HealthBench ($-2.3$pp) and LEXam ($-4.5$pp). As Section~\ref{sec:generative_analysis} shows, the WritingBench gains are confounded by verbosity and the WildBench gains are driven almost entirely by a single subtask.

\section{Analysis}
\label{sec:analysis}

\begin{figure}[t]
    \centering
    \includegraphics[width=\textwidth]{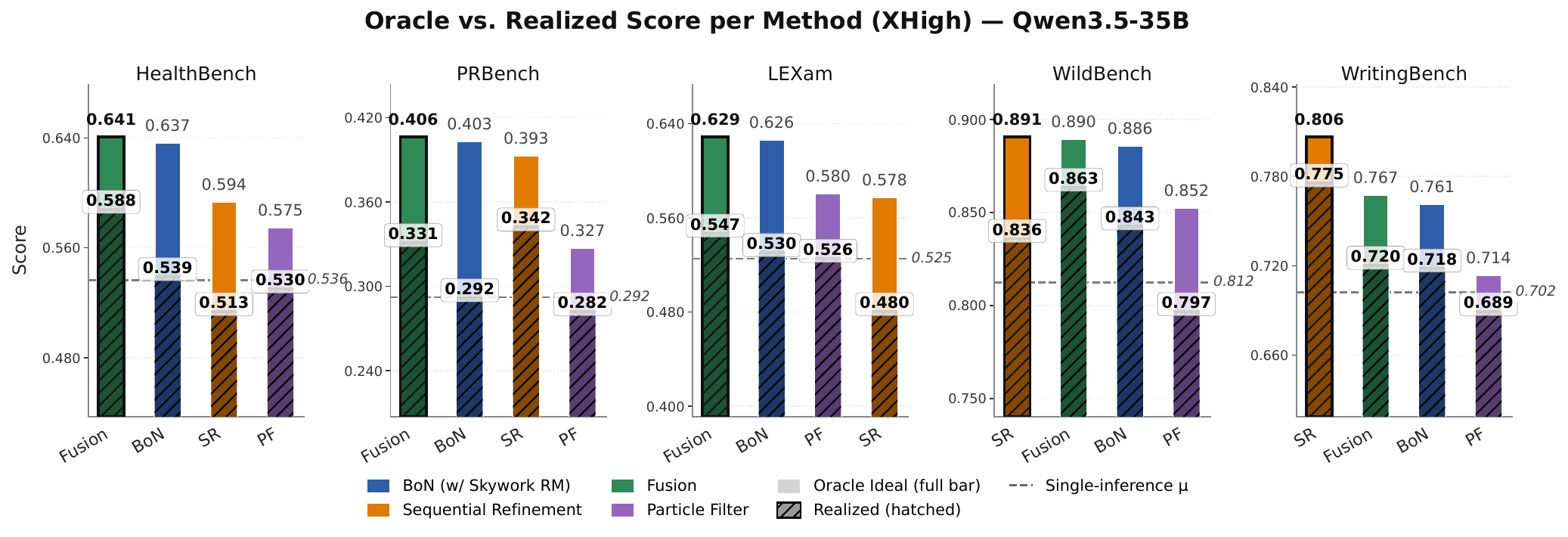}
    \caption{Oracle (full bar) vs.\ realised (hatched) quality per method
    at matched compute (\textbf{XHigh}), Qwen3.5-35B-A3B. All qualitative patterns above replicate under the naive oracle (Appendix~\ref{app:naive_oracle}). In that appendix figure, the naive estimator systematically overestimates oracle quality, so the exploitation gaps shown there are conservative lower bounds on the true exploitation gap.}
    \label{fig:oracle_quality}
\end{figure}

The explanation reduces to one observation: \textbf{the candidate pool is not the problem---exploitation is}. Figure~\ref{fig:oracle_quality} makes this concrete. At the highest compute budget, all four methods produce candidate pools with high oracle quality across all five benchmarks. Fusion and BoN generate nearly identical oracle pools (typically within $0.01$--$0.04$ oracle difference), yet their realised scores diverge consistently: averaged across benchmarks (Figure~\ref{fig:correlation_and_exploitation}), Fusion captures ${\sim}40\%$ of available quality while BoN captures only ${\sim}15\%$. Sequential Refinement is erratic, ranging from $-0.86$ to $0.70$ across benchmarks, and Particle Filter averages around ${-}40\%$, actively degrading quality. The bottleneck is not generating good candidates but selecting or synthesising the right answer from them. Table~\ref{tab:method_diagnosis} summarises the failure mode for each method family.

\begin{table}[t]
\centering
\small
\caption{Diagnosis per method family. Oracle quality reflects exploration;
headroom capture reflects exploitation.}
\label{tab:method_diagnosis}
\resizebox{\textwidth}{!}{%
\begin{tabular}{p{1.7cm}|p{4.3cm}|p{3.6cm}|p{6cm}}
\toprule
\textbf{Method} & \textbf{Oracle quality} & \textbf{Headroom capture} &
\textbf{Failure mode} \\
\midrule
BoN + ORM
  & High
  & ${\sim}15\%$ ($\hat{\rho}_v \approx 0.12$)
  & RM miscalibrated (\S\ref{sec:rm_analysis}) \\[2pt]
Tree search
  & Collapsed (diversity loss)
  & Negative on most tasks
  & PRM destroys diversity (\S\ref{sec:tree_failure}) \\[2pt]
Fusion
  & $\geq$ BoN
  & ${\sim}40\%$ (best, still low)
  & Residual gap (\S\ref{sec:generative_analysis}) \\[2pt]
SR
  & Lower than BoN on 3/5 benchmarks
  & Genuine gain on 1/5 (PRBench)
  & Refinement mostly hurts; where it helps, gains trace to
    subtask composition or verbosity  (\S\ref{sec:generative_analysis}) \\
\bottomrule
\end{tabular}%
}
\end{table}

\subsection{Verifier Correlation Explains Selection Failure}
\label{sec:rm_analysis}

\begin{figure}[t]
    \centering
    \includegraphics[width=0.43\textwidth]{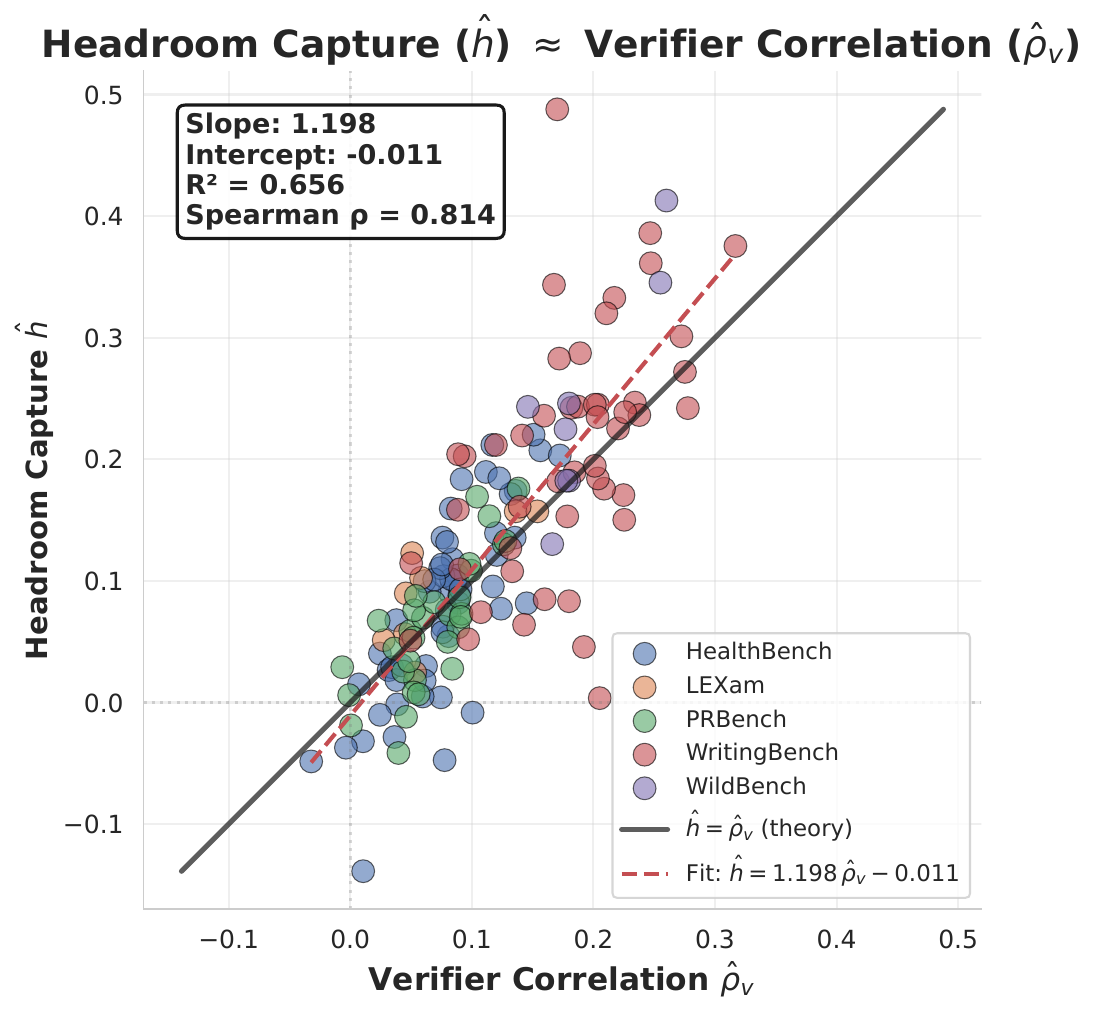}
    \hfill
    \includegraphics[width=0.56\textwidth]{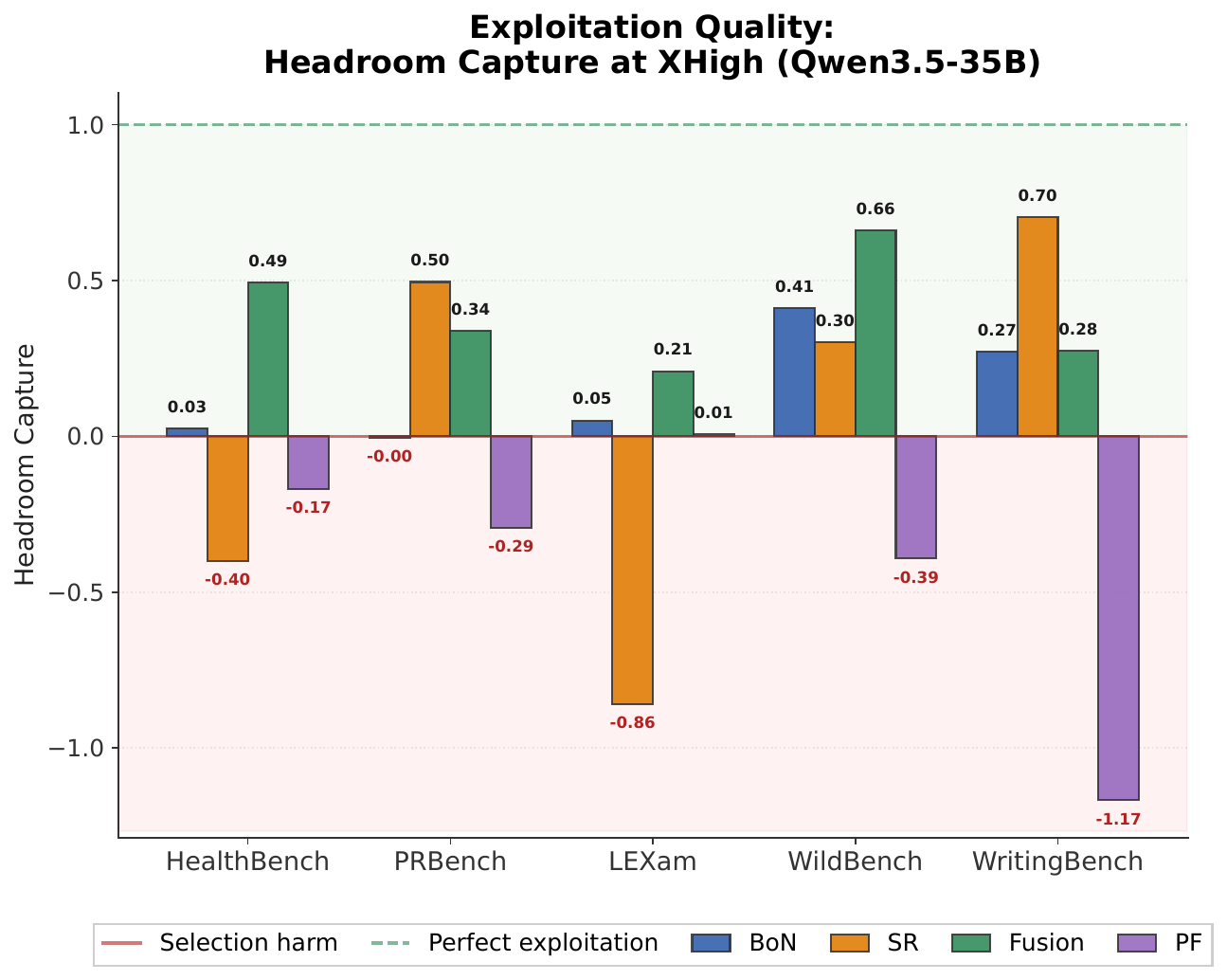}

    \caption{\textbf{Left:} $\hat{h}$ vs.\ $\hat{\rho}_v$ across $N=152$ task-level points.
Identity line $y=x$ is the theoretical prediction (Equation~\ref{eq:headroom_rho});
empirical fit $y = 1.198\,\hat{\rho}_v - 0.011$ confirms it
($R^2 = 0.66$, $\rho = 0.81$, $p < 10^{-36}$).
    \textbf{Right:} Headroom capture per method at XHigh compute. The headrooms for all models are in Appendix \ref{app:headroom_tables_adjusted}}
    \label{fig:correlation_and_exploitation}
\end{figure}

\paragraph{Current reward models are insufficient to improve strong generators via BoN.} The effectiveness of BoN depends critically on the generator-verifier gap. When the generator is weak relative to the RM, BoN yields clean positive scaling: our reproduction of the original Skywork-RM experiment, swapping in the much weaker generator from that paper, recovers strong TTS scaling (Appendix~\ref{app:skywork_reproduction}). When the generator is strong, however, we find that current reward models struggle to discriminate reliably among candidate outputs. On MATH-500 and GPQA Diamond, our generators gain only ${\sim}2$pp over single-sample (Appendix~\ref{app:deterministic}); on open-ended generation, Spearman correlations between RM scores and judge scores average just $0.12$ (Skywork) and $0.11$ (Llama) across 152 (generator, RM, benchmark) combinations (Appendix~\ref{app:rho_v_tables}), reducing BoN to near-random selection. 

\paragraph{$\boldsymbol{\hat{\rho}_v}$ quantitatively predicts headroom capture.}
Theory predicts the identity $h^{\mathrm{BoN}} \approx \rho_v$
(Equation~\ref{eq:headroom_rho}).
Regressing $\hat{h}$ on $\hat{\rho}_v$ across all $N=152$ combinations
yields slope $1.198$ and intercept $-0.011$: slope near $1$ and intercept
near $0$ directly validate the prediction
($R^2 = 0.66$, $\rho = 0.81$, $p < 10^{-36}$) Figure~\ref{fig:correlation_and_exploitation} (left)).
Figure~\ref{fig:correlation_and_exploitation} (right) shows the concrete
consequence: the same ${\sim}15\%$ vs.\ ${\sim}40\%$ headroom capture gap
between RM-based BoN and Fusion, now explained by $\hat{\rho}_v \approx  0.12$.

\subsection{Tree Search Failure: Diversity Collapse}
\label{sec:tree_failure}

\begin{figure}[t]
    \centering
     \includegraphics[width=0.48\textwidth]{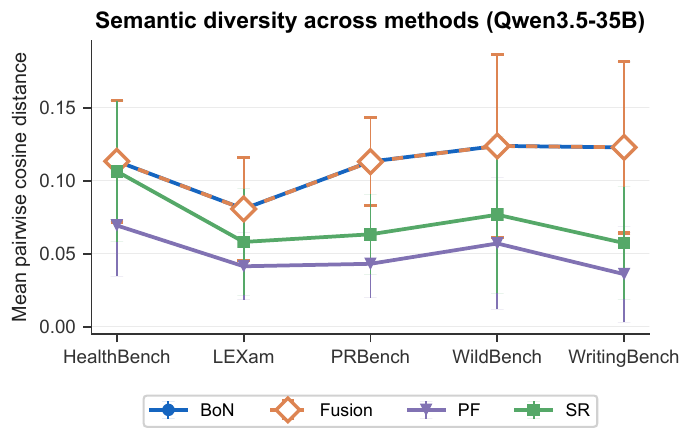}
    \hfill
    \includegraphics[width=0.48\textwidth]{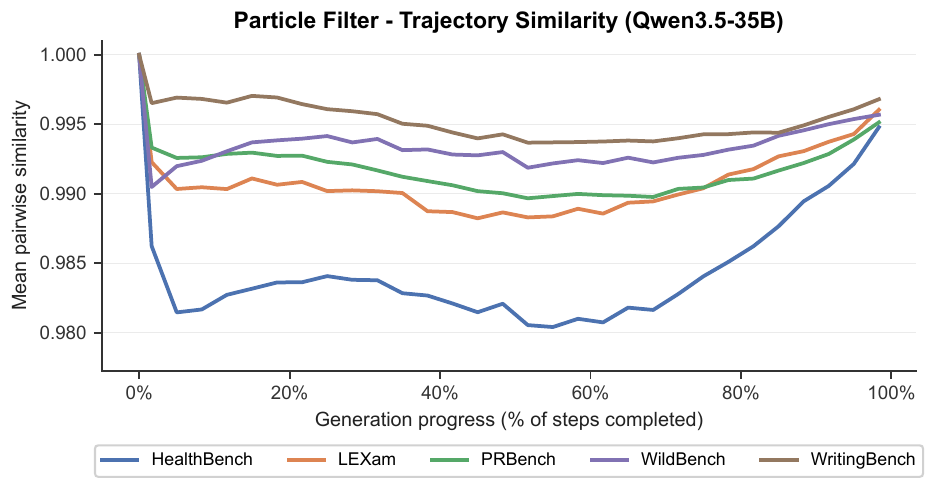}
    \caption{
    \textbf{Left:} Mean pairwise cosine distance across final outputs at High compute.
    \textbf{Right:} Particle trajectory similarity over normalised generation depth. 
    \label{fig:pf_failure}}
\end{figure}

While BoN fails at exploitation through poor selection, tree search fails
at both exploitation \emph{and} exploration: PRM-guided resampling collapses the candidate pool itself. Particle Filtering produces the least diverse final outputs across all benchmarks: for
Qwen3.5-35B, PF's mean cosine distance ranges from 0.036 (WritingBench)
to 0.069 (HealthBench), compared with 0.123--0.124 for BoN (Figure~\ref{fig:pf_failure}, left). Tracing
pairwise similarity across 16 particles over generation depth,
WritingBench particles remain near-identical throughout (similarity
$\geq 0.997$)---16 nominal particles are effectively a single trajectory (Figure~\ref{fig:pf_failure}, right).
The mechanism is the exponential pruning sensitivity from Equation~\ref{eq:pruning}:
a miscalibrated PRM prunes promising branches at every step, and the
probability that the best candidate survives decays exponentially with
depth. Naive parallel sampling preserves more diversity at equal cost.

\subsection{Generative Methods: Fusion Synthesises, Sequential Refinement Regresses}
\label{sec:generative_analysis}

\begin{figure}[t]
    \centering
    \includegraphics[width=0.44\textwidth]{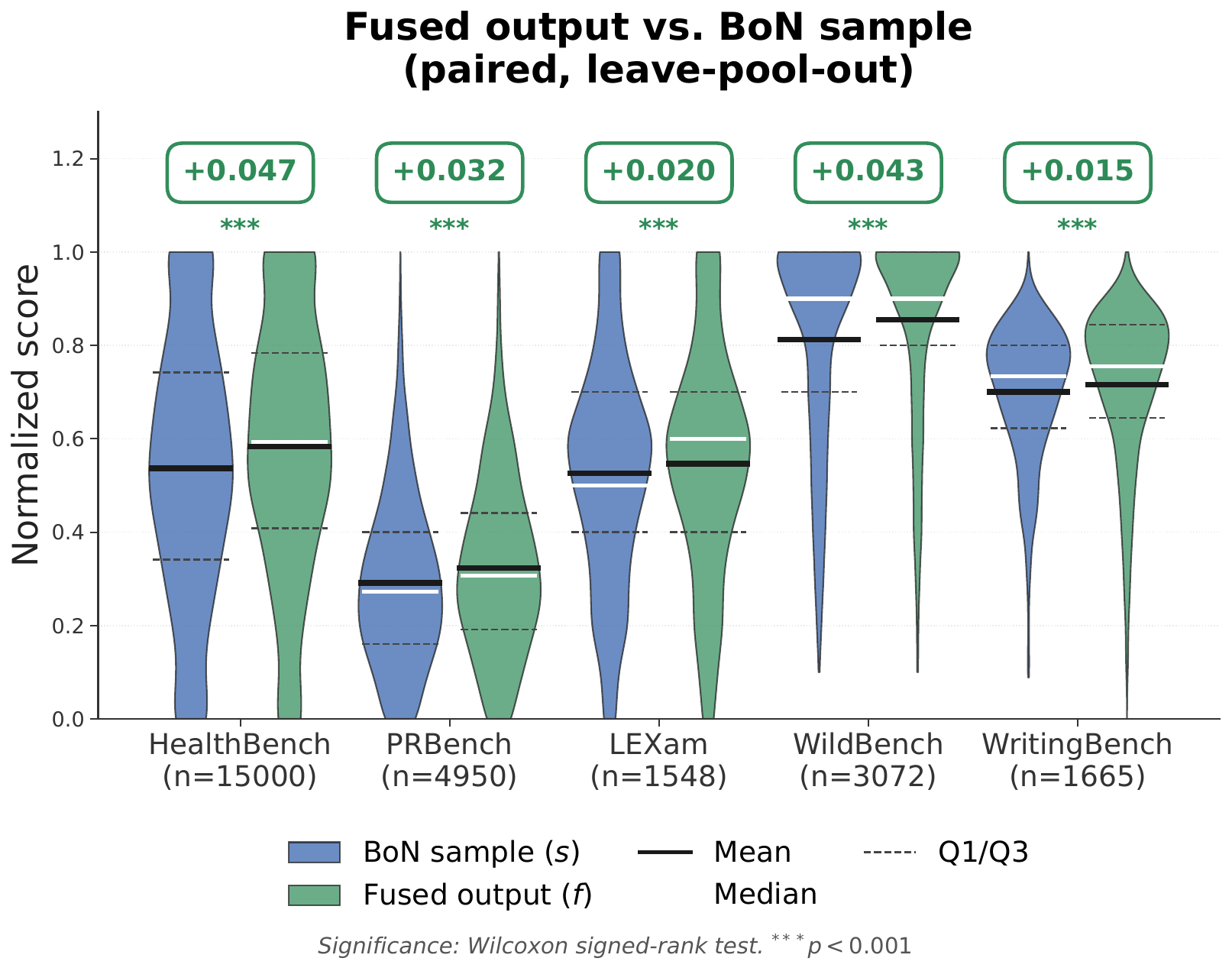}
    \hfill
    \includegraphics[width=0.55\textwidth]{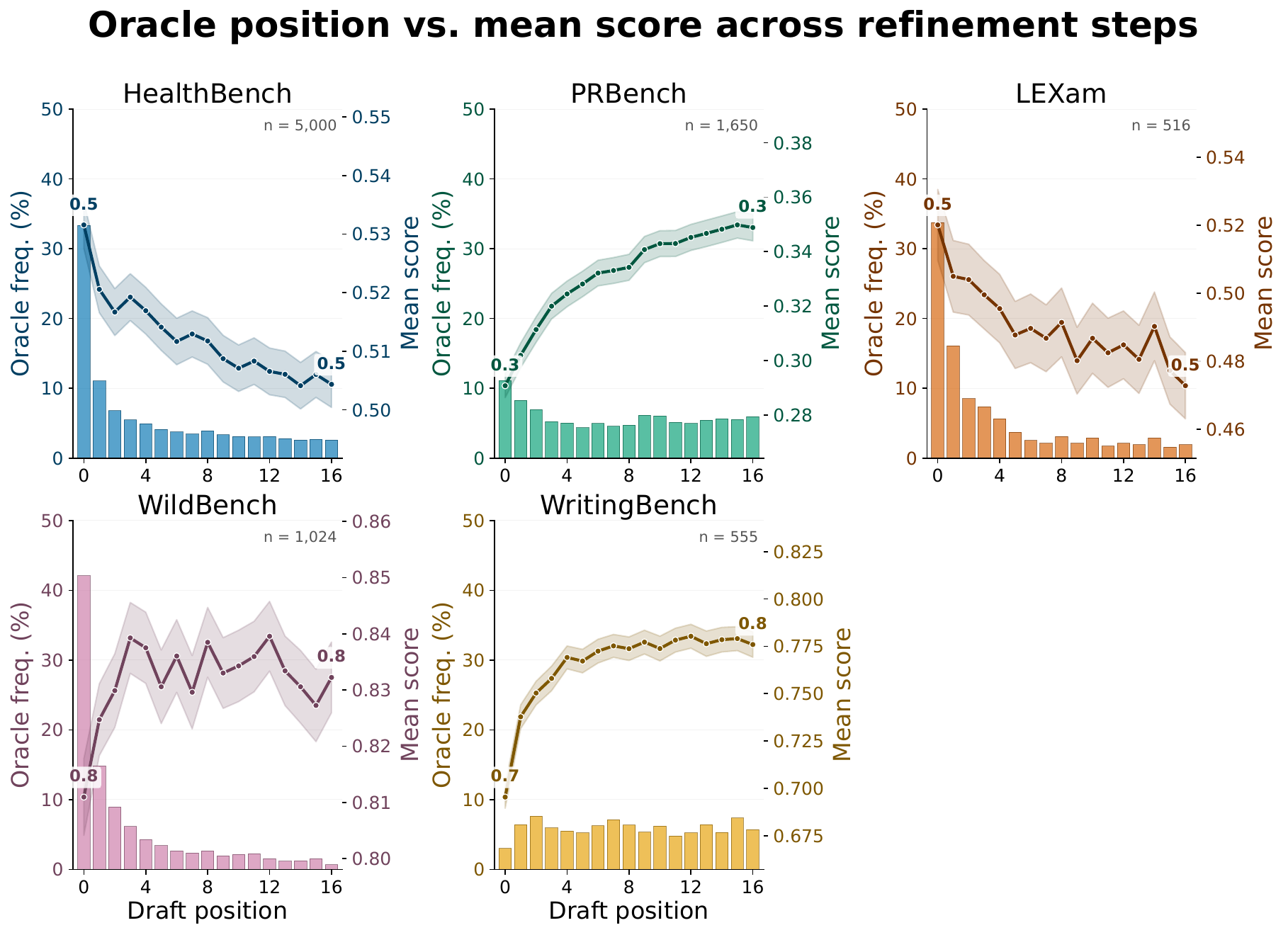}
    \caption{\textbf{Left:} Fusion output quality vs.\ a random candidate drawn from the candidates \emph{not} included in the shared sub-pool common to both BoN and Fusion, per benchmark. \textbf{Right:} Oracle-position frequency (bars, left axis) and mean score trajectory (line, right axis) across SR iterations.}
    \label{fig:fusion_self_refine_analysis}
\end{figure}

\paragraph{Fusion: synthesis exceeds selection.}
 Despite generating fewer independent candidates than BoN at equal compute, Fusion matches or exceeds BoN's oracle ceiling on all benchmarks (Figure~\ref{fig:oracle_quality}), meaning the synthesised output frequently lifts the pool ceiling more than another i.i.d. sample.  Figure~\ref{fig:fusion_self_refine_analysis} (left) shows that the fused output also consistently scores above a held-out
i.i.d. candidate drawn from outside the shared sub-pool used by both BoN
and Fusion---making this a fair comparison of synthesis quality against
an independent sample of equivalent compute cost. Together, the two figures rule out a variance-only interpretation: synthesis raises mean candidate quality, not just the upper tail, so Fusion's ceiling gain reflects systematically better candidates rather than fatter-tailed sampling.

\paragraph{Sequential Refinement: refinement is benchmark-dependent.}
On the exploration side, SR's oracle ceiling falls below BoN's on three of five benchmarks (Figure~\ref{fig:oracle_quality}), confirming that critique tokens displace more value than they add on most tasks. On the exploitation side, the per-iteration trajectory (Figure~\ref{fig:fusion_self_refine_analysis}, right) reveals benchmark-dependent behaviour. On HealthBench and LEXam, refinement is counterproductive: the initial draft is most often the oracle and mean score declines with each iteration. On WildBench, mean score rises despite the initial draft being the oracle 43\% of the time; however, this gain traces to a single subtask (Coding~\&~Debugging), which contributes more than the entire aggregate lift; excluding it produces a net regression (Appendix~\ref{app:wildbench_coding}). Only PRBench exhibits genuine improvement: oracle frequency is nearly uniform across iterations and mean score rises across all subtasks. WritingBench is discussed separately below.


\paragraph{The WritingBench exception is confounded by verbosity.}
WritingBench is the apparent exception: the oracle ceiling for
SR exceeds BoN's and the best draft shifts to later
iterations. However, the within-prompt length--score correlation is
$\hat{\rho} = +0.33$ for SR (vs.\ $+0.20$ for BoN), and the judge discriminates more on length relative to substance than other benchmarks do (PRBench: $4.3\times$ content-to-length discrimination; HealthBench: $2.5\times$; WritingBench: $0.7\times$). The fine-tuned judge released by the WritingBench authors shows an even stronger length--score correlation, confirming the verbosity bias is a
property of the benchmark's evaluation design, not of any particular judge. Full analysis in Appendix~\ref{app:verbosity}.

\subsection{Generation and Exploitation Are Distinct Capabilities}
\label{sec:scaling_analysis}
Self-verifier quality is bounded by generator capability~(Appendix~\ref{app:theory}), but capability is not reducible to parameter count. To separate scale from family, we compare Fusion and SR within Qwen3.5 and across families (Qwen3.5 vs.\ OLMo3); full per-model tables are in Appendices~\ref{app:olmo} and~\ref{app:qwen_results}; full model-capacity decomposition in Appendix~\ref{sec:model_capacity_study}. 

\paragraph{Generation capability does not imply exploitation capability.}                         
The most revealing comparison is cross-family. Qwen3.5 and OLMo3 produce candidate pools with comparable headrooms on HealthBench and LEXam (Figure~\ref{fig:fusion-oracle-gaps}), and on WritingBench OLMo3 in fact shows \emph{larger} headrooms. Yet headroom capture diverges sharply (Figure~\ref{fig:model-families-headroom}): at High compute, Qwen3.5 models (9B and 35B-A3B) close $\sim36\%$ of headroom on average under Fusion, while OLMo3.1-32B-Think produces \emph{negative} headroom on two of three benchmarks, meaning its synthesised output is worse than a random candidate.  OLMo3 generates pools with as much (or more) oracle gap as Qwen3.5 but largely fails to convert that gap into a better answer.                                                                         
\paragraph{Scaling generation capability within a family does not improve exploitation either.} Within Qwen3.5, both Fusion and SR exhibit the same qualitative behaviour at every model size we tested. Fusion improves over the single-sample baseline on every benchmark and at every scale, but its advantage relative
to a random candidate does not grow with model size --- consistent with synthesis being a structural strategy (bypassing selection) rather than a capability-dependent one. SR's behaviour is benchmark-dependent --- regression on HealthBench and LEXam, gains on WritingBench and PRBench --- with comparable exploitation across model sizes (Figure \ref{fig:self-verifier-abs-gain-per-bench}). 

\section{Conclusion}
\label{sec:conclusion}
We present the first systematic, compute-normalised comparison of five TTS families on open-ended generation, grounded in a unified exploration--exploitation framework, and find that exploitation — not exploration — is the bottleneck. Oracle quality rises steadily with compute, but no method converts more than a fraction into realised quality: RM-based selection fails structurally ($\hat{\rho}_v \approx 0.12$, both ORMs identical), tree search compounds this through diversity collapse, Fusion is the only consistently improving method yet captures only
${\sim}40\%$ of available headroom, and SR yields genuine gains on only one benchmark. These failures are not unique to open-ended generation — they arise whenever generators outpace RM training data and no scaling axis closes the gap.


\paragraph{Broader Impact and Limitations}
$\hat{\rho}_v$ provides a cheap diagnostic: practitioners can measure verifier-quality alignment on a small sample before committing inference budget. This matters most in high-stakes domains (medicine, law, finance), where RM-based TTS can actively degrade performance; our results reinforce that LLM-based evaluation is not a substitute for expert review, and deploying TTS in such settings without human oversight remains inadvisable.Our conclusions rest on two model families, and a single unified judge (Qwen3.5-397B-A17B), and do not cover all open-ended use cases. Our oracle estimator $\hat{O}^{\mathrm{ideal}}$ assumes i.i.d.\ candidates---exact for BoN, but only approximate for Refinement, Fusion, and Particle Filtering; we expect it to slightly underestimate the true oracle for the first two and slightly overestimate it for Particle Filtering (Appendix~\ref{app:oracle_bias}).

\paragraph{Future work.}
Two open problems follow from the exploitation gap. First, training
verifiers calibrated for open-ended evaluation: $\hat{\rho}_v$ provides a
measurable target. Second, developing exploitation mechanisms beyond
selection and synthesis: the high unrealised headroom 
represents a substantial opportunity.

\section*{Acknowledgements}
We sincerely thank our colleague, Guglielmo Bonifazi, for his valuable input and insightful discussion points throughout the development of this work. We also express our gratitude to Shirsha Ray for her support in facilitating Kanak's collaboration with our team, which was instrumental in achieving these results.


\bibliographystyle{abbrvnat}
\bibliography{references}

\appendix


\section{Compute Band Configurations}
\label{app:compute}

\begin{table}[H]
\centering
\small
\caption{Compute band configurations. Fusion High uses subset $= 7$ for
Qwen3.5 and subset $= 6$ for OLMo3. Entries marked --- were not run.
$\bar{\ell}_t$: mean thinking tokens per BoN sample;
$\bar{\ell}$: mean answer tokens per BoN sample.
$^{\dagger}$Two tasks ran at $W=3,k=4$ instead of $W=4,k=4$
(see Table~\ref{tab:bs-high}).}
\label{tab:compute_bands}
\resizebox{\textwidth}{!}{
\begin{tabular}{lllllll}
\toprule
\textbf{Band} & \textbf{BoN} & \textbf{Beam Search} & \textbf{PF} &
\textbf{Sequential Refinement} & \textbf{Budget Forcing} & \textbf{Fusion} \\
\midrule
Low   & $n=2$  & ---           & ---     & iter$=3$  & $2\bar{\ell}_t + \bar{\ell}$   & ---          \\
Mid   & $n=4$  & $W=2,k=2$    & $N=4$   & iter$=6$  & $4\bar{\ell}_t + \bar{\ell}$   & subset$=3$   \\
High  & $n=8$  & $W=2,k=4$    & $N=8$   & iter$=12$ & $8\bar{\ell}_t + \bar{\ell}$   & subset$=6/7$ \\
XHigh & $n=16$ & $W=4,k=4^{\dagger}$ & $N=16$  & iter$=16$ & $16\bar{\ell}_t + \bar{\ell}$ & subset$=15$  \\
\bottomrule
\end{tabular}}
\end{table}

\section{Benchmark Details}
\label{app:benchmarks}

\paragraph{LEXam.}
We use the \texttt{open\_question} configuration of
\texttt{LEXam-Benchmark/LEXam} (HuggingFace), drawn from law-school
exams administered at a Swiss university. The original release
contains 2{,}841 items across four legal areas (Private, Public,
Interdisciplinary, Criminal) in English and German. We filter to
\texttt{language == "en"}, retaining 516 items and dropping 2{,}325
German items. The English subset is dominated by international and
comparative law (jurisdiction: International 429, Generic 73, Swiss
14), reflecting the language of instruction at the source institution.
The pool is heavily skewed toward Private and Public law (85\% of
items combined), with Interdisciplinary (48) and Criminal (31)
forming the long tail. Each item carries a single professor reference
solution.

\paragraph{HealthBench.}
We use \texttt{openai/HealthBench} (5{,}000 items). Every example
carries exactly one \texttt{theme} tag; the tags are non-overlapping
and cover the full corpus, so we partition the dataset into seven
sub-files used as separate evaluation tasks: \texttt{global\_health}
(1{,}097), \texttt{hedging} (1{,}071), \texttt{communication} (919),
\texttt{context\_seeking} (594), \texttt{emergency\_referrals} (482),
\texttt{health\_data\_tasks} (477), \texttt{complex\_responses} (360). Each example carries on average
$11.4$ physician-authored rubric criteria, each tagged with a signed
integer point value: positive points reward desired behaviors,
negative points penalize undesired ones (e.g.\ recommending an unsafe
action). The judge marks each criterion \emph{met} or \emph{not met}.
An example's score is the sum of points across met criteria, divided
by the sum of all \emph{positive} points in the rubric (the maximum
positive total). The dataset-level metric is the mean across examples,
clipped to $[0,1]$.

\paragraph{PRBench.}
We use the four pre-split configurations of \texttt{ScaleAI/PRBench}:
\texttt{prbench\_finance} (600), \texttt{prbench\_finance\_hard}
(300), \texttt{prbench\_legal} (500), \texttt{prbench\_legal\_hard}
(250). Each item carries between 10 and 30 weighted rubric criteria
(29{,}252 criteria across 1{,}650 items, average $\sim 17$ per item).
Each criterion is tagged with one of six \texttt{weight\_class} levels,
but the per-criterion integer weight is not a fixed mapping from class
to value: the dataset stores a specific weight per criterion drawn
from the ranges in Table~\ref{tab:prbench_weights}, so a single
example's rubric can contain criteria with weights spanning
$[-10, +10]$. We report the clipped score: per-example
$s_i = \mathrm{raw}_i\big/w^{+}_{\max,i}$ where $w^{+}_{\max,i}$ is the
maximum positive weight present in example $i$'s rubric, and the
dataset-level score is $\max\!\bigl(0,\;\overline{s_i}\bigr)$.

\begin{table}[ht]
\centering
\small
\caption{PRBench weight class to integer point ranges.}
\label{tab:prbench_weights}
\begin{tabular}{ll}
\toprule
\textbf{Weight class} & \textbf{Integer weight} \\
\midrule
critically important    & $\{8, 9, 10\}$ \\
important               & $\{4, 5, 6, 7\}$ \\
slightly important      & $\{1, 2, 3\}$ \\
slightly detrimental    & $\{-3, -2, -1\}$ \\
detrimental             & $\{-7, -6, -5, -4\}$ \\
critically detrimental  & $\{-10, -9, -8\}$ \\
\bottomrule
\end{tabular}
\end{table}

\paragraph{WildBench.}
We use the \texttt{v2} configuration of \texttt{allenai/WildBench}
(1{,}024 items). Items are tagged with one of 11
\texttt{primary\_tag} categories, with a heavy long tail:
Information seeking (182), Coding \& Debugging (170), Creative Writing
(146), Reasoning (133), Planning (116), Math (87), Editing (82), Data
Analysis (33), Role playing (30), Brainstorming (24), Advice seeking
(21). Each example carries a task-specific checklist of 6--33 items
(avg $\sim 11$). \emph{Deviation from upstream that affects absolute
scores:} the WildBench paper macro-averages WB-Score across five task
groups (equal weight per group regardless of group size); we
micro-average across all 1{,}024 items, so larger categories dominate.
This shifts our absolute WB-Scores relative to the official
leaderboard but preserves relative ordering between TTS methods
evaluated on the same item set, which is the comparison we make.

\paragraph{WritingBench.}
We use \texttt{X-PLUG/WritingBench} (\texttt{benchmark\_all.jsonl}). The original release contains 1{,}000 items in Chinese and English; we filter to \texttt{lang == "en"}, retaining 555 items and dropping 445 Chinese items. The remaining items split fairly evenly across six domains: \texttt{finance\_business} (115), \texttt{academic\_engineering} (107), \texttt{politics\_law} (99), \texttt{literature\_arts} (96), \texttt{advertising\_marketing} (74), \texttt{education} (64). Unlike the other rubric-only benchmarks, WritingBench rubrics are \emph{dynamic per query}: every item ships with exactly five criteria, each consisting of a name, a free-text \texttt{criteria\_description}, and five score-band descriptions (\texttt{1--2}, \texttt{3--4}, \texttt{5--6}, \texttt{7--8}, \texttt{9--10}). Each criterion is scored on $1$--$10$; the example score is the mean across the five criteria.

\paragraph{Multi-turn handling.}
Three of the five benchmarks contain multi-turn items: HealthBench ($41.7\%$, max 19 messages), PRBench ($42.2\%$, max 19 messages), and WildBench ($27.0\%$, max 9 messages); LEXam and WritingBench are single-turn only. Multi-turn conversations are passed to both the generator and the judge as ordered \texttt{[\{role, content\}]} message lists, following the standard OpenAI / vLLM chat-completion convention.

\paragraph{Native judges.}
Each benchmark ships with a recommended evaluation judge whose model and protocol we adopt verbatim from the upstream codebase (Table~\ref{tab:native_judges}); these natives serve as the ground-truth comparator in our judge-agreement analysis (Sec~\ref{app:judge_agreement}).

\begin{table}[H]
\centering
\small
\caption{Native judges adopted verbatim from each benchmark's upstream release. Calls/example is the average number of judge
invocations per candidate completion under the native protocol; per-criterion designs (HealthBench, PRBench, WritingBench) fan out into many calls per example.}
\label{tab:native_judges}
\begin{tabular}{lllc}
\toprule
\textbf{Benchmark} & \textbf{Native judge} & \textbf{Protocol} &
\textbf{Calls/ex} \\
\midrule
LEXam        & GPT-4o + Qwen3-32B + DeepSeek-V3 & holistic, min-agg ensemble  & $3$        \\
HealthBench  & GPT-4.1                          & per-criterion binary        & $\sim 11$  \\
PRBench      & o4-mini                          & per-criterion binary        & $\sim 17$  \\
WildBench    & GPT-4.1                          & holistic + checklist        & $1$        \\
WritingBench & Claude 3.7 Sonnet                & per-criterion $1$--$10$     & $5$        \\
\bottomrule
\end{tabular}
\end{table}

\section{Hyperparameters and Prompt Templates}
\label{app:hyperparams}

This appendix documents (i)~the sampling and serving configuration used
across all generation calls, (ii)~method-specific algorithmic settings
beyond the compute-band axis already reported in
Appendix~\ref{app:compute}, (iii)~reward and process reward model
configuration, and (iv)~the prompt templates used by Sequential Refinement and
Fusion. Methods not listed under prompt templates use only the dataset's
prompt together with the system prompt below.

\paragraph{Sampling.}
All generation uses the sampling profile per generator family in Table~\ref{tab:sampling_profile}, with seed $42$ throughout. The system prompt is identical across all benchmarks and methods:
\texttt{"Please complete the following user request."}
Table~\ref{tab:sampling_profile} reports the canonical profile per generator. Fusion synthesis and Budget Forcing reduce \texttt{max\_tokens} from $16{,}384$ to $8{,}192$. The serving context length depends on generator and method: OLMo3 runs at $65{,}536$ throughout (the model's
maximum supported context length); Qwen3.5 runs at $65{,}536$ for BoN, Beam Search, Particle Filter, and Budget Forcing, raised to $131{,}072$ for Sequential Refinement to accommodate accumulating revision history and for Fusion at subset sizes $3$ and $7$, and to $262{,}144$ for Fusion at subset size $15$ to fit the larger candidate pool. Sequential Refinement and
Fusion strip everything up to and including the \texttt{</think>} tag from each draft before passing it to the next step.

\begin{table}[H]
\centering
\small
\caption{Canonical sampling profile per generator family.}
\label{tab:sampling_profile}
\begin{tabular}{lccccc}
\toprule
\textbf{Generator} & \textbf{Temperature} & \textbf{Top-$p$} &
\textbf{Top-$k$} & \textbf{Presence penalty} & \textbf{Thinking mode} \\
\midrule
Qwen3.5 (9B, 35B-A3B) & 1.0 & 0.95 & 20 & 1.5 & enabled \\
OLMo3 (7B, 32B)       & 0.8 & 0.9  & -1 & 0  & enabled \\
\bottomrule
\end{tabular}
\end{table}

\paragraph{Method configurations.}
Compute-band axes ($n$ for BoN, $W{,}k$ for Beam Search, $N$ for
Particle Filter, iteration count for Sequential Refinement, subset size for
Fusion) are listed in Table~\ref{tab:compute_bands}.
Table~\ref{tab:method_hyperparams} records the remaining fixed
hyperparameters per method.

\begin{table}[H]
\centering
\small
\caption{Method-specific algorithmic settings beyond the compute-band
axis.}
\label{tab:method_hyperparams}
\begin{tabular}{lp{0.72\textwidth}}
\toprule
\textbf{Method} & \textbf{Settings} \\
\midrule

BoN
  & XHigh pool ($N{=}16$) generated once; lower bands subsampled
    without replacement; selection $= \operatorname{argmax}$ ORM score. \\

Beam Search
  & $W$ beams, $k$ candidates/step (Table~\ref{tab:compute_bands});
    delimiter \texttt{\textbackslash n\textbackslash n};
    scored by min per-step PRM on full path;
    top-$W$ kept globally; EOS $=$ beam complete;
    PRM: VersaPRM-8B; depth 300 (500 for deterministic). \\

Particle Filter
  & $N$ particles (Table~\ref{tab:compute_bands});
    delimiter \texttt{\textbackslash n\textbackslash n};
    loop order: score $\to$ resample $\to$ generate;
    multinomial resampling with replacement,
    weights $= \mathrm{softmax}(\mathbf{w},\,T{=}1.0)$;
    PRM: VersaPRM-8B; max iterations 300. \\

Sequential Refinement
  & Forced iterations (stop-check ignored);
    history window $K{=}3$ (draft, feedback) pairs;
    feedback \& stop-check: thinking disabled, stop-check at $T{=}0$;
    \texttt{</think>} stripped from drafts before each prompt call. \\

Budget Forcing
  & 2-phase decode: Phase~1 suppresses end-of-thinking tag and injects
    a wait-string to extend thinking; Phase~2 generates answer freely
    from accumulated context. \\

Fusion
  & Random subset without replacement from precomputed BoN-$16$ pool;
    candidates shuffled and \texttt{</think>}-stripped before synthesis;
    single-shot direct synthesis with thinking enabled. \\

\bottomrule
\end{tabular}
\end{table}

\paragraph{BoN.}
Generate $N$ independent completions, score each with an ORM, return the argmax.
For compute bands Low through XHigh, a pool of $N{=}16$ completions is
generated once per prompt (the XHigh run). Lower-band results are
obtained by bootstrap subsampling: $n$ candidates are drawn without
replacement from the XHigh pool, scored by the ORM, and the argmax
is selected. All four bands therefore share the same underlying
generations; only the pool size visible to the reward model changes.

\paragraph{Beam Search.}
Maintain $W$ candidate paths; at each step expand each beam by $k$
continuations, score all $W{\times}k$ by PRM, keep the top-$W$.
At each step every beam independently generates $k$ candidate
continuations, stopping at the step delimiter
\texttt{\textbackslash n\textbackslash n} or EOS. All $W{\times}k$ candidates are scored by the PRM on the full accumulated path, using the minimum per-step positive probability as the sequence-level score.
The top-$W$ remaining candidates are retained as the next generation of beams. A beam is marked complete when generation stops at EOS
rather than the delimiter; completed beams are frozen with their
current score. The final output is the completed beam with the highest
PRM score, or the highest-scoring active beam if no beam reached
completion.

\paragraph{Particle Filter.}
Maintain $N$ particles; at each step score all by PRM, stochastically
resample with replacement, then generate one continuation each.
The loop order follows SCORE $\to$ RESAMPLE $\to$ GENERATE. All $N$ particles are scored by the PRM on their full accumulated paths; $N$ replacement indices are then drawn from $\mathrm{softmax}(\mathbf{w}/T)$ with $T{=}1.0$ via multinomial sampling with replacement. Finished particles (EOS) are frozen: they participate in resampling with their terminal score and, if re-selected, are copied as-is. PF terminates when all N particles reach EOS or after all iterations is exhausted. The final output is the highest-scoring completed (frozen) particle.

\paragraph{Sequential Refinement.}
Iteratively improve a single draft via two sequential calls per
iteration: feedback, and refinement --- all using the
same generator.
We define a window of maximum previous iterations present in the context at $J{=}3$, which deviates from the
full-history formulation of ~\citet{madaan2023self} to keep the prompt
within context.
The thinking blocks are stripped from drafts before each call.
The feedback generation is run with thinking mode off with Qwen3.5.

\paragraph{Budget Forcing.}
Force the model to think longer by suppressing its end-of-thinking token and injecting a wait-string, then generate a final answer from the extended context.
Phase~1 generates within a shared token budget; each time the End of Thinking token (e.g. \texttt{</think>}) is emitted, it is stripped and an injection token from
\{\textit{"Wait", "Actually", "However", "Let me", "Also", "Well"}\} is appended until the budget is exhausted, the suppression limit is reached, or the model halts
naturally. The token is selected randomly. Phase~2 generates the answer freely from the accumulated
context. Evaluated at Low and Mid only; higher budgets caused degeneration into repetitive self-termination.

\paragraph{Fusion.}
Generate a pool of candidates independently, then synthesise a single
fused output in one call rather than selecting among them.
The candidate pool is shared with BoN; a random subset of size
$s \in \{3, 7, 15\}$ is drawn without replacement, shuffled, and
stripped of the thinking part before the synthesis call.

\paragraph{Reward models.} Both ORMs (Skywork-Reward-V2-Llama-3.1-8B and Llama-3.1-70B-Instruct-RM-RB2) score each (prompt, completion) pair in a single classification forward pass. Each call returns one scalar score per candidate.

\paragraph{Process reward model.}
VersaPRM-8B returns per-step scores in a single forward pass with STEP pooling. The input format follows VersaPRM's training specification: the prompt is concatenated to the candidate's steps, separated by a space followed by four newlines, with steps obtained client-side by splitting on the configured delimiter
\texttt{\textbackslash n\textbackslash n}. At each step boundary (tokenizer id~$23535$), the model emits a $[\text{neg}, \text{pos}]$ logit pair over the $+/-$ discriminator tokens (ids~$\{12, 10\}$) it was trained on; the positive probability is taken as the per-step score, and the sequence-level score reported to the search algorithm is the minimum across per-step positives, following the aggregation strategy reported as best-performing by~\citet{versaprm2025}.

\paragraph{Sequential Refinement prompts.}
The initial draft ($y_0$) is generated with no custom template, using only the dataset prompt and the system prompt above. Each subsequent iteration consists of three generator calls in fixed order: (i)~a \textit{feedback} call that critiques the current draft; (ii)~a \textit{stop-check} call returning STOP or CONTINUE (computed
every iteration but ignored under our forced-iteration setup); and
(iii)~a \textit{refinement} call that produces the next draft from the
task input and the recent (draft, feedback) history. The three
templates are reproduced below verbatim, with placeholders:
\texttt{\{x\}} is the original task input,
\texttt{\{y\_t\}} the current draft, \texttt{\{fb\_t\}} the feedback on
that draft, \texttt{\{few\_shot\_examples\}} an optional few-shot prefix
(left empty in all our runs), and \texttt{\{history\}} the rendered
sequence of prior drafts and feedbacks in the format:
\texttt{\#\# Draft 0\textbackslash n\{y\_0\}\textbackslash n\textbackslash
n\#\# Feedback 0\textbackslash n\{fb\_0\}\textbackslash n\textbackslash
n\#\# Draft 1\textbackslash n\ldots} (capped at the most recent $K{=}3$
pairs).

\begin{promptbox}{Feedback prompt (step i)}
\begin{verbatim}
{few_shot_examples}### TASK INPUT
{x}

### CURRENT DRAFT
{y_t}

### FEEDBACK
Identify specific issues in the draft and propose concrete improvements
as bullet points:
\end{verbatim}
\end{promptbox}

\begin{promptbox}{Stop-check prompt (step ii)}
\begin{verbatim}
{few_shot_examples}### TASK INPUT
{x}

### CURRENT DRAFT
{y_t}

### FEEDBACK ON THIS DRAFT
{fb_t}

### DECISION
Based on the feedback above:
- Answer STOP if the draft does not need further refinement.
- Answer CONTINUE if the feedback raises significant issues worth
  addressing.

You MUST output exactly one of the following two lines and nothing else:
<decision>STOP</decision>
<decision>CONTINUE</decision>
\end{verbatim}
\end{promptbox}

\begin{promptbox}{Refinement prompt (step iii)}
\begin{verbatim}
{few_shot_examples}### TASK INPUT
{x}

### REVISION HISTORY
{history}

### INSTRUCTIONS
Revise the most recent draft by applying the most recent feedback.
- Fix every issue mentioned in the feedback.
- Preserve all task requirements and constraints.
- Do not introduce new unrelated content or regressions.
- Output ONLY the refined draft (no explanations, no scores, no
  feedback).

### REFINED OUTPUT
\end{verbatim}
\end{promptbox}

\paragraph{Fusion prompt.}
Fusion uses the direct-synthesis variant (a two-step compare-then-fuse
variant exists in the implementation but is not used in this paper).
\texttt{\{generations\}} is the candidate pool rendered as
\texttt{\#\# Generation 1}, \texttt{\#\# Generation 2}, \ldots,
and \texttt{\{instruction\}} is the original task input.

\begin{promptbox}{Fusion prompt (direct synthesis)}
\begin{verbatim}
Based on the provided Task Input and Generated Texts, fuse them into a
better generation that combines the strength of each of them. The fused
generation should adequately respond to the task input, sound natural to
a native speaker, and be focused on conveying the most relevant and
accurate information in a responsible and ethical way.

### Generated Texts
{generations}

### Task Input
{instruction}

Output only the fused generation text, as if you were directly answering
the task yourself. Do not reference or mention the previous generations
(e.g., avoid phrases like "combining the best of both responses" or "as
mentioned in Generation 1").

Please provide your fused text.
\end{verbatim}
\end{promptbox}

\paragraph{Other methods.}
BoN, Beam Search, Particle Filter, and Budget Forcing use only each
benchmark's native prompt together with the system prompt above; no
custom user template is added.

\paragraph{LLM-Judge input tokens.}
Table~\ref{tab:judge-input-tokens} reports the total input tokens
sent to the LLM-as-Judge across all the judge runs in our evaluation, aggregated
by algorithm family $\times$ benchmark. Grand total: 27{,}540\,M
tokens ($\approx 27.5$\,B).

\begin{table}[ht]
\centering
\small
\caption{Total LLM-Judge input tokens (millions) by algorithm family and benchmark.}
\label{tab:judge-input-tokens}
\resizebox{\textwidth}{!}{
\begin{tabular}{lrrrrrr}
\toprule
Algorithm & HealthBench & LEXam & PRBench & WildBench & WritingBench & Total \\
\midrule
BoN (oracle, $n=16$) & 4,556 & 202 & 4,995 & 173 & 506 & 10,432 \\
PF-High (oracle, $n=16$) & 1,519 & 74 & 1,704 & 54 & 176 & 3,527 \\
Fusion & 1,524 & 61 & 2,219 & 48 & 164 & 4,015 \\
Beam Search & 1,146 & 44 & 1,657 & 37 & 120 & 3,004 \\
PF (normal) & 1,146 & 44 & 1,657 & 37 & 120 & 3,004 \\
Budget Forcing & 764 & 29 & 1,105 & 25 & 80 & 2,003 \\
Sequential Refinement & 638 & 25 & 814 & 16 & 61 & 1,555 \\
\midrule
\textbf{Total} & 11,293 & 479 & 14,151 & 390 & 1,227 & 27,540 \\
\bottomrule
\end{tabular}}
\end{table}

\paragraph{Combined paper-level compute totals.}
Total compute across all experiments was approximately $23{,}800$ GPU-hours on $8\times$B200 nodes Table~\ref{tab:gpu-hours}

\begin{table}[ht]
\centering
\small
\caption{Combined paper-level totals across Open-QA and verifiable evaluation tracks. Wall-clock hours measure end-to-end run duration; GPU-hours scale by node size.}
\label{tab:gpu-hours}
\begin{tabular}{lrrr}
\toprule
            & Tasks & Wall-clock (h) & GPU-hours \\
\midrule
Open-QA     &   164 &        2{,}642.6 & 21{,}104 \\
Verifiable  &    93 &          342.2 &  2{,}737 \\
\midrule
\textbf{Total} &  257 &        2{,}984.8 & $\approx$23{,}800 \\
\bottomrule
\end{tabular}
\end{table}

\section{Judge Agreement Analysis}
\label{app:judge_agreement}

\paragraph{Why a unified judge.}
Running the native judges (Table~\ref{tab:native_judges},
Sec~\ref{app:benchmarks}) across the full evaluation grid is
prohibitive: per-criterion designs (HealthBench, PRBench,
WritingBench) fan out into hundreds of thousands of GPT-4.1,
o4-mini, and Claude 3.7 Sonnet calls, and consequently prohibitive cost. We therefore use
\textbf{Qwen3.5-397B-A17B}, as the unified judge for all main-text results,
and validate against each benchmark's native evaluation on a
stratified sample.

\paragraph{Unified setup.}
All judge models receive each native prompt verbatim and evaluate the same
candidate completions (with reasoning tags stripped), so the judge model is
the sole variable across comparisons. All runs share the following sampling
configuration:

\begin{promptbox}{Qwen3.5 Judge — Sampling Configuration}
\begin{verbatim}
sampling:
  temperature: 0.7
  top_p:        0.8
  top_k:        20
  min_p:        0
  chat_template_kwargs:
    enable_thinking: false
\end{verbatim}
\end{promptbox}

Full config can be shared upon request.

\paragraph{Sample selection.}
We draw a stratified random sample over task splits per benchmark,
preserving sub-population proportions. The sampling rate is $15\%$
of items for LEXam, WildBench, and WritingBench, and $5\%$ for
PRBench (reduced due to its high per-item criterion count,
averaging ${\sim}17.7$ criteria per item). Both judges score every
completion on identical inputs. For benchmarks with per-criterion
evaluation (PRBench, WritingBench), the pair counts in
Table~\ref{tab:judge_agreement} reflect the total number of
criterion-level judgments (sampled items $\times$ average criteria
per item); for LEXam and WildBench, which produce a single holistic
score per item, the pair count equals the number of sampled items
directly.

\paragraph{HealthBench: validation against physician annotations.}
HealthBench is the only benchmark in our suite that had public human expert
annotations, enabling a stronger validation than unified-vs-native
comparison. We evaluate our unified judge against the physician
rubric annotations released with the benchmark. Table~\ref{tab:healthbench_judges}
reports Macro~F1 for several judge models on this task.
Qwen3.5-397B-A17B achieves a Macro~F1 of $0.679$, comparable to
GPT-4.1 ($0.709$) and o4-mini ($0.692$). The
inter-physician pairwise agreement across the 184 physicians with
sufficient annotation data has a mean Macro~F1 of $0.655$
($\text{std} = 0.102$, range $[0.04, 1.0]$). Our unified judge
sits at the 57th percentile of this distribution, meaning it agrees
with physicians better than the majority of individual physicians
agree with each other.

\begin{table}[H]
\centering
\small
\caption{HealthBench judge agreement with physician annotations
(Macro~F1). The inter-physician baseline is the mean pairwise
Macro~F1 across 184 physicians with sufficient annotation data.}
\label{tab:healthbench_judges}
\begin{tabular}{lc}
\toprule
\textbf{Judge} & \textbf{Macro F1} \\
\midrule
GPT-4.1              & 0.709 \\
o4-mini              & 0.692 \\
o3                   & 0.681 \\
\textbf{Qwen3.5-397B-A17B (ours)} & \textbf{0.679} \\
GPT-4.1 mini         & 0.661 \\
\midrule
Inter-physician mean  & 0.655 ($\pm\,0.102$) \\
\bottomrule
\end{tabular}
\end{table}

\paragraph{Other benchmarks: unified vs.\ native judge.}
For the remaining four benchmarks, we compare the unified against
each benchmark's native judge (Table~\ref{tab:native_judges}).
Table~\ref{tab:judge_agreement} reports the agreement.

\begin{table}[H]
\centering
\small
\caption{Criterion-level agreement between the unified
(Qwen3.5-397B-A17B) and each benchmark's native judge. PRBench is
evaluated on a $5\%$ sample; the remaining benchmarks on a $15\%$
sample. Metrics are reported in the form most natural to each
native protocol and are not directly comparable across rows.}
\label{tab:judge_agreement}
\begin{tabular}{llll}
\toprule
\textbf{Benchmark} & \textbf{$n$ pairs} & \textbf{Metric} &
\textbf{Agreement} \\
\midrule
PRBench      & 1{,}186 & $\kappa$ / Macro F1 & 0.679 / 0.838 \\
LEXam        & 77      & Pearson $r$ / MAE    & 0.813 / 22.2  \\
WildBench    & 147     & QWK / MAE            & 0.564 / 1.35  \\
WritingBench & 404     & QWK / MAE            & 0.408 / 1.51  \\
\bottomrule
\end{tabular}
\end{table}

PRBench and LEXam show the strongest alignment. WildBench and WritingBench show lower per-pair agreement (QWK = 0.564 and 0.408), driven by the heterogeneity of evaluation criteria across task types and domains.



\section{Theory}
\label{app:theory}

\paragraph{Self-verifier methods (Fusion, Sequential Refinement).}
\label{app:self-verifier-bounds}
These methods use the generator itself to produce the final output, either
by synthesising across candidates (Fusion) or by iteratively critiquing and
rewriting (Sequential Refinement). Their headroom capture does not depend on an
external verifier's correlation with true quality. Instead, by the data
processing inequality, the mutual information between the self-verifier's
signal and true quality is bounded by the generator's own distributional
quality:
\begin{equation}
    I\!\left(\mathcal{V}_{=\theta}(y);\;Q_{\mathrm{true}}(y)\right)
    \;\leq\;
    I\!\left(\pi_\theta(y \mid x);\;Q_{\mathrm{true}}(y)\right).
    \label{app:dpi}
\end{equation}
A more capable generator has higher mutual information with true quality,
raising the ceiling on headroom capture for self-verifier methods.
Critically, this bound is on \emph{capability} (the ability to produce and
assess high-quality outputs), not on parameter count alone.

\paragraph{Verifier correlation under noisy scoring.}
\label{app:bon-noisy}
When the external verifier correlates imperfectly with true quality at rate
$\rho_v = \mathrm{Corr}[\mathcal{V}(y), \mathcal{V}^*(y)]$, the expected
true quality of the BoN-selected candidate satisfies:
\begin{equation}
    Q^{\mathrm{BoN}}(N, \rho_v)
    \;\approx\;
    \mu \;+\; \rho_v \cdot \sigma \cdot
    \Phi^{-1}\!\!\left(\frac{N}{N+1}\right),
    \label{eq:bon_noisy}
\end{equation}

Substituting into the headroom capture Equation~\ref{eq:headroom} with $Q^{*}(N) \approx \mu + \sigma\,\Phi^{-1}(N/(N+1))$
from Equation~\ref{eq:order_stat}:
\begin{equation}
    h^{\mathrm{BoN}} \;=\;
    \frac{Q^{\mathrm{BoN}} - \mu}{Q^{*} - \mu} \;=\;
    \frac{\rho_v \cdot \sigma \cdot
    \Phi^{-1}\!\!\left(\frac{N}{N+1}\right)}
    {\sigma \cdot \Phi^{-1}\!\!\left(\frac{N}{N+1}\right)}
    \;=\; \rho_v.
\end{equation}

Headroom capture for BoN \emph{is} the verifier correlation. When
$\rho_v = 1$ the method captures all available quality; when $\rho_v = 0$
additional compute yields no benefit; when $\rho_v < 0$ more compute
actively harms performance.

\paragraph{Tree search pruning sensitivity.}
\label{app:tree-pruning}
Tree search is exponentially more sensitive to verifier miscalibration than
BoN. Let $\epsilon_{\mathrm{prune}}$ denote the per-step probability that
the PRM incorrectly prunes the highest-quality surviving candidate. The
probability that the best candidate survives to depth $d$ decays as:
\begin{equation}
    P(\text{best survives to depth } d)
    \;=\; (1 - \epsilon_{\mathrm{prune}})^d.
    \label{eq:pruning}
\end{equation}
At $\epsilon_{\mathrm{prune}} = 0.33$ and $d = 4$, the best candidate
survives with probability $\leq 0.20$, meaning tree search is expected to
return a suboptimal candidate more than 80\% of the time even under
moderate miscalibration. This formalises why Beam Search and Particle
Filter perform strictly worse than naive parallel sampling on open-ended
QA, where PRM miscalibration is severe (Section~\ref{sec:tree_failure}).

\section{Oracle Bias From Judge Noise}
\label{app:oracle_bias}
       
  \textbf{Setup.} The judge here is the \emph{final reference-based evaluation} --- it scores candidates after the algorithm has already chosen what to do, and is not part of the algorithm itself. We start by studying the  the BoN ceiling (``oracle'') due to its simplicity of having independent candidates. Each of the $n$ candidates are passed through the judge and then the max is taken. \textbf{Judge noise inflates that max upward.}                                                      
  For a given input, candidate $i \in \{1, \ldots, n\}$ has true mean score $\mu_i$ (infinite-trial judge average). The judge realizes $X_i = \mu_i + \epsilon_i$ with $\epsilon_i                       
  \stackrel{\text{i.i.d.}}{\sim} \mathcal{N}(0, \sigma_J^2)$.                                 
  \textbf{Notation.} Let $Z_1, \ldots, Z_n \stackrel{\text{i.i.d.}}{\sim} \mathcal{N}(0, 1)$ be a generic set of i.i.d. standard normals (not the candidate scores --- introduced solely to define one constant),   
  and
  $$a_n := \mathbb{E}\bigl[\max_{i \in \{1, \ldots, n\}} Z_i\bigr],$$                                     
  the expected maximum of $n$ i.i.d. standard normals. Numerically: $a_2 \approx 0.564$, $a_4 \approx 1.029$, $a_8 \approx 1.424$, $a_{16} \approx 1.766$, growing like $\sqrt{2 \ln n}$.
  Two quantities of interest:                         
  \begin{enumerate}                                   
      \item \textbf{Naive oracle} $O^{\text{naive}} = \max_i X_i$ --- \textbf{overstates} the BoN ceiling.                                                      
      \item \textbf{Ideal oracle} $O^{\text{ideal}} = \max_i \mu_i$ --- true quality of the genuinely best candidate, i.e.\ what an infinite-trial judge would report.                                       
  \end{enumerate}                                                                                      
  \subsection{Derivation of the expectations}
  \label{app:idealoraclederivation}
  Two facts used repeatedly:                       
  \begin{itemize}                                        
      \item \textbf{(F1) Sum of independent normals.} If $A \sim \mathcal{N}(\alpha, \sigma_A^2)$ and $B \sim \mathcal{N}(\beta, \sigma_B^2)$ are independent, then $A + B \sim \mathcal{N}(\alpha + \beta,
  \sigma_A^2 + \sigma_B^2)$.                                                                             
      \item \textbf{(F2) Affine equivariance of the max.} For constants $a$ and $b > 0$ and any random variables $V_1, \ldots, V_n$, $\max_i(a + b V_i) = a + b\max_i V_i$, so $\mathbb{E}[\max_i(a + b V_i)]
  = a + b\,\mathbb{E}[\max_i V_i]$.                     
  \end{itemize}                                           
  \textbf{Ideal.} Assume the distribution of true scores from our algorithm follows $\mu_i \stackrel{\text{i.i.d.}}{\sim} \mathcal{N}(\bar\mu, \tau^2)$, independent of the $\epsilon_i$, so we can write $\mu_i = \bar\mu + \tau Z_i$ with $Z_i \stackrel{\text{i.i.d.}}{\sim} \mathcal{N}(0,
   1)$. By (F2),
  $$\mathbb{E}[O^{\text{ideal}}] = \mathbb{E}[\max_i \mu_i] = \mathbb{E}[\max_i(\bar\mu + \tau Z_i)] = \bar\mu + \tau\,\mathbb{E}[\max_i Z_i] = \boxed{\bar\mu + a_n\,\tau}.$$                     
  \textbf{Naive.} Apply (F1) to $X_i = \mu_i + \epsilon_i$: since $\mu_i$ and $\epsilon_i$ are independent and Gaussian,                                    
  $$X_i \sim \mathcal{N}(\bar\mu, \tau^2 + \sigma_J^2),$$                                                   
  and the $X_i$'s are i.i.d. (the pairs $(\mu_i, \epsilon_i)$ are mutually independent, so the sums are too). Standardize: $X_i = \bar\mu + \sqrt{\tau^2 + \sigma_J^2}\,Z_i'$ with $Z_i'                         
  \stackrel{\text{i.i.d.}}{\sim} \mathcal{N}(0, 1)$. By (F2),                                             
  $$\mathbb{E}[O^{\text{naive}}] = \mathbb{E}[\max_i X_i] = \bar\mu + \sqrt{\tau^2 + \sigma_J^2}\,\mathbb{E}[\max_i Z_i'] = \boxed{\bar\mu + a_n\sqrt{\tau^2 + \sigma_J^2}}.$$                                
         
  \textbf{Gap.} Subtracting,  $$\mathbb{E}[O^{\text{naive}}] - \mathbb{E}[O^{\text{ideal}}] = a_n\bigl[\sqrt{\tau^2 + \sigma_J^2} - \tau\bigr] = a_n \cdot \frac{\sigma_J^2}{\sqrt{\tau^2 + \sigma_J^2} + \tau}$$                         
  Can see that: $\sigma_J \to 0 \Rightarrow$ gap $\to 0$; $\tau = 0 \Rightarrow$ gap $= a_n \sigma_J$ (no true spread, all the apparent winner's lead is   
  noise); $\tau \to \infty \Rightarrow$ gap $\to 0$ (a real quality gap drowns out the noise); grows with $n$ via $a_n$.                                                                             
  \subsection*{Ideal per-entry estimator}                
  The correction is just the closed-form bias evaluated at the per-entry candidate spread:
  $$\hat O^{\text{ideal}}_{\text{entry}} = \max_i X_i - a_n\bigl[\sqrt{\hat\tau^2 + \sigma_J^2} - \hat\tau\bigr], \qquad \hat\tau^2 = \max\bigl(0,\; s_X^2 - \sigma_J^2\bigr),$$
  where $s_X^2$ is the sample variance of the $n$ candidate scores in this entry (so $\mathbb{E}[s_X^2] = \tau^2 + \sigma_J^2$ and $\hat\tau^2$ is a plug-in for $\tau^2$). Under the Gaussian model with     
  oracle $\tau$, this is unbiased for $\mathbb{E}[O^{\text{ideal}}]$. With plug-in $\hat\tau^2$ the estimator has small finite-sample bias driven by the clip-at-zero (which biases $\hat\tau$ upward when    
  true $\tau$ is small, hence biases the correction \emph{downward} --- i.e.\ corrects too little when $\tau$ is near 0; conservative).    

\paragraph{Approximated per-benchmark correction}
We sampled 15-20\% of prompts per benchmark with a minimum of 100, then generated a single response from Qwen3.5-35B for each, and ran the judge four times per generation to produce a within-entry variance estimate. We make two simplifying assumptions: (i) judge variance is constant across entries within a benchmark and (ii) constant across the models we evaluate. Under (i)--(ii) a single per-benchmark $\sigma_J^2$ suffices for every algorithm and model. The resulting estimates are     
  listed in Table~\ref{tab:judge-std-per-benchmark}.
  
\begin{table}[h]
  \centering
  \begin{tabular}{lr}                                           
  \toprule
  Benchmark & Square root of mean var (normalized) \\                         
  \midrule                                                      
  HealthBench & 0.076 \\
  PRBench & 0.034 \\                                           
  LEXam & 0.055 \\                                           
  WildBench & 0.055 \\
  WritingBench & 0.030 \\                                      
  \bottomrule          
  \end{tabular}
  \caption{Square root of the per-benchmark mean per-entry judge variance (micro-averaged across entries).}                            
  \label{tab:judge-std-per-benchmark}                           
  \end{table}     

\begin{figure}[H]
    \centering
    \includegraphics[width=\textwidth]{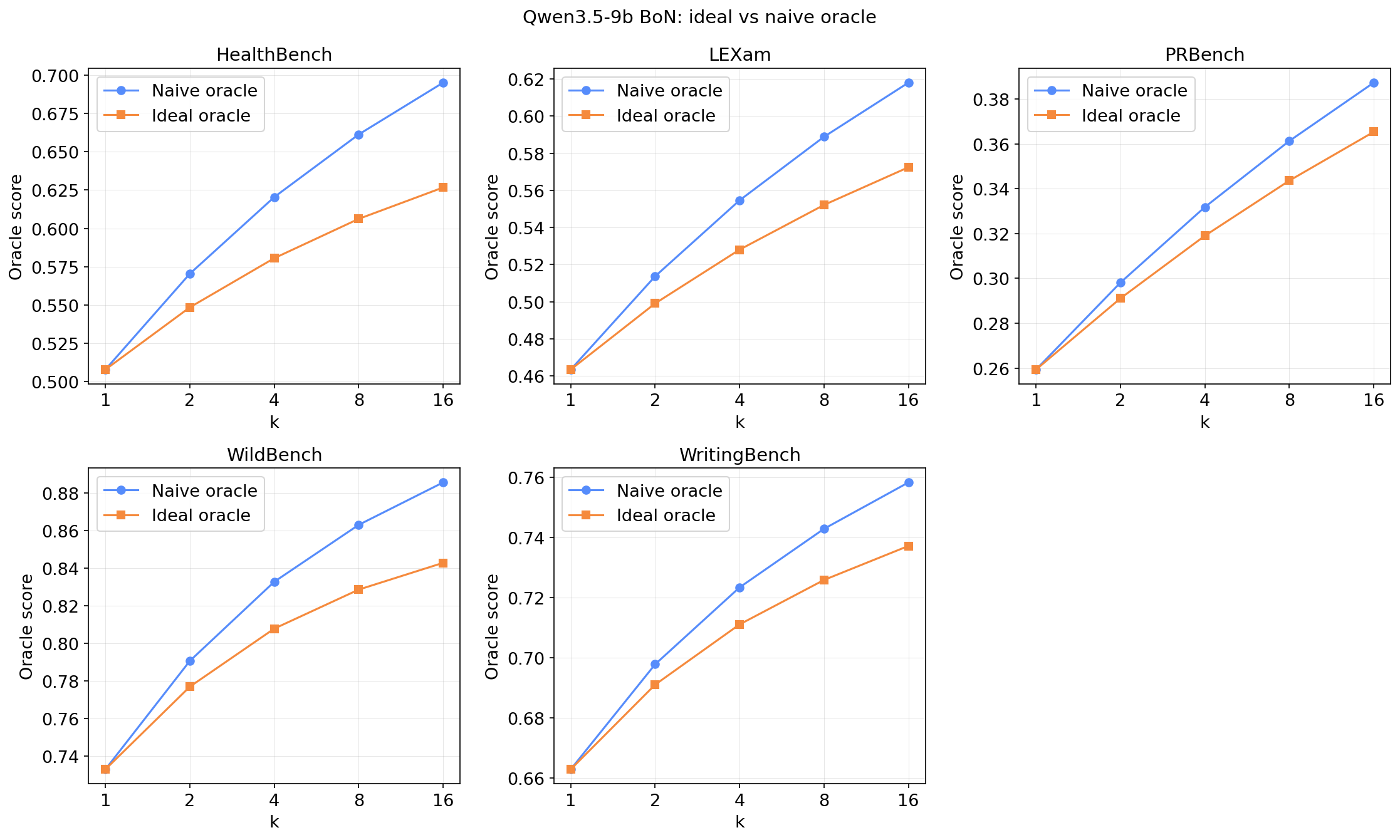}
    \caption{Naive (unadjusted) vs ideal oracle scores across compute for qwen3.5-9b BoN.}
    \label{fig:bon_ideal_vs_naive_adjustment}
\end{figure}

With $\sigma_J^2$ in hand, the per-entry estimator $\hat O^{\text{ideal}}_{\text{entry}}$ defined in Section~\ref{app:idealoraclederivation} can be evaluated directly: $a_n$ is known, $\sigma_J^2$ is read from Table~\ref{tab:judge-std-per-benchmark}, and $s_X^2$ is the empirical variance of the candidate scores within the entry. 
For BoN this is unambiguous --- the $n$ independent generations are the candidate pool by construction  
  --- and Figure~\ref{fig:bon_ideal_vs_naive_adjustment} shows the resulting adjustment across compute levels. 

The other algorithms in our study do not contain i.i.d. candidate pools, so applying the same correction requires algorithm-specific choices. For simplicity, we assume the candidate pools of Refinement, Fusion, and Particle Filtering are all i.i.d. and then discuss the limitations in the next section. The candidate pools are defined as the following for compute C (assume corresponds to BoN@k):
\begin{itemize}
    \item \textbf{Refinement.} The drafts in a trajectory that uses amount of compute C. 
    \item \textbf{Fusion.} The set of k-1 i.i.d. samples along with the fused response.
    \item \textbf{Particle Filter. } The pool is the final k particles.
\end{itemize}                                                   

\paragraph{Limitations of the i.i.d. assumption.}                      
  The closed-form bias is exact only when the $n$ candidates are i.i.d. Gaussian. Below we work through the qualitative behaviour we expect from
   the i.i.d. plug-in correction under each algorithm's structural deviation from this assumption; the magnitude and even the direction of the residual error depend on the underlying generator's behaviour and should be read as
  hypotheses rather than guarantees.                                                                 
  \begin{itemize}                                                                    
    \item \textbf{Fusion.} When fusion works well, the fused response has a higher mean than the i.i.d. input samples. In the limit as the fused-mean minus input-mean grows, the naive oracle picks the fused sample with probability     
  approaching one, judge noise on that single chosen candidate averages out, and the true bias of the naive oracle approaches zero. Our estimator, however, plugs in $s_X^2$ from the (fused, inputs) pool: the fused-vs-input gap
  inflates $s_X^2$ as if it were within-pool spread, and the formula continues to subtract a positive correction. The reported ideal oracle is therefore expected to be a slight \emph{lower bound} when the model is strong at fusing responses.                                           
  \item \textbf{Refinement.} Refinements within a trajectory are expected to be correlated. A scenario of interest is monotonic improvement, where the last
  refinement consistently has the highest mean and the naive oracle picks it with probability approaching one --- the same situation as Fusion, with the last refinement playing the role of the dominating fused response. The       
  fused-vs-input mean gap is replaced by the first-vs-last drift across the trajectory, $s_X^2$ is inflated by that drift, and the formula continues to subtract a positive correction even though the true bias has approached zero.
  Assuming strong refinement capabilities, we'd expect the ideal oracle to be a slight \emph{lower bound}. 
  
    \item \textbf{Particle Filter.} Resampling concentrates particles into clusters of near-identical trajectories that share a true mean within each cluster. A scenario of interest is when resampling produces a few well-separated
  clusters of size $c$: the naive oracle picks the maximum within the best-mean cluster, which is inflated above the true ideal by the maximum of $c$ judge-noise terms ($\approx a_c\sigma_J$). Our estimator's $s_X^2$, however, is
  dominated by the between-cluster mean gaps and inflates $\hat\tau$ well above the within-cluster noise scale, so the formula subtracts a much smaller correction ($\sim a_n\sigma_J^2/(2\hat\tau)$) than the true within-cluster    
  max-of-noise inflation. Therefore, we expect the ideal oracle to be a slight \emph{upper bound}, opposite in direction from Fusion and Refinement.                                                                                               
  \end{itemize}                                           

\subsection[Consistency of h under judge noise]{Consistency of $h \approx \rho_v$ under judge noise}
\label{app:sec:rho_bias}
Looking at Eq.~\ref{eq:headroom_rho}
, judge noise affects both sides of the identity $h \approx \rho_v$ 
symmetrically, so the linear relationship is preserved even without 
the oracle correction. To see this, note that the judge scores 
$J(y) = \mathcal{V}^*(y) + \epsilon_J$ with 
$\epsilon_J \sim \mathcal{N}(0, \sigma_J^2)$ enter two quantities:

\paragraph{Effect on measured headroom capture $\hat{h}$.}
The numerator $\hat{Q}(T) - \mu$ is the expected judge score of the 
RM-selected candidate minus baseline. Since RM selection is based on 
$\mathcal{V}(y)$ and judge noise $\epsilon_J$ is independent of 
$\mathcal{V}$, the noise averages out in expectation:
$$
\hat{Q}(T) - \mu \;\approx\; \rho_v \cdot a_N \tau.
$$
The denominator $\hat{Q}^*(T) - \mu$ is the expected maximum of $N$ 
i.i.d. judge scores, which by the derivation in 
Section~\ref{app:idealoraclederivation} equals $a_N\sqrt{\tau^2 + \sigma_J^2}$. 
Therefore:
$$
\hat{h} \;=\; \frac{\hat{Q}(T) - \mu}{\hat{Q}^*(T) - \mu} 
\;\approx\; \rho_v \cdot \frac{\tau}{\sqrt{\tau^2 + \sigma_J^2}}.
$$

\subsection*{Effect on measured verifier correlation $\hat{\rho}_v$.}
We measure $\hat{\rho}_v = \mathrm{Corr}[\mathcal{V}(y),\, J(y)]$, 
whereas the theoretical $\rho_v = \mathrm{Corr}[\mathcal{V}(y),\, 
\mathcal{V}^*(y)]$. Since $J(y) = \mathcal{V}^*(y) + \epsilon_J$ with 
$\epsilon_J$ independent of $\mathcal{V}$, standard attenuation gives:
$$
\hat{\rho}_v 
\;=\; \rho_v \cdot \sqrt{\frac{\tau^2}{\tau^2 + \sigma_J^2}}
\;=\; \rho_v \cdot \frac{\tau}{\sqrt{\tau^2 + \sigma_J^2}}.
$$

\paragraph{Cancellation.}
Both quantities are attenuated by the same factor 
$\tau/\sqrt{\tau^2 + \sigma_J^2}$, so:
$$
\hat{h} \;\approx\; \hat{\rho}_v,
$$
and the identity $h \approx \rho_v$ holds at the level of 
\emph{measured} quantities without any correction. The oracle 
correction in Section~\ref{app:idealoraclederivation} is therefore not required 
for the validity of Equation~\ref{eq:headroom_rho}; it serves the 
separate purpose of obtaining unbiased estimates of true oracle quality 
$\mathbb{E}[O^{\mathrm{ideal}}]$.

\subsection*{Heterogeneous attenuation across benchmarks.}
The attenuation factor $\tau/\sqrt{\tau^2 + \sigma_J^2}$ varies across 
benchmarks because both $\tau$ (true spread of candidate quality) and 
$\sigma_J$ (judge noise, Table~\ref{tab:judge-std-per-benchmark}) 
differ. This introduces benchmark-level offsets that appear as scatter 
around the identity line in Figure~\ref{fig:correlation_and_exploitation} 
(left), but do not bias the slope in expectation. The observed slope of $1.198$ and $R^2 = 0.656$ are consistent with this prediction.



\section{Naive Oracle Results}
\label{app:naive_oracle}
\begin{figure}[H]
    \centering
    \includegraphics[width=\textwidth]{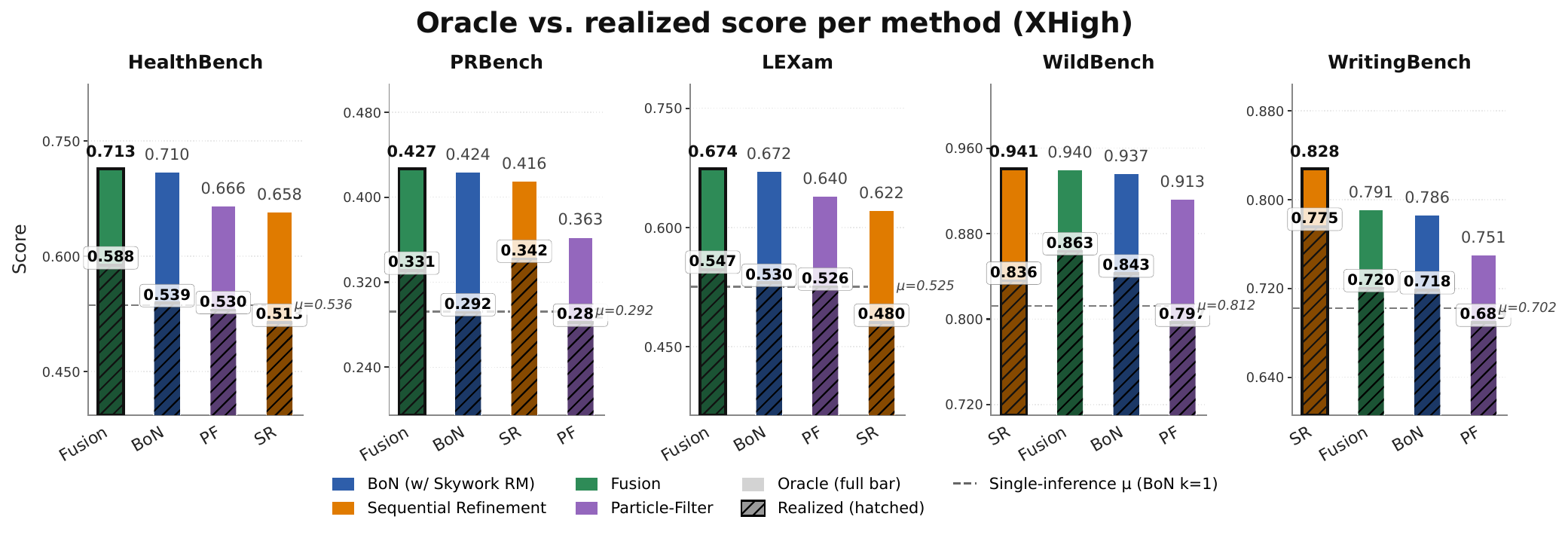}
    \caption{Oracle (full bar) vs.\ realised (hatched) quality per method
    at matched compute (\textbf{XHigh}), Qwen3.5-35B-A3B, using the
    \textbf{naive} (uncorrected) oracle.}
    \label{fig:oracle_quality_naive}
\end{figure}

\begin{figure}[H]
    \centering
    \includegraphics[width=\textwidth]{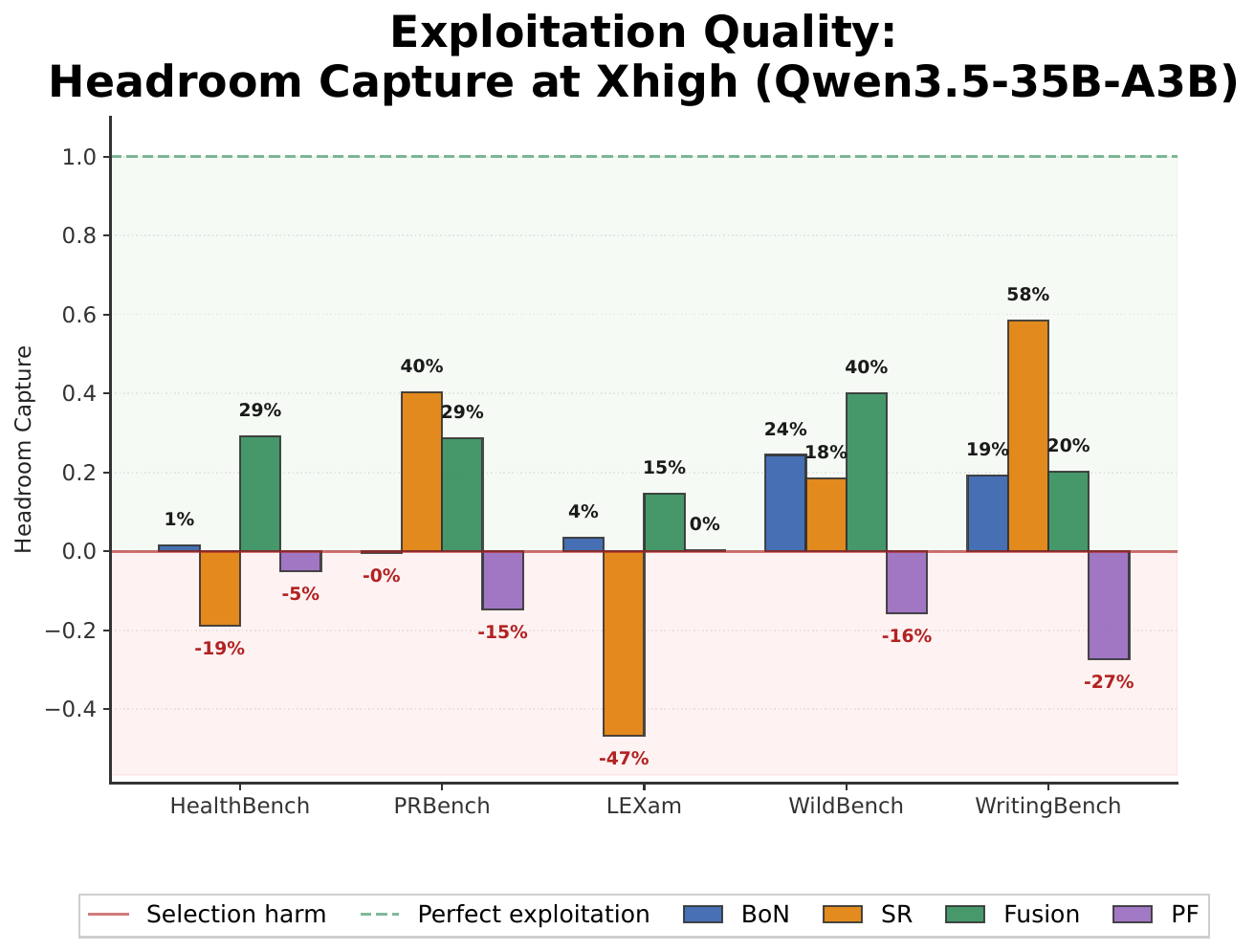}
    \caption{Headroom capture per method at XHigh compute, using the
    \textbf{naive} (uncorrected) oracle. Compare with
    Figure~\ref{fig:correlation_and_exploitation} (right) which uses
    the bias-corrected oracle.}
    \label{fig:exploitation_naive}
\end{figure}

\begin{figure}[H]
      \centering
      \includegraphics[width=0.65\textwidth]{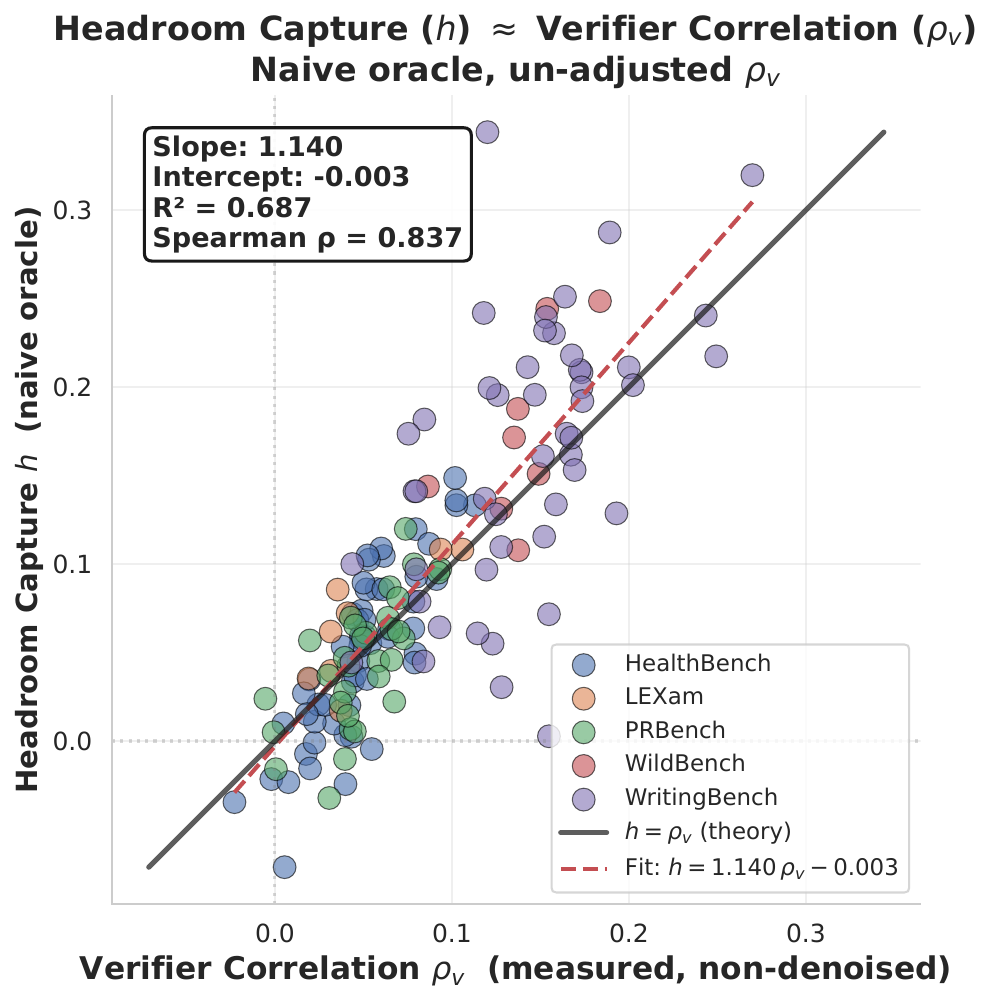}
      \caption{$h$ vs.\ $\rho_v$ across $N=152$ task-level points using the \textbf{naive}
      (uncorrected) oracle and \textbf{measured} (non-denoised) $\rho_v$. The empirical fit
      closely tracks the identity line $y=x$, mirroring the linear relationship
      $\hat{h}\approx\hat{\rho}_v$ that Figure~\ref{fig:correlation_and_exploitation} (left)
      demonstrates under the bias-corrected oracle and denoised $\hat{\rho}_v$. Same
      conclusion holds under both conventions: headroom capture is bounded by verifier
      correlation regardless of whether we de-noise the oracle and the correlation estimate.}
      \label{fig:headroom_scatter_naive}
  \end{figure}

Figures~\ref{fig:oracle_quality_naive}, \ref{fig:exploitation_naive}, and~\ref{fig:headroom_scatter_naive} replicate Figures~\ref{fig:oracle_quality} and~\ref{fig:correlation_and_exploitation} using the naive oracle $\hat{O}^{\mathrm{naive}} = \max_i J(y_i)$, which overstates true pool quality by selecting partly on judge noise. Despite the inflated ceiling, the conclusion is unchanged: oracle quality rises with compute while realised quality stagnates, and exploitation remains the binding constraint. The naive oracle inflates the apparent gap, making the exploitation failure look more severe, not less.

    \subsection{Headroom capture tables (Unadjusted oracles)}
    \label{app:headroom_tables_unadjusted}
    
\begin{table}[H]
\centering
\setlength{\tabcolsep}{2.5pt}
\tiny
\caption{Headroom capture for \textbf{Qwen3.5-35B}: $h = (\text{realized} - \mu) / (\text{oracle}_\text{ideal} - \mu)$, where $\mu$ is the BoN $N{=}1$ baseline (single-inference mean). $h = 1$ means full exploitation of the oracle gap; $h = 0$ matches $\mu$; $h < 0$ underperforms $\mu$. \colorbox[RGB]{56,142,60}{\textcolor{white}{green}} = positive (gain captured), \colorbox[RGB]{200,60,60}{\textcolor{white}{red}} = negative (below baseline). Bold when $|h| \geq 0.20$.}
\label{tab:headroom_naive_qwen35_35b_unadjusted}
\resizebox{\textwidth}{!}{%
\begin{tabular}{l cccc cccc cccc}
  \toprule
  & \multicolumn{4}{c}{\textbf{HealthBench}} & \multicolumn{4}{c}{\textbf{PRBench}} & \multicolumn{4}{c}{\textbf{LEXam}} \\
  \cmidrule(lr){2-5} \cmidrule(lr){6-9} \cmidrule(lr){10-13}
  & \footnotesize Low & \footnotesize Mid & \footnotesize High & \footnotesize XHigh & \footnotesize Low & \footnotesize Mid & \footnotesize High & \footnotesize XHigh & \footnotesize Low & \footnotesize Mid & \footnotesize High & \footnotesize XHigh \\
  \midrule
  BoN+Skywork
    & \cellcolor[RGB]{239,246,239}+0.047
    & \cellcolor[RGB]{242,247,242}+0.038
    & \cellcolor[RGB]{244,249,245}+0.030
    & \cellcolor[RGB]{246,250,246}+0.025
    & \cellcolor[RGB]{242,247,242}+0.038
    & \cellcolor[RGB]{247,250,247}+0.023
    & \cellcolor[RGB]{250,252,250}+0.014
    & \cellcolor[RGB]{254,253,253}-0.005
    & \cellcolor[RGB]{239,246,239}+0.047
    & \cellcolor[RGB]{234,243,234}+0.062
    & \cellcolor[RGB]{233,242,234}+0.065
    & \cellcolor[RGB]{237,245,238}+0.051 \\
  Fusion
    & \cellcolor{white}---
    & \cellcolor[RGB]{56,142,60}\textbf{+0.670}
    & \cellcolor[RGB]{62,145,66}\textbf{+0.580}
    & \cellcolor[RGB]{90,161,94}\textbf{+0.495}
    & \cellcolor{white}---
    & \cellcolor[RGB]{125,181,128}\textbf{+0.390}
    & \cellcolor[RGB]{136,187,138}\textbf{+0.357}
    & \cellcolor[RGB]{142,191,144}\textbf{+0.339}
    & \cellcolor{white}---
    & \cellcolor[RGB]{137,188,139}\textbf{+0.355}
    & \cellcolor[RGB]{173,208,175}\textbf{+0.246}
    & \cellcolor[RGB]{185,215,186}\textbf{+0.209} \\
  Particle Filter
    & \cellcolor{white}---
    & \cellcolor{white}---
    & \cellcolor{white}---
    & \cellcolor[RGB]{239,200,200}-0.168
    & \cellcolor{white}---
    & \cellcolor{white}---
    & \cellcolor{white}---
    & \cellcolor[RGB]{228,159,159}\textbf{-0.293}
    & \cellcolor{white}---
    & \cellcolor{white}---
    & \cellcolor{white}---
    & \cellcolor[RGB]{253,253,253}+0.006 \\
  Self\mbox{-}Refine
    & \cellcolor[RGB]{200,60,60}\textbf{-1.096}
    & \cellcolor[RGB]{200,60,60}\textbf{-0.907}
    & \cellcolor[RGB]{204,75,75}\textbf{-0.553}
    & \cellcolor[RGB]{218,125,125}\textbf{-0.399}
    & \cellcolor[RGB]{104,169,107}\textbf{+0.455}
    & \cellcolor[RGB]{99,166,103}\textbf{+0.467}
    & \cellcolor[RGB]{90,161,94}\textbf{+0.495}
    & \cellcolor[RGB]{90,161,93}\textbf{+0.496}
    & \cellcolor[RGB]{200,60,60}\textbf{-1.148}
    & \cellcolor[RGB]{200,60,60}\textbf{-0.833}
    & \cellcolor[RGB]{200,60,60}\textbf{-0.959}
    & \cellcolor[RGB]{200,60,60}\textbf{-0.858} \\
  \bottomrule
\end{tabular}%
}
\vspace{0.4em}
\resizebox{\textwidth}{!}{%
\begin{tabular}{l cccc cccc cccc}
  \toprule
  & \multicolumn{4}{c}{\textbf{WildBench}} & \multicolumn{4}{c}{\textbf{WritingBench}} & \multicolumn{4}{c}{\textbf{Overall}} \\
  \cmidrule(lr){2-5} \cmidrule(lr){6-9} \cmidrule(lr){10-13}
  & \footnotesize Low & \footnotesize Mid & \footnotesize High & \footnotesize XHigh & \footnotesize Low & \footnotesize Mid & \footnotesize High & \footnotesize XHigh & \footnotesize Low & \footnotesize Mid & \footnotesize High & \footnotesize XHigh \\
  \midrule
  BoN+Skywork
    & \cellcolor[RGB]{130,184,133}\textbf{+0.375}
    & \cellcolor[RGB]{125,181,127}\textbf{+0.391}
    & \cellcolor[RGB]{122,179,124}\textbf{+0.400}
    & \cellcolor[RGB]{118,177,120}\textbf{+0.413}
    & \cellcolor[RGB]{171,207,172}\textbf{+0.252}
    & \cellcolor[RGB]{168,205,170}\textbf{+0.261}
    & \cellcolor[RGB]{167,205,168}\textbf{+0.265}
    & \cellcolor[RGB]{164,203,166}\textbf{+0.273}
    & \cellcolor[RGB]{204,226,205}+0.152
    & \cellcolor[RGB]{203,225,204}+0.155
    & \cellcolor[RGB]{203,225,204}+0.155
    & \cellcolor[RGB]{204,226,205}+0.152 \\
  Fusion
    & \cellcolor{white}---
    & \cellcolor[RGB]{73,152,77}\textbf{+0.547}
    & \cellcolor[RGB]{56,142,60}\textbf{+0.642}
    & \cellcolor[RGB]{56,142,60}\textbf{+0.662}
    & \cellcolor{white}---
    & \cellcolor[RGB]{159,200,161}\textbf{+0.288}
    & \cellcolor[RGB]{195,220,196}+0.181
    & \cellcolor[RGB]{163,203,165}\textbf{+0.275}
    & \cellcolor{white}---
    & \cellcolor[RGB]{105,170,108}\textbf{+0.450}
    & \cellcolor[RGB]{121,179,124}\textbf{+0.401}
    & \cellcolor[RGB]{123,180,126}\textbf{+0.396} \\
  Particle Filter
    & \cellcolor{white}---
    & \cellcolor{white}---
    & \cellcolor{white}---
    & \cellcolor[RGB]{219,127,127}\textbf{-0.392}
    & \cellcolor{white}---
    & \cellcolor{white}---
    & \cellcolor{white}---
    & \cellcolor[RGB]{200,60,60}\textbf{-1.167}
    & \cellcolor{white}---
    & \cellcolor{white}---
    & \cellcolor{white}---
    & \cellcolor[RGB]{218,124,124}\textbf{-0.403} \\
  Self\mbox{-}Refine
    & \cellcolor[RGB]{137,188,140}\textbf{+0.353}
    & \cellcolor[RGB]{141,190,143}\textbf{+0.342}
    & \cellcolor[RGB]{160,201,162}\textbf{+0.286}
    & \cellcolor[RGB]{154,198,156}\textbf{+0.303}
    & \cellcolor[RGB]{56,142,60}\textbf{+0.869}
    & \cellcolor[RGB]{56,142,60}\textbf{+0.767}
    & \cellcolor[RGB]{56,142,60}\textbf{+0.792}
    & \cellcolor[RGB]{56,142,60}\textbf{+0.703}
    & \cellcolor[RGB]{244,218,218}-0.114
    & \cellcolor[RGB]{251,244,244}-0.033
    & \cellcolor[RGB]{251,252,251}+0.012
    & \cellcolor[RGB]{238,245,239}+0.049 \\
  \bottomrule
\end{tabular}%
}
\end{table}
    
    \begin{table}[H]
\centering
\setlength{\tabcolsep}{2.5pt}
\tiny
\caption{Headroom capture for \textbf{Qwen3.5-9B}: $h = (\text{realized} - \mu) / (\text{oracle}_\text{ideal} - \mu)$, where $\mu$ is the BoN $N{=}1$ baseline (single-inference mean). $h = 1$ means full exploitation of the oracle gap; $h = 0$ matches $\mu$; $h < 0$ underperforms $\mu$. \colorbox[RGB]{56,142,60}{\textcolor{white}{green}} = positive (gain captured), \colorbox[RGB]{200,60,60}{\textcolor{white}{red}} = negative (below baseline). Bold when $|h| \geq 0.20$.}
\label{tab:headroom_naive_qwen35_9b}
\resizebox{\textwidth}{!}{%
\begin{tabular}{l cccc cccc cccc}
  \toprule
  & \multicolumn{4}{c}{\textbf{HealthBench}} & \multicolumn{4}{c}{\textbf{PRBench}} & \multicolumn{4}{c}{\textbf{LEXam}} \\
  \cmidrule(lr){2-5} \cmidrule(lr){6-9} \cmidrule(lr){10-13}
  & \footnotesize Low & \footnotesize Mid & \footnotesize High & \footnotesize XHigh & \footnotesize Low & \footnotesize Mid & \footnotesize High & \footnotesize XHigh & \footnotesize Low & \footnotesize Mid & \footnotesize High & \footnotesize XHigh \\
  \midrule
  BoN+Skywork
    & \cellcolor[RGB]{227,239,227}+0.084
    & \cellcolor[RGB]{228,239,229}+0.080
    & \cellcolor[RGB]{227,239,227}+0.084
    & \cellcolor[RGB]{225,238,225}+0.089
    & \cellcolor[RGB]{229,240,230}+0.077
    & \cellcolor[RGB]{231,241,231}+0.071
    & \cellcolor[RGB]{234,243,235}+0.061
    & \cellcolor[RGB]{238,245,238}+0.050
    & \cellcolor[RGB]{230,241,231}+0.073
    & \cellcolor[RGB]{235,244,236}+0.058
    & \cellcolor[RGB]{235,243,235}+0.059
    & \cellcolor[RGB]{236,244,236}+0.056 \\
  Fusion
    & \cellcolor{white}---
    & \cellcolor[RGB]{56,142,60}\textbf{+0.615}
    & \cellcolor[RGB]{85,158,88}\textbf{+0.512}
    & \cellcolor[RGB]{98,165,101}\textbf{+0.473}
    & \cellcolor{white}---
    & \cellcolor[RGB]{155,198,157}\textbf{+0.301}
    & \cellcolor[RGB]{140,189,142}\textbf{+0.346}
    & \cellcolor[RGB]{151,196,153}\textbf{+0.313}
    & \cellcolor{white}---
    & \cellcolor[RGB]{116,176,118}\textbf{+0.419}
    & \cellcolor[RGB]{107,171,110}\textbf{+0.445}
    & \cellcolor[RGB]{142,191,144}\textbf{+0.339} \\
  Self\mbox{-}Refine
    & \cellcolor[RGB]{200,60,60}\textbf{-3.615}
    & \cellcolor[RGB]{200,60,60}\textbf{-1.681}
    & \cellcolor[RGB]{200,60,60}\textbf{-1.175}
    & \cellcolor[RGB]{200,60,60}\textbf{-0.918}
    & \cellcolor[RGB]{171,207,173}\textbf{+0.251}
    & \cellcolor[RGB]{147,194,149}\textbf{+0.323}
    & \cellcolor[RGB]{121,178,123}\textbf{+0.404}
    & \cellcolor[RGB]{145,193,148}\textbf{+0.329}
    & \cellcolor[RGB]{200,60,60}\textbf{-1.104}
    & \cellcolor[RGB]{200,60,60}\textbf{-0.970}
    & \cellcolor[RGB]{200,60,60}\textbf{-0.952}
    & \cellcolor[RGB]{200,60,60}\textbf{-1.147} \\
  \bottomrule
\end{tabular}%
}
\vspace{0.4em}
\resizebox{\textwidth}{!}{%
\begin{tabular}{l cccc cccc cccc}
  \toprule
  & \multicolumn{4}{c}{\textbf{WildBench}} & \multicolumn{4}{c}{\textbf{WritingBench}} & \multicolumn{4}{c}{\textbf{Overall}} \\
  \cmidrule(lr){2-5} \cmidrule(lr){6-9} \cmidrule(lr){10-13}
  & \footnotesize Low & \footnotesize Mid & \footnotesize High & \footnotesize XHigh & \footnotesize Low & \footnotesize Mid & \footnotesize High & \footnotesize XHigh & \footnotesize Low & \footnotesize Mid & \footnotesize High & \footnotesize XHigh \\
  \midrule
  BoN+Skywork
    & \cellcolor[RGB]{133,186,136}\textbf{+0.365}
    & \cellcolor[RGB]{136,187,138}\textbf{+0.358}
    & \cellcolor[RGB]{137,188,139}\textbf{+0.355}
    & \cellcolor[RGB]{140,189,142}\textbf{+0.345}
    & \cellcolor[RGB]{188,217,190}+0.200
    & \cellcolor[RGB]{187,216,188}\textbf{+0.204}
    & \cellcolor[RGB]{185,215,186}\textbf{+0.211}
    & \cellcolor[RGB]{186,216,188}\textbf{+0.206}
    & \cellcolor[RGB]{201,224,203}+0.160
    & \cellcolor[RGB]{203,225,204}+0.154
    & \cellcolor[RGB]{203,226,204}+0.154
    & \cellcolor[RGB]{205,226,206}+0.149 \\
  Fusion
    & \cellcolor{white}---
    & \cellcolor[RGB]{91,162,95}\textbf{+0.492}
    & \cellcolor[RGB]{87,159,90}\textbf{+0.506}
    & \cellcolor[RGB]{69,149,73}\textbf{+0.560}
    & \cellcolor{white}---
    & \cellcolor[RGB]{173,208,174}\textbf{+0.247}
    & \cellcolor[RGB]{195,221,196}+0.179
    & \cellcolor[RGB]{214,231,215}+0.123
    & \cellcolor{white}---
    & \cellcolor[RGB]{117,176,120}\textbf{+0.415}
    & \cellcolor[RGB]{123,180,125}\textbf{+0.398}
    & \cellcolor[RGB]{135,186,137}\textbf{+0.362} \\
  Self\mbox{-}Refine
    & \cellcolor[RGB]{235,244,236}+0.058
    & \cellcolor[RGB]{252,246,246}-0.027
    & \cellcolor[RGB]{239,246,239}+0.047
    & \cellcolor[RGB]{246,250,246}+0.026
    & \cellcolor[RGB]{56,142,60}\textbf{+0.833}
    & \cellcolor[RGB]{56,142,60}\textbf{+0.770}
    & \cellcolor[RGB]{56,142,60}\textbf{+0.700}
    & \cellcolor[RGB]{56,142,60}\textbf{+0.630}
    & \cellcolor[RGB]{200,60,60}\textbf{-0.715}
    & \cellcolor[RGB]{225,152,152}\textbf{-0.317}
    & \cellcolor[RGB]{237,191,191}-0.195
    & \cellcolor[RGB]{235,184,184}\textbf{-0.216} \\
  \bottomrule
\end{tabular}%
}
\end{table}
    
    \begin{table}[H]
\centering
\setlength{\tabcolsep}{2.5pt}
\tiny
\caption{Headroom capture for \textbf{OLMo3-7B}: $h = (\text{realized} - \mu) / (\text{oracle}_\text{ideal} - \mu)$, where $\mu$ is the BoN $N{=}1$ baseline (single-inference mean). $h = 1$ means full exploitation of the oracle gap; $h = 0$ matches $\mu$; $h < 0$ underperforms $\mu$. \colorbox[RGB]{56,142,60}{\textcolor{white}{green}} = positive (gain captured), \colorbox[RGB]{200,60,60}{\textcolor{white}{red}} = negative (below baseline). Bold when $|h| \geq 0.20$.}
\label{tab:headroom_naive_olmo3_7b}
\resizebox{\textwidth}{!}{%
\begin{tabular}{l cccc cccc cccc}
  \toprule
  & \multicolumn{4}{c}{\textbf{HealthBench}} & \multicolumn{4}{c}{\textbf{PRBench}} & \multicolumn{4}{c}{\textbf{LEXam}} \\
  \cmidrule(lr){2-5} \cmidrule(lr){6-9} \cmidrule(lr){10-13}
  & \footnotesize Low & \footnotesize Mid & \footnotesize High & \footnotesize XHigh & \footnotesize Low & \footnotesize Mid & \footnotesize High & \footnotesize XHigh & \footnotesize Low & \footnotesize Mid & \footnotesize High & \footnotesize XHigh \\
  \midrule
  BoN+Skywork
    & \cellcolor[RGB]{196,221,197}+0.177
    & \cellcolor[RGB]{200,223,201}+0.165
    & \cellcolor[RGB]{201,224,202}+0.160
    & \cellcolor[RGB]{200,223,201}+0.165
    & \cellcolor[RGB]{198,222,199}+0.170
    & \cellcolor[RGB]{209,229,210}+0.137
    & \cellcolor[RGB]{215,232,216}+0.119
    & \cellcolor[RGB]{218,234,219}+0.110
    & \cellcolor[RGB]{183,214,184}\textbf{+0.216}
    & \cellcolor[RGB]{198,222,199}+0.171
    & \cellcolor[RGB]{204,226,205}+0.152
    & \cellcolor[RGB]{202,225,203}+0.157 \\
  Fusion
    & \cellcolor{white}---
    & \cellcolor[RGB]{186,216,187}\textbf{+0.207}
    & \cellcolor[RGB]{158,200,160}\textbf{+0.292}
    & \cellcolor{white}---
    & \cellcolor{white}---
    & \cellcolor[RGB]{235,243,235}+0.059
    & \cellcolor[RGB]{180,212,182}\textbf{+0.224}
    & \cellcolor{white}---
    & \cellcolor{white}---
    & \cellcolor[RGB]{137,188,140}\textbf{+0.353}
    & \cellcolor[RGB]{159,201,161}\textbf{+0.286}
    & \cellcolor{white}--- \\
  \bottomrule
\end{tabular}%
}
\vspace{0.4em}
\resizebox{\textwidth}{!}{%
\begin{tabular}{l cccc cccc cccc}
  \toprule
  & \multicolumn{4}{c}{\textbf{WildBench}} & \multicolumn{4}{c}{\textbf{WritingBench}} & \multicolumn{4}{c}{\textbf{Overall}} \\
  \cmidrule(lr){2-5} \cmidrule(lr){6-9} \cmidrule(lr){10-13}
  & \footnotesize Low & \footnotesize Mid & \footnotesize High & \footnotesize XHigh & \footnotesize Low & \footnotesize Mid & \footnotesize High & \footnotesize XHigh & \footnotesize Low & \footnotesize Mid & \footnotesize High & \footnotesize XHigh \\
  \midrule
  BoN+Skywork
    & \cellcolor[RGB]{178,211,180}\textbf{+0.230}
    & \cellcolor[RGB]{189,218,191}+0.196
    & \cellcolor[RGB]{194,220,195}+0.184
    & \cellcolor[RGB]{194,220,195}+0.182
    & \cellcolor[RGB]{202,225,203}+0.158
    & \cellcolor[RGB]{209,228,210}+0.138
    & \cellcolor[RGB]{213,231,214}+0.126
    & \cellcolor[RGB]{213,231,214}+0.125
    & \cellcolor[RGB]{191,219,193}+0.190
    & \cellcolor[RGB]{201,224,202}+0.161
    & \cellcolor[RGB]{205,227,206}+0.148
    & \cellcolor[RGB]{205,227,206}+0.148 \\
  Fusion
    & \cellcolor{white}---
    & \cellcolor[RGB]{237,193,193}-0.190
    & \cellcolor{white}---
    & \cellcolor{white}---
    & \cellcolor{white}---
    & \cellcolor[RGB]{227,239,228}+0.083
    & \cellcolor[RGB]{206,227,207}+0.145
    & \cellcolor{white}---
    & \cellcolor{white}---
    & \cellcolor[RGB]{221,235,221}+0.102
    & \cellcolor[RGB]{176,210,178}\textbf{+0.237}
    & \cellcolor{white}--- \\
  \bottomrule
\end{tabular}%
}
\end{table}
    
\begin{table}[t]
\centering
\setlength{\tabcolsep}{2.5pt}
\tiny
\caption{Headroom capture for \textbf{OLMo3-32B}: $h = (\text{realized} - \mu) / (\text{oracle}_\text{ideal} - \mu)$, where $\mu$ is the BoN $N{=}1$ baseline (single-inference mean). $h = 1$ means full exploitation of the oracle gap; $h = 0$ matches $\mu$; $h < 0$ underperforms $\mu$. \colorbox[RGB]{56,142,60}{\textcolor{white}{green}} = positive (gain captured), \colorbox[RGB]{200,60,60}{\textcolor{white}{red}} = negative (below baseline). Bold when $|h| \geq 0.20$.}
\label{tab:headroom_naive_olmo3_32b}
\resizebox{\textwidth}{!}{%
\begin{tabular}{l cccc cccc cccc}
  \toprule
  & \multicolumn{4}{c}{\textbf{HealthBench}} & \multicolumn{4}{c}{\textbf{PRBench}} & \multicolumn{4}{c}{\textbf{LEXam}} \\
  \cmidrule(lr){2-5} \cmidrule(lr){6-9} \cmidrule(lr){10-13}
  & \footnotesize Low & \footnotesize Mid & \footnotesize High & \footnotesize XHigh & \footnotesize Low & \footnotesize Mid & \footnotesize High & \footnotesize XHigh & \footnotesize Low & \footnotesize Mid & \footnotesize High & \footnotesize XHigh \\
  \midrule
  BoN+Skywork
    & \cellcolor[RGB]{222,236,222}+0.099
    & \cellcolor[RGB]{223,236,223}+0.096
    & \cellcolor[RGB]{225,238,225}+0.090
    & \cellcolor[RGB]{226,238,227}+0.086
    & \cellcolor[RGB]{221,235,221}+0.102
    & \cellcolor[RGB]{227,239,228}+0.082
    & \cellcolor[RGB]{231,241,232}+0.070
    & \cellcolor[RGB]{229,240,230}+0.076
    & \cellcolor[RGB]{226,238,227}+0.085
    & \cellcolor[RGB]{233,242,233}+0.065
    & \cellcolor[RGB]{238,245,238}+0.049
    & \cellcolor[RGB]{246,250,246}+0.025 \\
  Fusion
    & \cellcolor{white}---
    & \cellcolor[RGB]{215,116,116}\textbf{-0.426}
    & \cellcolor[RGB]{225,149,149}\textbf{-0.326}
    & \cellcolor{white}---
    & \cellcolor{white}---
    & \cellcolor[RGB]{200,60,60}\textbf{-0.979}
    & \cellcolor[RGB]{200,60,60}\textbf{-1.016}
    & \cellcolor{white}---
    & \cellcolor{white}---
    & \cellcolor[RGB]{178,211,180}\textbf{+0.231}
    & \cellcolor[RGB]{179,211,180}\textbf{+0.228}
    & \cellcolor{white}--- \\
  \bottomrule
\end{tabular}%
}
\vspace{0.4em}
\resizebox{\textwidth}{!}{%
\begin{tabular}{l cccc cccc cccc}
  \toprule
  & \multicolumn{4}{c}{\textbf{WildBench}} & \multicolumn{4}{c}{\textbf{WritingBench}} & \multicolumn{4}{c}{\textbf{Overall}} \\
  \cmidrule(lr){2-5} \cmidrule(lr){6-9} \cmidrule(lr){10-13}
  & \footnotesize Low & \footnotesize Mid & \footnotesize High & \footnotesize XHigh & \footnotesize Low & \footnotesize Mid & \footnotesize High & \footnotesize XHigh & \footnotesize Low & \footnotesize Mid & \footnotesize High & \footnotesize XHigh \\
  \midrule
  BoN+Skywork
    & \cellcolor[RGB]{176,210,177}\textbf{+0.237}
    & \cellcolor[RGB]{181,213,183}\textbf{+0.221}
    & \cellcolor[RGB]{183,214,184}\textbf{+0.216}
    & \cellcolor[RGB]{180,212,181}\textbf{+0.225}
    & \cellcolor[RGB]{143,191,145}\textbf{+0.336}
    & \cellcolor[RGB]{151,196,153}\textbf{+0.312}
    & \cellcolor[RGB]{158,199,160}\textbf{+0.292}
    & \cellcolor[RGB]{170,207,172}\textbf{+0.255}
    & \cellcolor[RGB]{197,222,199}+0.172
    & \cellcolor[RGB]{203,225,204}+0.155
    & \cellcolor[RGB]{207,227,208}+0.143
    & \cellcolor[RGB]{210,229,211}+0.133 \\
  Fusion
    & \cellcolor{white}---
    & \cellcolor[RGB]{200,60,60}\textbf{-0.892}
    & \cellcolor{white}---
    & \cellcolor{white}---
    & \cellcolor{white}---
    & \cellcolor[RGB]{200,60,60}\textbf{-0.833}
    & \cellcolor[RGB]{236,188,188}\textbf{-0.204}
    & \cellcolor{white}---
    & \cellcolor{white}---
    & \cellcolor[RGB]{201,66,66}\textbf{-0.580}
    & \cellcolor[RGB]{224,147,147}\textbf{-0.330}
    & \cellcolor{white}--- \\
  \bottomrule
\end{tabular}%
}
\end{table}

\section{Qwen3.5 Results}
\label{app:qwen_results}

\subsection{Qwen3.5-35B-A3B Oracle Results}
\begin{table}[H]
\centering
\setlength{\tabcolsep}{2.5pt}
\tiny
\caption{Oracle quality for \textbf{Qwen3.5-35B} across five benchmarks
and four compute levels. Cell colour encodes delta from BoN oracle N=1 baseline:
\colorbox[RGB]{173,210,240}{steel blue} = large gain,
\colorbox[RGB]{225,237,248}{pale blue} = small gain over BoN N=1 baseline.
Bold = gain $\geq 0.03$ over baseline.}
\label{tab:oracle_scores_qwen35_35b}
\resizebox{\textwidth}{!}{%
\begin{tabular}{l ccccc ccccc ccccc}
  \toprule
  & \multicolumn{5}{c}{\textbf{HealthBench}} & \multicolumn{5}{c}{\textbf{PRBench}} & \multicolumn{5}{c}{\textbf{LEXam}} \\
  \cmidrule(lr){2-6} \cmidrule(lr){7-11} \cmidrule(lr){12-16}
  & \footnotesize N=1 & \footnotesize Low & \footnotesize Mid & \footnotesize High & \footnotesize XHigh & \footnotesize N=1 & \footnotesize Low & \footnotesize Mid & \footnotesize High & \footnotesize XHigh & \footnotesize N=1 & \footnotesize Low & \footnotesize Mid & \footnotesize High & \footnotesize XHigh \\
  \midrule
  BoN oracle
    & \cellbase{0.536}
    & \cellcolor[RGB]{231,242,251}\textbf{0.571}
    & \cellcolor[RGB]{212,232,247}\textbf{0.599}
    & \cellcolor[RGB]{198,223,244}\textbf{0.620}
    & \cellcolor[RGB]{186,217,242}\textbf{0.637}
    & \cellbase{0.292}
    & \cellcolor[RGB]{233,243,251}\textbf{0.325}
    & \cellcolor[RGB]{213,232,247}\textbf{0.354}
    & \cellcolor[RGB]{195,222,244}\textbf{0.380}
    & \cellcolor[RGB]{179,213,241}\textbf{0.403}
    & \cellbase{0.525}
    & \cellcolor[RGB]{231,242,251}\textbf{0.560}
    & \cellcolor[RGB]{213,232,247}\textbf{0.587}
    & \cellcolor[RGB]{198,224,245}\textbf{0.609}
    & \cellcolor[RGB]{186,217,242}\textbf{0.626} \\
  Fusion
    & \cellcolor{white}---
    & \cellcolor{white}---
    & \cellcolor[RGB]{216,234,248}\textbf{0.593}
    & \cellcolor[RGB]{195,222,244}\textbf{0.624}
    & \cellcolor[RGB]{184,216,242}\textbf{0.641}
    & \cellcolor{white}---
    & \cellcolor{white}---
    & \cellcolor[RGB]{216,234,248}\textbf{0.349}
    & \cellcolor[RGB]{194,221,244}\textbf{0.382}
    & \cellcolor[RGB]{177,212,241}\textbf{0.406}
    & \cellcolor{white}---
    & \cellcolor{white}---
    & \cellcolor[RGB]{215,233,248}\textbf{0.584}
    & \cellcolor[RGB]{196,223,244}\textbf{0.612}
    & \cellcolor[RGB]{184,216,242}\textbf{0.629} \\
  Particle Filter
    & \cellcolor{white}---
    & \cellcolor{white}---
    & \cellcolor{white}---
    & \cellcolor{white}---
    & \cellcolor[RGB]{229,241,250}\textbf{0.575}
    & \cellcolor{white}---
    & \cellcolor{white}---
    & \cellcolor{white}---
    & \cellcolor{white}---
    & \cellcolor[RGB]{231,242,251}\textbf{0.327}
    & \cellcolor{white}---
    & \cellcolor{white}---
    & \cellcolor{white}---
    & \cellcolor{white}---
    & \cellcolor[RGB]{217,234,248}\textbf{0.580} \\
  Self\mbox{-}Refine
    & \cellcolor{white}---
    & \cellcolor[RGB]{245,250,253}0.551
    & \cellcolor[RGB]{240,247,252}0.558
    & \cellcolor[RGB]{231,242,251}\textbf{0.572}
    & \cellcolor[RGB]{216,234,248}\textbf{0.594}
    & \cellcolor{white}---
    & \cellcolor[RGB]{240,247,252}0.313
    & \cellcolor[RGB]{227,240,250}\textbf{0.333}
    & \cellcolor[RGB]{207,229,246}\textbf{0.363}
    & \cellcolor[RGB]{186,217,242}\textbf{0.393}
    & \cellcolor{white}---
    & \cellcolor[RGB]{243,248,253}0.543
    & \cellcolor[RGB]{238,245,252}0.551
    & \cellcolor[RGB]{228,240,250}\textbf{0.564}
    & \cellcolor[RGB]{219,235,248}\textbf{0.578} \\
  \bottomrule
\end{tabular}%
}
\vspace{0.4em}
\resizebox{\textwidth}{!}{%
\begin{tabular}{l ccccc ccccc ccccc}
  \toprule
  & \multicolumn{5}{c}{\textbf{WildBench}} & \multicolumn{5}{c}{\textbf{WritingBench}} & \multicolumn{5}{c}{\textbf{Overall}} \\
  \cmidrule(lr){2-6} \cmidrule(lr){7-11} \cmidrule(lr){12-16}
  & \footnotesize N=1 & \footnotesize Low & \footnotesize Mid & \footnotesize High & \footnotesize XHigh & \footnotesize N=1 & \footnotesize Low & \footnotesize Mid & \footnotesize High & \footnotesize XHigh & \footnotesize N=1 & \footnotesize Low & \footnotesize Mid & \footnotesize High & \footnotesize XHigh \\
  \midrule
  BoN oracle
    & \cellbase{0.812}
    & \cellcolor[RGB]{232,242,251}\textbf{0.846}
    & \cellcolor[RGB]{218,234,248}\textbf{0.867}
    & \cellcolor[RGB]{209,230,247}\textbf{0.879}
    & \cellcolor[RGB]{205,227,246}\textbf{0.886}
    & \cellbase{0.702}
    & \cellcolor[RGB]{240,247,252}0.724
    & \cellcolor[RGB]{229,241,250}\textbf{0.740}
    & \cellcolor[RGB]{221,236,249}\textbf{0.752}
    & \cellcolor[RGB]{215,233,248}\textbf{0.761}
    & \cellbase{0.574}
    & \cellcolor[RGB]{233,243,251}\textbf{0.605}
    & \cellcolor[RGB]{217,234,248}\textbf{0.629}
    & \cellcolor[RGB]{204,227,246}\textbf{0.648}
    & \cellcolor[RGB]{194,222,244}\textbf{0.663} \\
  Fusion
    & \cellcolor{white}---
    & \cellcolor{white}---
    & \cellcolor[RGB]{215,233,248}\textbf{0.870}
    & \cellcolor[RGB]{207,229,246}\textbf{0.882}
    & \cellcolor[RGB]{202,226,245}\textbf{0.890}
    & \cellcolor{white}---
    & \cellcolor{white}---
    & \cellcolor[RGB]{225,238,249}\textbf{0.747}
    & \cellcolor[RGB]{218,234,248}\textbf{0.757}
    & \cellcolor[RGB]{211,231,247}\textbf{0.767}
    & \cellcolor{white}---
    & \cellcolor{white}---
    & \cellcolor[RGB]{217,234,248}\textbf{0.629}
    & \cellcolor[RGB]{202,226,245}\textbf{0.651}
    & \cellcolor[RGB]{192,220,243}\textbf{0.666} \\
  Particle Filter
    & \cellcolor{white}---
    & \cellcolor{white}---
    & \cellcolor{white}---
    & \cellcolor{white}---
    & \cellcolor[RGB]{228,240,250}\textbf{0.852}
    & \cellcolor{white}---
    & \cellcolor{white}---
    & \cellcolor{white}---
    & \cellcolor{white}---
    & \cellcolor[RGB]{247,251,254}0.714
    & \cellcolor{white}---
    & \cellcolor{white}---
    & \cellcolor{white}---
    & \cellcolor{white}---
    & \cellcolor[RGB]{230,241,250}\textbf{0.610} \\
  Self\mbox{-}Refine
    & \cellcolor{white}---
    & \cellcolor[RGB]{231,242,251}\textbf{0.848}
    & \cellcolor[RGB]{220,236,249}\textbf{0.864}
    & \cellcolor[RGB]{211,231,247}\textbf{0.876}
    & \cellcolor[RGB]{202,226,245}\textbf{0.891}
    & \cellcolor{white}---
    & \cellcolor[RGB]{227,240,250}\textbf{0.743}
    & \cellcolor[RGB]{212,232,247}\textbf{0.765}
    & \cellcolor[RGB]{197,223,244}\textbf{0.787}
    & \cellcolor[RGB]{184,216,242}\textbf{0.806}
    & \cellcolor{white}---
    & \cellcolor[RGB]{237,245,252}0.600
    & \cellcolor[RGB]{227,240,250}\textbf{0.614}
    & \cellcolor[RGB]{215,233,248}\textbf{0.632}
    & \cellcolor[RGB]{201,226,245}\textbf{0.652} \\
  \bottomrule
\end{tabular}%
}
\end{table}

\subsection{Qwen3.5-9B Realised Results}
\begin{table}[H]
\centering
\setlength{\tabcolsep}{2.5pt}
\tiny
\caption{Realised quality for \textbf{Qwen3.5-9B} across
benchmarks and four compute levels. Cell colour encodes delta from the
single-sample baseline (BoN@1):
\colorbox[RGB]{56,142,60}{\textcolor{white}{green}} = improvement,
\colorbox[RGB]{200,60,60}{\textcolor{white}{red}} = regression.
Bold = gain $\geq 0.03$ over baseline.}
\label{tab:results_qwen35_9b}
\resizebox{\textwidth}{!}{%
\begin{tabular}{l ccccc ccccc ccccc}
\toprule
  & \multicolumn{5}{c}{\textbf{HealthBench}} & \multicolumn{5}{c}{\textbf{PRBench}} & \multicolumn{5}{c}{\textbf{LEXam}} \\
  \cmidrule(lr){2-6} \cmidrule(lr){7-11} \cmidrule(lr){12-16}
  & \footnotesize N=1 & \footnotesize Low & \footnotesize Mid & \footnotesize High & \footnotesize XHigh & \footnotesize N=1 & \footnotesize Low & \footnotesize Mid & \footnotesize High & \footnotesize XHigh & \footnotesize N=1 & \footnotesize Low & \footnotesize Mid & \footnotesize High & \footnotesize XHigh \\
\midrule
  BoN+Skywork
    & \cellbase{0.508}
    & \cellcolor[RGB]{248,251,248}0.511
    & \cellcolor[RGB]{243,248,243}0.514
    & \cellcolor[RGB]{237,245,238}0.516
    & \cellcolor[RGB]{232,242,233}0.518
    & \cellbase{0.259}
    & \cellcolor[RGB]{250,252,250}0.262
    & \cellcolor[RGB]{246,250,246}0.264
    & \cellcolor[RGB]{244,249,244}0.265
    & \cellcolor[RGB]{244,249,244}0.265
    & \cellbase{0.463}
    & \cellcolor[RGB]{249,252,250}0.466
    & \cellcolor[RGB]{247,250,247}0.467
    & \cellcolor[RGB]{244,249,244}0.469
    & \cellcolor[RGB]{242,248,242}0.470 \\
  BoN+Llama70B
    & \cellbase{0.508}
    & \cellcolor[RGB]{247,251,247}0.511
    & \cellcolor[RGB]{242,248,242}0.514
    & \cellcolor[RGB]{236,244,237}0.517
    & \cellcolor[RGB]{230,241,231}0.519
    & \cellbase{0.259}
    & \cellcolor[RGB]{251,253,251}0.261
    & \cellcolor[RGB]{249,251,249}0.262
    & \cellcolor[RGB]{247,250,247}0.263
    & \cellcolor[RGB]{249,251,249}0.262
    & \cellbase{0.463}
    & \cellcolor[RGB]{245,250,246}0.468
    & \cellcolor[RGB]{238,245,238}0.471
    & \cellcolor[RGB]{231,241,231}0.475
    & \cellcolor[RGB]{231,241,231}0.475 \\
  Fusion
    & \cellcolor{white}---
    & \cellcolor{white}---
    & \cellcolor[RGB]{168,206,170}\textbf{0.548}
    & \cellcolor[RGB]{144,192,146}\textbf{0.560}
    & \cellcolor[RGB]{130,184,133}\textbf{0.566}
    & \cellcolor{white}---
    & \cellcolor{white}---
    & \cellcolor[RGB]{221,236,221}0.275
    & \cellcolor[RGB]{190,218,191}\textbf{0.290}
    & \cellcolor[RGB]{182,214,183}\textbf{0.293}
    & \cellcolor{white}---
    & \cellcolor{white}---
    & \cellcolor[RGB]{203,226,205}0.487
    & \cellcolor[RGB]{173,208,175}\textbf{0.502}
    & \cellcolor[RGB]{175,210,177}\textbf{0.501} \\
  Beam Search
    & \cellcolor{white}---
    & \cellcolor{white}---
    & \cellcolor[RGB]{254,254,254}0.508
    & \cellcolor[RGB]{254,249,249}0.505
    & \cellcolor[RGB]{254,250,250}0.506
    & \cellcolor{white}---
    & \cellcolor{white}---
    & \cellcolor[RGB]{253,244,244}0.254
    & \cellcolor[RGB]{252,241,241}0.253
    & \cellcolor[RGB]{252,243,243}0.254
    & \cellcolor{white}---
    & \cellcolor{white}---
    & \cellcolor[RGB]{230,241,231}0.475
    & \cellcolor[RGB]{219,235,220}0.480
    & \cellcolor[RGB]{255,255,255}0.463 \\
  Particle Filter
    & \cellcolor{white}---
    & \cellcolor{white}---
    & \cellcolor[RGB]{252,243,243}0.503
    & \cellcolor[RGB]{254,250,250}0.506
    & \cellcolor[RGB]{252,242,242}0.502
    & \cellcolor{white}---
    & \cellcolor{white}---
    & \cellcolor[RGB]{251,237,237}0.251
    & \cellcolor[RGB]{251,235,235}0.251
    & \cellcolor[RGB]{249,229,229}0.248
    & \cellcolor{white}---
    & \cellcolor{white}---
    & \cellcolor[RGB]{238,246,239}0.471
    & \cellcolor[RGB]{252,253,252}0.465
    & \cellcolor[RGB]{249,251,249}0.467 \\
  Sequential Refinement
    & \cellcolor{white}---
    & \cellcolor[RGB]{241,189,189}0.479
    & \cellcolor[RGB]{237,172,172}0.472
    & \cellcolor[RGB]{233,155,155}0.464
    & \cellcolor[RGB]{230,140,140}0.455
    & \cellcolor{white}---
    & \cellcolor[RGB]{246,250,246}0.264
    & \cellcolor[RGB]{234,243,235}0.269
    & \cellcolor[RGB]{205,227,206}0.283
    & \cellcolor[RGB]{196,222,197}0.287
    & \cellcolor{white}---
    & \cellcolor[RGB]{245,210,210}0.444
    & \cellcolor[RGB]{242,195,195}0.437
    & \cellcolor[RGB]{238,177,177}0.429
    & \cellcolor[RGB]{230,140,140}0.413 \\
  Budget Forcing
    & \cellcolor{white}---
    & \cellcolor[RGB]{232,242,232}0.519
    & \cellcolor[RGB]{221,236,222}0.524
    & \cellcolor{white}---
    & \cellcolor{white}---
    & \cellcolor{white}---
    & \cellcolor[RGB]{255,253,253}0.258
    & \cellcolor[RGB]{254,251,251}0.258
    & \cellcolor{white}---
    & \cellcolor{white}---
    & \cellcolor{white}---
    & \cellcolor[RGB]{250,252,250}0.466
    & \cellcolor[RGB]{253,254,253}0.465
    & \cellcolor{white}---
    & \cellcolor{white}--- \\
\bottomrule
\end{tabular}%
}
\vspace{0.4em}
\resizebox{\textwidth}{!}{%
\begin{tabular}{l ccccc ccccc ccccc}
\toprule
  & \multicolumn{5}{c}{\textbf{WildBench}} & \multicolumn{5}{c}{\textbf{WritingBench}} & \multicolumn{5}{c}{\textbf{Overall}} \\
  \cmidrule(lr){2-6} \cmidrule(lr){7-11} \cmidrule(lr){12-16}
  & \footnotesize N=1 & \footnotesize Low & \footnotesize Mid & \footnotesize High & \footnotesize XHigh & \footnotesize N=1 & \footnotesize Low & \footnotesize Mid & \footnotesize High & \footnotesize XHigh & \footnotesize N=1 & \footnotesize Low & \footnotesize Mid & \footnotesize High & \footnotesize XHigh \\
\midrule
  BoN+Skywork
    & \cellbase{0.733}
    & \cellcolor[RGB]{221,236,221}0.749
    & \cellcolor[RGB]{197,222,199}0.760
    & \cellcolor[RGB]{182,214,184}\textbf{0.767}
    & \cellcolor[RGB]{174,209,175}\textbf{0.771}
    & \cellbase{0.663}
    & \cellcolor[RGB]{243,248,243}0.668
    & \cellcolor[RGB]{234,243,234}0.673
    & \cellcolor[RGB]{226,239,227}0.676
    & \cellcolor[RGB]{222,236,223}0.678
    & \cellbase{0.525}
    & \cellcolor[RGB]{242,248,242}0.531
    & \cellcolor[RGB]{233,243,234}0.535
    & \cellcolor[RGB]{227,239,227}0.538
    & \cellcolor[RGB]{223,237,224}0.540 \\
  BoN+Llama70B
    & \cellbase{0.733}
    & \cellcolor[RGB]{231,241,231}0.744
    & \cellcolor[RGB]{218,234,219}0.750
    & \cellcolor[RGB]{213,231,214}0.753
    & \cellcolor[RGB]{212,231,213}0.753
    & \cellbase{0.663}
    & \cellcolor[RGB]{243,248,243}0.669
    & \cellcolor[RGB]{232,242,233}0.673
    & \cellcolor[RGB]{227,239,227}0.676
    & \cellcolor[RGB]{223,237,224}0.678
    & \cellbase{0.525}
    & \cellcolor[RGB]{243,248,244}0.531
    & \cellcolor[RGB]{236,244,236}0.534
    & \cellcolor[RGB]{231,241,231}0.537
    & \cellcolor[RGB]{229,240,230}0.537 \\
  Fusion
    & \cellcolor{white}---
    & \cellcolor{white}---
    & \cellcolor[RGB]{172,208,174}\textbf{0.771}
    & \cellcolor[RGB]{148,194,150}\textbf{0.783}
    & \cellcolor[RGB]{126,182,129}\textbf{0.798}
    & \cellcolor{white}---
    & \cellcolor{white}---
    & \cellcolor[RGB]{230,241,230}0.675
    & \cellcolor[RGB]{229,241,230}0.675
    & \cellcolor[RGB]{235,244,235}0.672
    & \cellcolor{white}---
    & \cellcolor{white}---
    & \cellcolor[RGB]{199,223,200}0.551
    & \cellcolor[RGB]{177,211,179}\textbf{0.562}
    & \cellcolor[RGB]{167,205,169}\textbf{0.566} \\
  Beam Search
    & \cellcolor{white}---
    & \cellcolor{white}---
    & \cellcolor[RGB]{254,255,254}0.733
    & \cellcolor[RGB]{228,240,229}0.746
    & \cellcolor[RGB]{254,250,250}0.731
    & \cellcolor{white}---
    & \cellcolor{white}---
    & \cellcolor[RGB]{254,252,252}0.662
    & \cellcolor[RGB]{251,235,235}0.654
    & \cellcolor[RGB]{252,241,241}0.657
    & \cellcolor{white}---
    & \cellcolor{white}---
    & \cellcolor[RGB]{252,254,252}0.527
    & \cellcolor[RGB]{250,252,250}0.528
    & \cellcolor[RGB]{253,248,248}0.522 \\
  Particle Filter
    & \cellcolor{white}---
    & \cellcolor{white}---
    & \cellcolor[RGB]{248,224,224}0.720
    & \cellcolor[RGB]{249,227,227}0.721
    & \cellcolor[RGB]{248,221,221}0.718
    & \cellcolor{white}---
    & \cellcolor{white}---
    & \cellcolor[RGB]{251,235,235}0.654
    & \cellcolor[RGB]{254,250,250}0.661
    & \cellcolor[RGB]{248,224,224}0.650
    & \cellcolor{white}---
    & \cellcolor{white}---
    & \cellcolor[RGB]{252,242,242}0.520
    & \cellcolor[RGB]{253,244,244}0.521
    & \cellcolor[RGB]{251,236,236}0.517 \\
  Sequential Refinement
    & \cellcolor{white}---
    & \cellcolor[RGB]{250,252,250}0.736
    & \cellcolor[RGB]{254,251,251}0.731
    & \cellcolor[RGB]{246,250,246}0.737
    & \cellcolor[RGB]{249,252,249}0.736
    & \cellcolor{white}---
    & \cellcolor[RGB]{184,215,186}\textbf{0.696}
    & \cellcolor[RGB]{154,198,157}\textbf{0.710}
    & \cellcolor[RGB]{126,182,129}\textbf{0.724}
    & \cellcolor[RGB]{126,182,129}\textbf{0.731}
    & \cellcolor{white}---
    & \cellcolor[RGB]{254,251,251}0.524
    & \cellcolor[RGB]{254,252,252}0.524
    & \cellcolor[RGB]{250,252,250}0.528
    & \cellcolor[RGB]{254,252,252}0.524 \\
  Budget Forcing
    & \cellcolor{white}---
    & \cellcolor[RGB]{215,232,216}0.752
    & \cellcolor[RGB]{215,232,216}0.752
    & \cellcolor{white}---
    & \cellcolor{white}---
    & \cellcolor{white}---
    & \cellcolor[RGB]{239,181,181}0.631
    & \cellcolor[RGB]{232,150,150}0.617
    & \cellcolor{white}---
    & \cellcolor{white}---
    & \cellcolor{white}---
    & \cellcolor[RGB]{255,255,255}0.525
    & \cellcolor[RGB]{254,249,249}0.523
    & \cellcolor{white}---
    & \cellcolor{white}--- \\
\bottomrule
\end{tabular}%
}
\end{table}

\subsection{Qwen3.5-9B Oracle Results}
\begin{table}[H]
\centering
\setlength{\tabcolsep}{2.5pt}
\tiny
\caption{Oracle quality for \textbf{Qwen3.5-9B} across five benchmarks
and four compute levels. Cell colour encodes delta from BoN oracle N=1 baseline:
\colorbox[RGB]{173,210,240}{steel blue} = large gain,
\colorbox[RGB]{225,237,248}{pale blue} = small gain over BoN N=1 baseline.
Bold = gain $\geq 0.03$ over baseline.}
\label{tab:oracle_scores_qwen35_9b}
\resizebox{\textwidth}{!}{%
\begin{tabular}{l ccccc ccccc ccccc}
  \toprule
  & \multicolumn{5}{c}{\textbf{HealthBench}} & \multicolumn{5}{c}{\textbf{PRBench}} & \multicolumn{5}{c}{\textbf{LEXam}} \\
  \cmidrule(lr){2-6} \cmidrule(lr){7-11} \cmidrule(lr){12-16}
  & \footnotesize N=1 & \footnotesize Low & \footnotesize Mid & \footnotesize High & \footnotesize XHigh & \footnotesize N=1 & \footnotesize Low & \footnotesize Mid & \footnotesize High & \footnotesize XHigh & \footnotesize N=1 & \footnotesize Low & \footnotesize Mid & \footnotesize High & \footnotesize XHigh \\
  \midrule
  BoN oracle
    & \cellbase{0.508}
    & \cellcolor[RGB]{227,240,250}\textbf{0.549}
    & \cellcolor[RGB]{205,228,246}\textbf{0.581}
    & \cellcolor[RGB]{188,218,243}\textbf{0.606}
    & \cellcolor[RGB]{174,210,240}\textbf{0.627}
    & \cellbase{0.259}
    & \cellcolor[RGB]{233,243,251}\textbf{0.291}
    & \cellcolor[RGB]{214,233,248}\textbf{0.319}
    & \cellcolor[RGB]{197,223,244}\textbf{0.344}
    & \cellcolor[RGB]{183,215,242}\textbf{0.365}
    & \cellbase{0.463}
    & \cellcolor[RGB]{231,242,251}\textbf{0.499}
    & \cellcolor[RGB]{211,231,247}\textbf{0.528}
    & \cellcolor[RGB]{194,222,244}\textbf{0.552}
    & \cellcolor[RGB]{181,214,241}\textbf{0.572} \\
  Fusion
    & \cellcolor{white}---
    & \cellcolor{white}---
    & \cellcolor[RGB]{210,230,247}\textbf{0.574}
    & \cellcolor[RGB]{186,217,242}\textbf{0.609}
    & \cellcolor[RGB]{173,210,240}\textbf{0.631}
    & \cellcolor{white}---
    & \cellcolor{white}---
    & \cellcolor[RGB]{219,235,248}\textbf{0.312}
    & \cellcolor[RGB]{195,222,244}\textbf{0.347}
    & \cellcolor[RGB]{180,214,241}\textbf{0.368}
    & \cellcolor{white}---
    & \cellcolor{white}---
    & \cellcolor[RGB]{216,234,248}\textbf{0.521}
    & \cellcolor[RGB]{196,223,244}\textbf{0.549}
    & \cellcolor[RGB]{180,214,241}\textbf{0.573} \\
  Self\mbox{-}Refine
    & \cellcolor{white}---
    & \cellcolor[RGB]{250,252,254}0.516
    & \cellcolor[RGB]{240,247,252}0.529
    & \cellcolor[RGB]{230,241,250}\textbf{0.545}
    & \cellcolor[RGB]{215,233,248}\textbf{0.566}
    & \cellcolor{white}---
    & \cellcolor[RGB]{243,249,253}0.276
    & \cellcolor[RGB]{235,244,251}0.289
    & \cellcolor[RGB]{216,233,248}\textbf{0.317}
    & \cellcolor[RGB]{198,224,245}\textbf{0.343}
    & \cellcolor{white}---
    & \cellcolor[RGB]{243,248,253}0.481
    & \cellcolor[RGB]{237,245,252}0.490
    & \cellcolor[RGB]{231,242,251}\textbf{0.499}
    & \cellcolor[RGB]{225,238,249}\textbf{0.508} \\
  \bottomrule
\end{tabular}%
}
\vspace{0.4em}
\resizebox{\textwidth}{!}{%
\begin{tabular}{l ccccc ccccc ccccc}
  \toprule
  & \multicolumn{5}{c}{\textbf{WildBench}} & \multicolumn{5}{c}{\textbf{WritingBench}} & \multicolumn{5}{c}{\textbf{Overall}} \\
  \cmidrule(lr){2-6} \cmidrule(lr){7-11} \cmidrule(lr){12-16}
  & \footnotesize N=1 & \footnotesize Low & \footnotesize Mid & \footnotesize High & \footnotesize XHigh & \footnotesize N=1 & \footnotesize Low & \footnotesize Mid & \footnotesize High & \footnotesize XHigh & \footnotesize N=1 & \footnotesize Low & \footnotesize Mid & \footnotesize High & \footnotesize XHigh \\
  \midrule
  BoN oracle
    & \cellbase{0.733}
    & \cellcolor[RGB]{225,239,250}\textbf{0.777}
    & \cellcolor[RGB]{204,227,246}\textbf{0.808}
    & \cellcolor[RGB]{190,219,243}\textbf{0.829}
    & \cellcolor[RGB]{180,214,241}\textbf{0.843}
    & \cellbase{0.663}
    & \cellcolor[RGB]{236,244,251}0.691
    & \cellcolor[RGB]{222,237,249}\textbf{0.711}
    & \cellcolor[RGB]{212,231,247}\textbf{0.726}
    & \cellcolor[RGB]{204,227,246}\textbf{0.737}
    & \cellbase{0.525}
    & \cellcolor[RGB]{230,241,250}\textbf{0.561}
    & \cellcolor[RGB]{211,231,247}\textbf{0.589}
    & \cellcolor[RGB]{196,223,244}\textbf{0.611}
    & \cellcolor[RGB]{184,216,242}\textbf{0.629} \\
  Fusion
    & \cellcolor{white}---
    & \cellcolor{white}---
    & \cellcolor[RGB]{202,226,245}\textbf{0.811}
    & \cellcolor[RGB]{188,218,243}\textbf{0.832}
    & \cellcolor[RGB]{176,212,241}\textbf{0.849}
    & \cellcolor{white}---
    & \cellcolor{white}---
    & \cellcolor[RGB]{223,237,249}\textbf{0.710}
    & \cellcolor[RGB]{210,230,247}\textbf{0.729}
    & \cellcolor[RGB]{203,226,245}\textbf{0.740}
    & \cellcolor{white}---
    & \cellcolor{white}---
    & \cellcolor[RGB]{214,232,247}\textbf{0.586}
    & \cellcolor[RGB]{195,222,244}\textbf{0.613}
    & \cellcolor[RGB]{182,215,242}\textbf{0.632} \\
  Self\mbox{-}Refine
    & \cellcolor{white}---
    & \cellcolor[RGB]{225,239,250}\textbf{0.777}
    & \cellcolor[RGB]{213,232,247}\textbf{0.795}
    & \cellcolor[RGB]{192,220,243}\textbf{0.826}
    & \cellcolor[RGB]{180,214,241}\textbf{0.843}
    & \cellcolor{white}---
    & \cellcolor[RGB]{228,240,250}\textbf{0.702}
    & \cellcolor[RGB]{213,232,247}\textbf{0.724}
    & \cellcolor[RGB]{195,222,244}\textbf{0.750}
    & \cellcolor[RGB]{182,215,242}\textbf{0.770}
    & \cellcolor{white}---
    & \cellcolor[RGB]{238,246,252}0.551
    & \cellcolor[RGB]{228,240,250}\textbf{0.565}
    & \cellcolor[RGB]{213,232,247}\textbf{0.587}
    & \cellcolor[RGB]{200,225,245}\textbf{0.606} \\
  \bottomrule
\end{tabular}%
}
\end{table}

\section{OLMo3 Results}
\label{app:olmo}

Tables~\ref{tab:olmo32b_results} and~\ref{tab:olmo7b_results} report
realised quality for OLMo3-32B and OLMo3-7B respectively.
Tables~\ref{tab:oracle_scores_olmo3_7b}--\ref{tab:oracle_scores_olmo3_32b}
report the corresponding oracle scores. Sequential Refinement was not evaluated
on OLMo3 due to compute constraints.
 
The OLMo3 results serve as the primary negative case for our central
thesis: generation capability does not imply exploitation capability.
 
\paragraph{Fusion severely degrades OLMo3-32B.}
Fusion---the only method to consistently improve Qwen3.5---produces
large regressions on OLMo3-32B. On PRBench, realised quality drops from
$0.211$ (baseline) to $0.164$ at High ($-22\%$ relative). WildBench
drops from $0.721$ to $0.653$, HealthBench from $0.480$ to $0.450$, and
WritingBench from $0.569$ to $0.516$. Only LEXam shows marginal
improvement ($0.395 \to 0.414$). Overall, OLMo3-32B Fusion scores
$0.441$ at Mid and $0.394$ at High, well below the $0.475$ baseline,
and the damage worsens with additional compute. This constitutes
\emph{negative} headroom capture: the synthesised output is
systematically worse than a random candidate from the same pool.
The contrast with Qwen3.5-35B-A3B, where Fusion captures ${\sim}40\%$
of headroom and improves on every benchmark, demonstrates that
exploitation is a distinct capability from generation.
 
\paragraph{OLMo3-7B Fusion is mixed but not broken.}
The smaller OLMo3-7B does not exhibit the same Fusion failure. At High
compute it improves over baseline on HealthBench ($+3.0$pp), LEXam
($+2.3$pp), PRBench ($+1.2$pp), and WritingBench ($+1.2$pp), while
regressing on WildBench at Mid ($-1.7$pp). This partial success
suggests that the 32B failure is not a simple property of the OLMo3
architecture, but reflects how scaling interacts with synthesis
capability in this model family.
 
\paragraph{Budget Forcing degenerates on OLMo3.}
Budget Forcing produces the largest regressions of any method.
OLMo3-32B drops from $0.475$ overall to $0.393$ at Mid; individual
benchmarks are worse: PRBench $0.211 \to 0.157$, WritingBench
$0.569 \to 0.421$. OLMo3-7B follows the same pattern (overall
$0.356 \to 0.288$). Additional thinking tokens produce degenerate
outputs rather than deeper reasoning.
 
\paragraph{Oracle quality confirms the dissociation.}
OLMo3-7B oracle quality rises steadily with compute
(Table~\ref{tab:oracle_scores_olmo3_7b}): overall from $0.356$ at
$N{=}1$ to $0.474$ at XHigh, with Fusion oracle tracking BoN oracle closely across compute levels. The candidate pool quality is proportionally
comparable to Qwen3.5, confirming that the divergence in realised
quality is driven entirely by exploitation, not exploration.

\begin{table}[H]
\centering
\setlength{\tabcolsep}{2.5pt}
\tiny
\caption{Realised quality for \textbf{OLMo3-32B} across five
benchmarks and four compute levels. Cell colour encodes delta from the
single-sample baseline (BoN@1):
\colorbox[RGB]{56,142,60}{\textcolor{white}{green}} = improvement,
\colorbox[RGB]{200,60,60}{\textcolor{white}{red}} = regression.
Bold = gain $\geq 0.03$ over baseline.}
\label{tab:olmo32b_results}
\resizebox{\textwidth}{!}{%
\begin{tabular}{l ccccc ccccc ccccc}
\toprule
  & \multicolumn{5}{c}{\textbf{HealthBench}} & \multicolumn{5}{c}{\textbf{PRBench}} & \multicolumn{5}{c}{\textbf{LEXam}} \\
  \cmidrule(lr){2-6} \cmidrule(lr){7-11} \cmidrule(lr){12-16}
  & \footnotesize N=1 & \footnotesize Low & \footnotesize Mid & \footnotesize High & \footnotesize XHigh & \footnotesize N=1 & \footnotesize Low & \footnotesize Mid & \footnotesize High & \footnotesize XHigh & \footnotesize N=1 & \footnotesize Low & \footnotesize Mid & \footnotesize High & \footnotesize XHigh \\
\midrule
  BoN+Skywork
    & \cellbase{0.480}
    & \cellcolor[RGB]{246,250,247}0.484
    & \cellcolor[RGB]{240,246,240}0.487
    & \cellcolor[RGB]{236,244,236}0.489
    & \cellcolor[RGB]{233,242,233}0.490
    & \cellbase{0.211}
    & \cellcolor[RGB]{249,252,250}0.213
    & \cellcolor[RGB]{246,250,247}0.215
    & \cellcolor[RGB]{244,249,245}0.215
    & \cellcolor[RGB]{240,247,241}0.217
    & \cellbase{0.395}
    & \cellcolor[RGB]{249,251,249}0.398
    & \cellcolor[RGB]{246,250,246}0.399
    & \cellcolor[RGB]{246,250,246}0.399
    & \cellcolor[RGB]{249,252,250}0.398 \\
  BoN+Llama70B
    & \cellbase{0.480}
    & \cellcolor[RGB]{247,251,247}0.483
    & \cellcolor[RGB]{242,248,243}0.486
    & \cellcolor[RGB]{239,246,240}0.487
    & \cellcolor[RGB]{236,244,237}0.488
    & \cellbase{0.211}
    & \cellcolor[RGB]{251,253,251}0.212
    & \cellcolor[RGB]{248,251,248}0.214
    & \cellcolor[RGB]{246,250,246}0.215
    & \cellcolor[RGB]{239,246,239}0.218
    & \cellbase{0.395}
    & \cellcolor[RGB]{246,250,246}0.399
    & \cellcolor[RGB]{241,247,241}0.401
    & \cellcolor[RGB]{234,243,235}0.405
    & \cellcolor[RGB]{227,239,227}0.408 \\
  Fusion
    & \cellcolor{white}---
    & \cellcolor{white}---
    & \cellcolor[RGB]{242,196,196}0.454
    & \cellcolor[RGB]{240,186,186}0.450
    & \cellcolor{white}---
    & \cellcolor{white}---
    & \cellcolor{white}---
    & \cellcolor[RGB]{237,171,171}0.174
    & \cellcolor[RGB]{232,149,149}0.164
    & \cellcolor{white}---
    & \cellcolor{white}---
    & \cellcolor{white}---
    & \cellcolor[RGB]{225,238,225}0.409
    & \cellcolor[RGB]{214,232,215}0.414
    & \cellcolor{white}--- \\
  Beam Search
    & \cellcolor{white}---
    & \cellcolor{white}---
    & \cellcolor[RGB]{254,253,253}0.479
    & \cellcolor[RGB]{254,254,254}0.480
    & \cellcolor{white}---
    & \cellcolor{white}---
    & \cellcolor{white}---
    & \cellcolor[RGB]{253,254,253}0.211
    & \cellcolor[RGB]{251,253,251}0.212
    & \cellcolor{white}---
    & \cellcolor{white}---
    & \cellcolor{white}---
    & \cellcolor[RGB]{254,251,251}0.393
    & \cellcolor[RGB]{250,234,234}0.386
    & \cellcolor{white}--- \\
  Particle Filter
    & \cellcolor{white}---
    & \cellcolor{white}---
    & \cellcolor[RGB]{254,252,252}0.479
    & \cellcolor[RGB]{255,255,255}0.480
    & \cellcolor[RGB]{254,249,249}0.477
    & \cellcolor{white}---
    & \cellcolor{white}---
    & \cellcolor[RGB]{255,253,253}0.210
    & \cellcolor[RGB]{253,246,246}0.207
    & \cellcolor[RGB]{255,254,254}0.210
    & \cellcolor{white}---
    & \cellcolor{white}---
    & \cellcolor[RGB]{253,244,244}0.390
    & \cellcolor[RGB]{231,242,232}0.406
    & \cellcolor[RGB]{255,255,255}0.395 \\
  Sequential Refinement
    & \cellcolor{white}---
    & \cellcolor{white}---
    & \cellcolor{white}---
    & \cellcolor{white}---
    & \cellcolor{white}---
    & \cellcolor{white}---
    & \cellcolor{white}---
    & \cellcolor{white}---
    & \cellcolor{white}---
    & \cellcolor{white}---
    & \cellcolor{white}---
    & \cellcolor{white}---
    & \cellcolor{white}---
    & \cellcolor{white}---
    & \cellcolor{white}--- \\
  Budget Forcing
    & \cellcolor{white}---
    & \cellcolor[RGB]{237,174,174}0.445
    & \cellcolor[RGB]{230,140,140}0.417
    & \cellcolor{white}---
    & \cellcolor{white}---
    & \cellcolor{white}---
    & \cellcolor[RGB]{241,192,192}0.183
    & \cellcolor[RGB]{230,140,140}0.157
    & \cellcolor{white}---
    & \cellcolor{white}---
    & \cellcolor{white}---
    & \cellcolor[RGB]{253,247,247}0.391
    & \cellcolor[RGB]{243,199,199}0.370
    & \cellcolor{white}---
    & \cellcolor{white}--- \\
\bottomrule
\end{tabular}%
}
\vspace{0.4em}
\resizebox{\textwidth}{!}{%
\begin{tabular}{l ccccc ccccc ccccc}
\toprule
  & \multicolumn{5}{c}{\textbf{WildBench}} & \multicolumn{5}{c}{\textbf{WritingBench}} & \multicolumn{5}{c}{\textbf{Overall}} \\
  \cmidrule(lr){2-6} \cmidrule(lr){7-11} \cmidrule(lr){12-16}
  & \footnotesize N=1 & \footnotesize Low & \footnotesize Mid & \footnotesize High & \footnotesize XHigh & \footnotesize N=1 & \footnotesize Low & \footnotesize Mid & \footnotesize High & \footnotesize XHigh & \footnotesize N=1 & \footnotesize Low & \footnotesize Mid & \footnotesize High & \footnotesize XHigh \\
\midrule
  BoN+Skywork
    & \cellbase{0.721}
    & \cellcolor[RGB]{230,241,231}0.733
    & \cellcolor[RGB]{216,233,217}0.739
    & \cellcolor[RGB]{206,227,207}0.744
    & \cellcolor[RGB]{195,221,196}0.749
    & \cellbase{0.569}
    & \cellcolor[RGB]{227,239,227}0.582
    & \cellcolor[RGB]{209,229,210}0.590
    & \cellcolor[RGB]{198,223,199}0.595
    & \cellcolor[RGB]{195,221,197}0.596
    & \cellbase{0.475}
    & \cellcolor[RGB]{240,247,241}0.482
    & \cellcolor[RGB]{231,242,232}0.486
    & \cellcolor[RGB]{226,238,226}0.488
    & \cellcolor[RGB]{223,237,223}0.490 \\
  BoN+Llama70B
    & \cellbase{0.721}
    & \cellcolor[RGB]{230,241,230}0.733
    & \cellcolor[RGB]{212,231,213}0.741
    & \cellcolor[RGB]{201,225,203}0.746
    & \cellcolor[RGB]{189,218,191}\textbf{0.752}
    & \cellbase{0.569}
    & \cellcolor[RGB]{233,242,233}0.579
    & \cellcolor[RGB]{218,234,219}0.586
    & \cellcolor[RGB]{208,229,209}0.590
    & \cellcolor[RGB]{199,223,200}0.595
    & \cellbase{0.475}
    & \cellcolor[RGB]{241,247,242}0.481
    & \cellcolor[RGB]{232,242,233}0.486
    & \cellcolor[RGB]{226,239,227}0.488
    & \cellcolor[RGB]{218,234,219}0.492 \\
  Fusion
    & \cellcolor{white}---
    & \cellcolor{white}---
    & \cellcolor[RGB]{230,140,140}0.653
    & \cellcolor{white}---
    & \cellcolor{white}---
    & \cellcolor{white}---
    & \cellcolor{white}---
    & \cellcolor[RGB]{230,140,140}0.516
    & \cellcolor[RGB]{246,212,212}0.550
    & \cellcolor{white}---
    & \cellcolor{white}---
    & \cellcolor{white}---
    & \cellcolor[RGB]{238,177,177}0.441
    & \cellcolor[RGB]{230,140,140}0.394
    & \cellcolor{white}--- \\
  Beam Search
    & \cellcolor{white}---
    & \cellcolor{white}---
    & \cellcolor[RGB]{248,222,222}0.707
    & \cellcolor[RGB]{249,226,226}0.709
    & \cellcolor{white}---
    & \cellcolor{white}---
    & \cellcolor{white}---
    & \cellcolor[RGB]{255,253,253}0.568
    & \cellcolor[RGB]{249,228,228}0.557
    & \cellcolor{white}---
    & \cellcolor{white}---
    & \cellcolor{white}---
    & \cellcolor[RGB]{253,247,247}0.472
    & \cellcolor[RGB]{252,241,241}0.469
    & \cellcolor{white}--- \\
  Particle Filter
    & \cellcolor{white}---
    & \cellcolor{white}---
    & \cellcolor[RGB]{248,221,221}0.707
    & \cellcolor[RGB]{249,226,226}0.709
    & \cellcolor[RGB]{251,236,236}0.713
    & \cellcolor{white}---
    & \cellcolor{white}---
    & \cellcolor[RGB]{245,249,245}0.573
    & \cellcolor[RGB]{255,253,253}0.568
    & \cellcolor[RGB]{254,250,250}0.566
    & \cellcolor{white}---
    & \cellcolor{white}---
    & \cellcolor[RGB]{253,247,247}0.472
    & \cellcolor[RGB]{254,252,252}0.474
    & \cellcolor[RGB]{254,249,249}0.472 \\
  Sequential Refinement
    & \cellcolor{white}---
    & \cellcolor{white}---
    & \cellcolor{white}---
    & \cellcolor{white}---
    & \cellcolor{white}---
    & \cellcolor{white}---
    & \cellcolor{white}---
    & \cellcolor{white}---
    & \cellcolor{white}---
    & \cellcolor{white}---
    & \cellcolor{white}---
    & \cellcolor{white}---
    & \cellcolor{white}---
    & \cellcolor{white}---
    & \cellcolor{white}--- \\
  Budget Forcing
    & \cellcolor{white}---
    & \cellcolor[RGB]{232,151,151}0.676
    & \cellcolor[RGB]{230,140,140}0.599
    & \cellcolor{white}---
    & \cellcolor{white}---
    & \cellcolor{white}---
    & \cellcolor[RGB]{230,140,140}0.501
    & \cellcolor[RGB]{230,140,140}0.421
    & \cellcolor{white}---
    & \cellcolor{white}---
    & \cellcolor{white}---
    & \cellcolor[RGB]{237,173,173}0.439
    & \cellcolor[RGB]{230,140,140}0.393
    & \cellcolor{white}---
    & \cellcolor{white}--- \\
\bottomrule
\end{tabular}%
}
\end{table}

\begin{table}[H]
\centering
\setlength{\tabcolsep}{2.5pt}
\tiny
\caption{Realised quality for \textbf{OLMo3-7B} across five
benchmarks and four compute levels. Cell colour encodes delta from the
single-sample baseline (BoN@1):
\colorbox[RGB]{56,142,60}{\textcolor{white}{green}} = improvement,
\colorbox[RGB]{200,60,60}{\textcolor{white}{red}} = regression.
Bold = gain $\geq 0.03$ over baseline.}
\label{tab:olmo7b_results}
\resizebox{\textwidth}{!}{%
\begin{tabular}{l ccccc ccccc ccccc}
\toprule
  & \multicolumn{5}{c}{\textbf{HealthBench}} & \multicolumn{5}{c}{\textbf{PRBench}} & \multicolumn{5}{c}{\textbf{LEXam}} \\
  \cmidrule(lr){2-6} \cmidrule(lr){7-11} \cmidrule(lr){12-16}
  & \footnotesize N=1 & \footnotesize Low & \footnotesize Mid & \footnotesize High & \footnotesize XHigh & \footnotesize N=1 & \footnotesize Low & \footnotesize Mid & \footnotesize High & \footnotesize XHigh & \footnotesize N=1 & \footnotesize Low & \footnotesize Mid & \footnotesize High & \footnotesize XHigh \\
\midrule
  BoN+Skywork
    & \cellbase{0.335}
    & \cellcolor[RGB]{239,246,239}0.342
    & \cellcolor[RGB]{227,239,228}0.348
    & \cellcolor[RGB]{217,233,218}0.353
    & \cellcolor[RGB]{206,227,207}0.358
    & \cellbase{0.142}
    & \cellcolor[RGB]{248,251,248}0.145
    & \cellcolor[RGB]{244,249,244}0.147
    & \cellcolor[RGB]{241,247,241}0.148
    & \cellcolor[RGB]{238,245,238}0.150
    & \cellbase{0.281}
    & \cellcolor[RGB]{240,247,241}0.288
    & \cellcolor[RGB]{233,242,233}0.291
    & \cellcolor[RGB]{227,239,227}0.294
    & \cellcolor[RGB]{218,234,219}0.298 \\
  BoN+Llama70B
    & \cellbase{0.335}
    & \cellcolor[RGB]{240,247,241}0.342
    & \cellcolor[RGB]{230,241,231}0.346
    & \cellcolor[RGB]{222,236,223}0.350
    & \cellcolor[RGB]{215,232,216}0.354
    & \cellbase{0.142}
    & \cellcolor[RGB]{249,252,249}0.144
    & \cellcolor[RGB]{245,250,246}0.146
    & \cellcolor[RGB]{243,248,243}0.147
    & \cellcolor[RGB]{238,245,238}0.150
    & \cellbase{0.281}
    & \cellcolor[RGB]{238,245,238}0.289
    & \cellcolor[RGB]{226,239,227}0.294
    & \cellcolor[RGB]{221,236,222}0.297
    & \cellcolor[RGB]{218,234,219}0.298 \\
  Fusion
    & \cellcolor{white}---
    & \cellcolor{white}---
    & \cellcolor[RGB]{225,238,226}0.349
    & \cellcolor[RGB]{190,218,192}\textbf{0.365}
    & \cellcolor{white}---
    & \cellcolor{white}---
    & \cellcolor{white}---
    & \cellcolor[RGB]{251,253,251}0.144
    & \cellcolor[RGB]{227,239,228}0.154
    & \cellcolor{white}---
    & \cellcolor{white}---
    & \cellcolor{white}---
    & \cellcolor[RGB]{214,232,215}0.300
    & \cellcolor[RGB]{206,227,207}0.304
    & \cellcolor{white}--- \\
  Beam Search
    & \cellcolor{white}---
    & \cellcolor{white}---
    & \cellcolor[RGB]{253,247,247}0.332
    & \cellcolor[RGB]{252,239,239}0.328
    & \cellcolor[RGB]{252,241,241}0.329
    & \cellcolor{white}---
    & \cellcolor{white}---
    & \cellcolor[RGB]{254,255,254}0.142
    & \cellcolor[RGB]{254,249,249}0.139
    & \cellcolor[RGB]{255,254,254}0.141
    & \cellcolor{white}---
    & \cellcolor{white}---
    & \cellcolor[RGB]{254,248,248}0.278
    & \cellcolor[RGB]{254,248,248}0.278
    & \cellcolor[RGB]{250,234,234}0.272 \\
  Particle Filter
    & \cellcolor{white}---
    & \cellcolor{white}---
    & \cellcolor[RGB]{255,253,253}0.334
    & \cellcolor[RGB]{244,249,244}0.340
    & \cellcolor[RGB]{253,254,253}0.336
    & \cellcolor{white}---
    & \cellcolor{white}---
    & \cellcolor[RGB]{253,248,248}0.139
    & \cellcolor[RGB]{255,253,253}0.141
    & \cellcolor[RGB]{254,250,250}0.140
    & \cellcolor{white}---
    & \cellcolor{white}---
    & \cellcolor[RGB]{251,237,237}0.273
    & \cellcolor[RGB]{255,255,255}0.281
    & \cellcolor[RGB]{254,249,249}0.279 \\
  Sequential Refinement
    & \cellcolor{white}---
    & \cellcolor{white}---
    & \cellcolor{white}---
    & \cellcolor{white}---
    & \cellcolor{white}---
    & \cellcolor{white}---
    & \cellcolor{white}---
    & \cellcolor{white}---
    & \cellcolor{white}---
    & \cellcolor{white}---
    & \cellcolor{white}---
    & \cellcolor{white}---
    & \cellcolor{white}---
    & \cellcolor{white}---
    & \cellcolor{white}--- \\
  Budget Forcing
    & \cellcolor{white}---
    & \cellcolor[RGB]{244,204,204}0.313
    & \cellcolor[RGB]{238,175,175}0.300
    & \cellcolor{white}---
    & \cellcolor{white}---
    & \cellcolor{white}---
    & \cellcolor[RGB]{242,196,196}0.116
    & \cellcolor[RGB]{232,150,150}0.096
    & \cellcolor{white}---
    & \cellcolor{white}---
    & \cellcolor{white}---
    & \cellcolor[RGB]{246,212,212}0.262
    & \cellcolor[RGB]{244,202,202}0.258
    & \cellcolor{white}---
    & \cellcolor{white}--- \\
\bottomrule
\end{tabular}%
}
\vspace{0.4em}
\resizebox{\textwidth}{!}{%
\begin{tabular}{l ccccc ccccc ccccc}
\toprule
  & \multicolumn{5}{c}{\textbf{WildBench}} & \multicolumn{5}{c}{\textbf{WritingBench}} & \multicolumn{5}{c}{\textbf{Overall}} \\
  \cmidrule(lr){2-6} \cmidrule(lr){7-11} \cmidrule(lr){12-16}
  & \footnotesize N=1 & \footnotesize Low & \footnotesize Mid & \footnotesize High & \footnotesize XHigh & \footnotesize N=1 & \footnotesize Low & \footnotesize Mid & \footnotesize High & \footnotesize XHigh & \footnotesize N=1 & \footnotesize Low & \footnotesize Mid & \footnotesize High & \footnotesize XHigh \\
\midrule
  BoN+Skywork
    & \cellbase{0.556}
    & \cellcolor[RGB]{228,240,229}0.568
    & \cellcolor[RGB]{213,231,214}0.575
    & \cellcolor[RGB]{201,225,202}0.581
    & \cellcolor[RGB]{190,218,192}\textbf{0.586}
    & \cellbase{0.466}
    & \cellcolor[RGB]{242,248,243}0.472
    & \cellcolor[RGB]{236,244,236}0.475
    & \cellcolor[RGB]{232,242,232}0.477
    & \cellcolor[RGB]{227,239,227}0.479
    & \cellbase{0.356}
    & \cellcolor[RGB]{240,246,240}0.363
    & \cellcolor[RGB]{230,241,231}0.367
    & \cellcolor[RGB]{223,237,224}0.371
    & \cellcolor[RGB]{216,233,217}0.374 \\
  BoN+Llama70B
    & \cellbase{0.556}
    & \cellcolor[RGB]{230,241,231}0.567
    & \cellcolor[RGB]{215,233,216}0.574
    & \cellcolor[RGB]{206,228,208}0.578
    & \cellcolor[RGB]{209,229,210}0.577
    & \cellbase{0.466}
    & \cellcolor[RGB]{244,249,244}0.471
    & \cellcolor[RGB]{238,245,238}0.474
    & \cellcolor[RGB]{233,242,233}0.476
    & \cellcolor[RGB]{227,239,227}0.479
    & \cellbase{0.356}
    & \cellcolor[RGB]{240,247,241}0.363
    & \cellcolor[RGB]{231,241,232}0.367
    & \cellcolor[RGB]{225,238,226}0.370
    & \cellcolor[RGB]{221,236,222}0.372 \\
  Fusion
    & \cellcolor{white}---
    & \cellcolor{white}---
    & \cellcolor[RGB]{247,218,218}0.539
    & \cellcolor{white}---
    & \cellcolor{white}---
    & \cellcolor{white}---
    & \cellcolor{white}---
    & \cellcolor[RGB]{244,249,244}0.471
    & \cellcolor[RGB]{228,240,229}0.478
    & \cellcolor{white}---
    & \cellcolor{white}---
    & \cellcolor{white}---
    & \cellcolor[RGB]{245,249,245}0.361
    & \cellcolor[RGB]{240,185,185}0.326
    & \cellcolor{white}--- \\
  Beam Search
    & \cellcolor{white}---
    & \cellcolor{white}---
    & \cellcolor[RGB]{252,239,239}0.549
    & \cellcolor[RGB]{254,250,250}0.553
    & \cellcolor[RGB]{250,233,233}0.546
    & \cellcolor{white}---
    & \cellcolor{white}---
    & \cellcolor[RGB]{247,250,247}0.470
    & \cellcolor[RGB]{239,246,239}0.473
    & \cellcolor[RGB]{253,254,254}0.467
    & \cellcolor{white}---
    & \cellcolor{white}---
    & \cellcolor[RGB]{254,251,251}0.354
    & \cellcolor[RGB]{254,252,252}0.354
    & \cellcolor[RGB]{253,244,244}0.351 \\
  Particle Filter
    & \cellcolor{white}---
    & \cellcolor{white}---
    & \cellcolor[RGB]{252,242,242}0.550
    & \cellcolor[RGB]{252,241,241}0.550
    & \cellcolor[RGB]{250,233,233}0.546
    & \cellcolor{white}---
    & \cellcolor{white}---
    & \cellcolor[RGB]{254,248,248}0.463
    & \cellcolor[RGB]{248,251,248}0.469
    & \cellcolor[RGB]{249,251,249}0.469
    & \cellcolor{white}---
    & \cellcolor{white}---
    & \cellcolor[RGB]{253,246,246}0.352
    & \cellcolor[RGB]{254,255,254}0.356
    & \cellcolor[RGB]{254,250,250}0.354 \\
  Sequential Refinement
    & \cellcolor{white}---
    & \cellcolor{white}---
    & \cellcolor{white}---
    & \cellcolor{white}---
    & \cellcolor{white}---
    & \cellcolor{white}---
    & \cellcolor{white}---
    & \cellcolor{white}---
    & \cellcolor{white}---
    & \cellcolor{white}---
    & \cellcolor{white}---
    & \cellcolor{white}---
    & \cellcolor{white}---
    & \cellcolor{white}---
    & \cellcolor{white}--- \\
  Budget Forcing
    & \cellcolor{white}---
    & \cellcolor[RGB]{230,140,140}0.489
    & \cellcolor[RGB]{230,140,140}0.435
    & \cellcolor{white}---
    & \cellcolor{white}---
    & \cellcolor{white}---
    & \cellcolor[RGB]{230,140,140}0.372
    & \cellcolor[RGB]{230,140,140}0.349
    & \cellcolor{white}---
    & \cellcolor{white}---
    & \cellcolor{white}---
    & \cellcolor[RGB]{232,150,150}0.310
    & \cellcolor[RGB]{230,140,140}0.288
    & \cellcolor{white}---
    & \cellcolor{white}--- \\
\bottomrule
\end{tabular}%
}
\end{table}

\subsection{Oracle Results Olmo-7B}

\begin{table}[H]
\centering
\setlength{\tabcolsep}{2.5pt}
\tiny
\caption{Oracle quality for \textbf{OLMo3-7B} across five benchmarks
and four compute levels. Cell colour encodes delta from BoN oracle N=1 baseline:
\colorbox[RGB]{173,210,240}{steel blue} = large gain,
\colorbox[RGB]{225,237,248}{pale blue} = small gain over BoN N=1 baseline.
Bold = gain $\geq 0.03$ over baseline.}
\label{tab:oracle_scores_olmo3_7b}
\resizebox{\textwidth}{!}{%
\begin{tabular}{l ccccc ccccc ccccc}
  \toprule
  & \multicolumn{5}{c}{\textbf{HealthBench}} & \multicolumn{5}{c}{\textbf{PRBench}} & \multicolumn{5}{c}{\textbf{LEXam}} \\
  \cmidrule(lr){2-6} \cmidrule(lr){7-11} \cmidrule(lr){12-16}
  & \footnotesize N=1 & \footnotesize Low & \footnotesize Mid & \footnotesize High & \footnotesize XHigh & \footnotesize N=1 & \footnotesize Low & \footnotesize Mid & \footnotesize High & \footnotesize XHigh & \footnotesize N=1 & \footnotesize Low & \footnotesize Mid & \footnotesize High & \footnotesize XHigh \\
  \midrule
  BoN oracle
    & \cellbase{0.335}
    & \cellcolor[RGB]{227,239,250}\textbf{0.376}
    & \cellcolor[RGB]{201,226,245}\textbf{0.413}
    & \cellcolor[RGB]{180,214,241}\textbf{0.445}
    & \cellcolor[RGB]{173,210,240}\textbf{0.473}
    & \cellbase{0.142}
    & \cellcolor[RGB]{242,248,253}0.161
    & \cellcolor[RGB]{229,241,250}\textbf{0.180}
    & \cellcolor[RGB]{217,234,248}\textbf{0.197}
    & \cellcolor[RGB]{206,228,246}\textbf{0.213}
    & \cellbase{0.281}
    & \cellcolor[RGB]{233,243,251}\textbf{0.313}
    & \cellcolor[RGB]{213,232,247}\textbf{0.342}
    & \cellcolor[RGB]{196,223,244}\textbf{0.368}
    & \cellcolor[RGB]{180,214,241}\textbf{0.391} \\
  Fusion
    & \cellcolor{white}---
    & \cellcolor{white}---
    & \cellcolor[RGB]{209,230,247}\textbf{0.403}
    & \cellcolor[RGB]{185,216,242}\textbf{0.438}
    & \cellcolor{white}---
    & \cellcolor{white}---
    & \cellcolor{white}---
    & \cellcolor[RGB]{232,243,251}\textbf{0.175}
    & \cellcolor[RGB]{216,234,248}\textbf{0.199}
    & \cellcolor{white}---
    & \cellcolor{white}---
    & \cellcolor{white}---
    & \cellcolor[RGB]{218,235,248}\textbf{0.335}
    & \cellcolor[RGB]{200,225,245}\textbf{0.361}
    & \cellcolor{white}--- \\
  \bottomrule
\end{tabular}%
}
\vspace{0.4em}
\resizebox{\textwidth}{!}{%
\begin{tabular}{l ccccc ccccc ccccc}
  \toprule
  & \multicolumn{5}{c}{\textbf{WildBench}} & \multicolumn{5}{c}{\textbf{WritingBench}} & \multicolumn{5}{c}{\textbf{Overall}} \\
  \cmidrule(lr){2-6} \cmidrule(lr){7-11} \cmidrule(lr){12-16}
  & \footnotesize N=1 & \footnotesize Low & \footnotesize Mid & \footnotesize High & \footnotesize XHigh & \footnotesize N=1 & \footnotesize Low & \footnotesize Mid & \footnotesize High & \footnotesize XHigh & \footnotesize N=1 & \footnotesize Low & \footnotesize Mid & \footnotesize High & \footnotesize XHigh \\
  \midrule
  BoN oracle
    & \cellbase{0.556}
    & \cellcolor[RGB]{218,235,248}\textbf{0.610}
    & \cellcolor[RGB]{187,218,243}\textbf{0.655}
    & \cellcolor[RGB]{173,210,240}\textbf{0.691}
    & \cellcolor[RGB]{173,210,240}\textbf{0.720}
    & \cellbase{0.466}
    & \cellcolor[RGB]{230,241,250}\textbf{0.503}
    & \cellcolor[RGB]{211,231,247}\textbf{0.530}
    & \cellcolor[RGB]{196,223,244}\textbf{0.553}
    & \cellcolor[RGB]{183,216,242}\textbf{0.571}
    & \cellbase{0.356}
    & \cellcolor[RGB]{230,241,250}\textbf{0.393}
    & \cellcolor[RGB]{208,229,246}\textbf{0.424}
    & \cellcolor[RGB]{190,219,243}\textbf{0.451}
    & \cellcolor[RGB]{174,211,240}\textbf{0.474} \\
  Fusion
    & \cellcolor{white}---
    & \cellcolor{white}---
    & \cellcolor[RGB]{197,223,244}\textbf{0.641}
    & \cellcolor{white}---
    & \cellcolor{white}---
    & \cellcolor{white}---
    & \cellcolor{white}---
    & \cellcolor[RGB]{212,232,247}\textbf{0.529}
    & \cellcolor[RGB]{196,223,244}\textbf{0.552}
    & \cellcolor{white}---
    & \cellcolor{white}---
    & \cellcolor{white}---
    & \cellcolor[RGB]{214,232,247}\textbf{0.416}
    & \cellcolor[RGB]{233,243,251}\textbf{0.387}
    & \cellcolor{white}--- \\
  \bottomrule
\end{tabular}%
}
\end{table}

\subsection{Oracle results Olmo3-32B}
\begin{table}[H]
\centering
\setlength{\tabcolsep}{2.5pt}
\tiny
\caption{Oracle quality for \textbf{OLMo3-32B} across five benchmarks
and four compute levels. Cell colour encodes delta from BoN oracle N=1 baseline:
\colorbox[RGB]{173,210,240}{steel blue} = large gain,
\colorbox[RGB]{225,237,248}{pale blue} = small gain over BoN N=1 baseline.
Bold = gain $\geq 0.03$ over baseline.}
\label{tab:oracle_scores_olmo3_32b}
\resizebox{\textwidth}{!}{%
\begin{tabular}{l ccccc ccccc ccccc}
  \toprule
  & \multicolumn{5}{c}{\textbf{HealthBench}} & \multicolumn{5}{c}{\textbf{PRBench}} & \multicolumn{5}{c}{\textbf{LEXam}} \\
  \cmidrule(lr){2-6} \cmidrule(lr){7-11} \cmidrule(lr){12-16}
  & \footnotesize N=1 & \footnotesize Low & \footnotesize Mid & \footnotesize High & \footnotesize XHigh & \footnotesize N=1 & \footnotesize Low & \footnotesize Mid & \footnotesize High & \footnotesize XHigh & \footnotesize N=1 & \footnotesize Low & \footnotesize Mid & \footnotesize High & \footnotesize XHigh \\
  \midrule
  BoN oracle
    & \cellbase{0.480}
    & \cellcolor[RGB]{227,240,250}\textbf{0.521}
    & \cellcolor[RGB]{205,227,246}\textbf{0.554}
    & \cellcolor[RGB]{186,217,242}\textbf{0.580}
    & \cellcolor[RGB]{173,210,240}\textbf{0.601}
    & \cellbase{0.211}
    & \cellcolor[RGB]{238,245,252}0.236
    & \cellcolor[RGB]{222,237,249}\textbf{0.259}
    & \cellcolor[RGB]{207,229,246}\textbf{0.281}
    & \cellcolor[RGB]{194,222,244}\textbf{0.300}
    & \cellbase{0.395}
    & \cellcolor[RGB]{231,242,251}\textbf{0.429}
    & \cellcolor[RGB]{211,231,247}\textbf{0.459}
    & \cellcolor[RGB]{195,222,244}\textbf{0.483}
    & \cellcolor[RGB]{182,215,242}\textbf{0.501} \\
  Fusion
    & \cellcolor{white}---
    & \cellcolor{white}---
    & \cellcolor[RGB]{214,232,247}\textbf{0.540}
    & \cellcolor[RGB]{192,221,244}\textbf{0.571}
    & \cellcolor{white}---
    & \cellcolor{white}---
    & \cellcolor{white}---
    & \cellcolor[RGB]{229,241,250}\textbf{0.248}
    & \cellcolor[RGB]{224,238,249}\textbf{0.256}
    & \cellcolor{white}---
    & \cellcolor{white}---
    & \cellcolor{white}---
    & \cellcolor[RGB]{213,232,247}\textbf{0.456}
    & \cellcolor[RGB]{198,224,245}\textbf{0.478}
    & \cellcolor{white}--- \\
  \bottomrule
\end{tabular}%
}
\vspace{0.4em}
\resizebox{\textwidth}{!}{%
\begin{tabular}{l ccccc ccccc ccccc}
  \toprule
  & \multicolumn{5}{c}{\textbf{WildBench}} & \multicolumn{5}{c}{\textbf{WritingBench}} & \multicolumn{5}{c}{\textbf{Overall}} \\
  \cmidrule(lr){2-6} \cmidrule(lr){7-11} \cmidrule(lr){12-16}
  & \footnotesize N=1 & \footnotesize Low & \footnotesize Mid & \footnotesize High & \footnotesize XHigh & \footnotesize N=1 & \footnotesize Low & \footnotesize Mid & \footnotesize High & \footnotesize XHigh & \footnotesize N=1 & \footnotesize Low & \footnotesize Mid & \footnotesize High & \footnotesize XHigh \\
  \midrule
  BoN oracle
    & \cellbase{0.721}
    & \cellcolor[RGB]{222,237,249}\textbf{0.769}
    & \cellcolor[RGB]{199,224,245}\textbf{0.804}
    & \cellcolor[RGB]{182,215,242}\textbf{0.828}
    & \cellcolor[RGB]{173,210,240}\textbf{0.846}
    & \cellbase{0.569}
    & \cellcolor[RGB]{228,240,250}\textbf{0.608}
    & \cellcolor[RGB]{208,229,246}\textbf{0.637}
    & \cellcolor[RGB]{193,221,244}\textbf{0.660}
    & \cellcolor[RGB]{180,214,241}\textbf{0.678}
    & \cellbase{0.475}
    & \cellcolor[RGB]{229,241,250}\textbf{0.513}
    & \cellcolor[RGB]{209,230,247}\textbf{0.543}
    & \cellcolor[RGB]{193,221,244}\textbf{0.566}
    & \cellcolor[RGB]{180,214,241}\textbf{0.585} \\
  Fusion
    & \cellcolor{white}---
    & \cellcolor{white}---
    & \cellcolor[RGB]{203,226,245}\textbf{0.797}
    & \cellcolor{white}---
    & \cellcolor{white}---
    & \cellcolor{white}---
    & \cellcolor{white}---
    & \cellcolor[RGB]{212,232,247}\textbf{0.631}
    & \cellcolor[RGB]{192,221,244}\textbf{0.660}
    & \cellcolor{white}---
    & \cellcolor{white}---
    & \cellcolor{white}---
    & \cellcolor[RGB]{214,233,248}\textbf{0.535}
    & \cellcolor[RGB]{244,249,253}0.491
    & \cellcolor{white}--- \\
  \bottomrule
\end{tabular}%
}
\end{table}


\section{Headroom capture tables}
\label{app:headroom_tables_adjusted}

\subsection{Qwen3.5-35B-A3B results}

\begin{table}[H]
\centering
\setlength{\tabcolsep}{2.5pt}
\tiny
\caption{Headroom capture for \textbf{Qwen3.5-35B}: $h = (\text{realized} - \mu) / (\text{oracle}_\text{ideal} - \mu)$, where $\mu$ is the BoN $N{=}1$ baseline (single-inference mean). $h = 1$ means full exploitation of the oracle gap; $h = 0$ matches $\mu$; $h < 0$ underperforms $\mu$. \colorbox[RGB]{56,142,60}{\textcolor{white}{green}} = positive (gain captured), \colorbox[RGB]{200,60,60}{\textcolor{white}{red}} = negative (below baseline). Bold when $|h| \geq 0.20$.}
\label{tab:headroom_qwen35_35b}
\resizebox{\textwidth}{!}{%
\begin{tabular}{l cccc cccc cccc}
  \toprule
  & \multicolumn{4}{c}{\textbf{HealthBench}} & \multicolumn{4}{c}{\textbf{PRBench}} & \multicolumn{4}{c}{\textbf{LEXam}} \\
  \cmidrule(lr){2-5} \cmidrule(lr){6-9} \cmidrule(lr){10-13}
  & \footnotesize XLow & \footnotesize Low & \footnotesize Mid & \footnotesize XHigh & \footnotesize XLow & \footnotesize Low & \footnotesize Mid & \footnotesize XHigh & \footnotesize XLow & \footnotesize Low & \footnotesize Mid & \footnotesize XHigh \\
  \midrule
  BoN+Skywork
    & \cellcolor[RGB]{239,246,239}+0.047
    & \cellcolor[RGB]{242,247,242}+0.038
    & \cellcolor[RGB]{244,249,245}+0.030
    & \cellcolor[RGB]{246,250,246}+0.025
    & \cellcolor[RGB]{242,247,242}+0.038
    & \cellcolor[RGB]{247,250,247}+0.023
    & \cellcolor[RGB]{250,252,250}+0.014
    & \cellcolor[RGB]{254,253,253}-0.005
    & \cellcolor[RGB]{239,246,239}+0.047
    & \cellcolor[RGB]{234,243,234}+0.062
    & \cellcolor[RGB]{233,242,234}+0.065
    & \cellcolor[RGB]{237,245,238}+0.051 \\
  Fusion
    & \cellcolor{white}---
    & \cellcolor[RGB]{56,142,60}\textbf{+0.670}
    & \cellcolor[RGB]{62,145,66}\textbf{+0.580}
    & \cellcolor[RGB]{90,161,94}\textbf{+0.495}
    & \cellcolor{white}---
    & \cellcolor[RGB]{125,181,128}\textbf{+0.390}
    & \cellcolor[RGB]{136,187,138}\textbf{+0.357}
    & \cellcolor[RGB]{142,191,144}\textbf{+0.339}
    & \cellcolor{white}---
    & \cellcolor[RGB]{137,188,139}\textbf{+0.355}
    & \cellcolor[RGB]{173,208,175}\textbf{+0.246}
    & \cellcolor[RGB]{185,215,186}\textbf{+0.209} \\
  Particle Filter
    & \cellcolor{white}---
    & \cellcolor{white}---
    & \cellcolor{white}---
    & \cellcolor[RGB]{239,200,200}-0.168
    & \cellcolor{white}---
    & \cellcolor{white}---
    & \cellcolor{white}---
    & \cellcolor[RGB]{228,159,159}\textbf{-0.293}
    & \cellcolor{white}---
    & \cellcolor{white}---
    & \cellcolor{white}---
    & \cellcolor[RGB]{253,253,253}+0.006 \\
  Self\mbox{-}Refine
    & \cellcolor[RGB]{200,60,60}\textbf{-1.096}
    & \cellcolor[RGB]{200,60,60}\textbf{-0.907}
    & \cellcolor[RGB]{204,75,75}\textbf{-0.553}
    & \cellcolor[RGB]{218,125,125}\textbf{-0.399}
    & \cellcolor[RGB]{104,169,107}\textbf{+0.455}
    & \cellcolor[RGB]{99,166,103}\textbf{+0.467}
    & \cellcolor[RGB]{90,161,94}\textbf{+0.495}
    & \cellcolor[RGB]{90,161,93}\textbf{+0.496}
    & \cellcolor[RGB]{200,60,60}\textbf{-1.148}
    & \cellcolor[RGB]{200,60,60}\textbf{-0.833}
    & \cellcolor[RGB]{200,60,60}\textbf{-0.959}
    & \cellcolor[RGB]{200,60,60}\textbf{-0.858} \\
  \bottomrule
\end{tabular}%
}
\vspace{0.4em}
\resizebox{\textwidth}{!}{%
\begin{tabular}{l cccc cccc cccc}
  \toprule
  & \multicolumn{4}{c}{\textbf{WildBench}} & \multicolumn{4}{c}{\textbf{WritingBench}} & \multicolumn{4}{c}{\textbf{Overall}} \\
  \cmidrule(lr){2-5} \cmidrule(lr){6-9} \cmidrule(lr){10-13}
  & \footnotesize XLow & \footnotesize Low & \footnotesize Mid & \footnotesize XHigh & \footnotesize XLow & \footnotesize Low & \footnotesize Mid & \footnotesize XHigh & \footnotesize XLow & \footnotesize Low & \footnotesize Mid & \footnotesize XHigh \\
  \midrule
  BoN+Skywork
    & \cellcolor[RGB]{130,184,133}\textbf{+0.375}
    & \cellcolor[RGB]{125,181,127}\textbf{+0.391}
    & \cellcolor[RGB]{122,179,124}\textbf{+0.400}
    & \cellcolor[RGB]{118,177,120}\textbf{+0.413}
    & \cellcolor[RGB]{171,207,172}\textbf{+0.252}
    & \cellcolor[RGB]{168,205,170}\textbf{+0.261}
    & \cellcolor[RGB]{167,205,168}\textbf{+0.265}
    & \cellcolor[RGB]{164,203,166}\textbf{+0.273}
    & \cellcolor[RGB]{204,226,205}+0.152
    & \cellcolor[RGB]{203,225,204}+0.155
    & \cellcolor[RGB]{203,225,204}+0.155
    & \cellcolor[RGB]{204,226,205}+0.152 \\
  Fusion
    & \cellcolor{white}---
    & \cellcolor[RGB]{73,152,77}\textbf{+0.547}
    & \cellcolor[RGB]{56,142,60}\textbf{+0.642}
    & \cellcolor[RGB]{56,142,60}\textbf{+0.662}
    & \cellcolor{white}---
    & \cellcolor[RGB]{159,200,161}\textbf{+0.288}
    & \cellcolor[RGB]{195,220,196}+0.181
    & \cellcolor[RGB]{163,203,165}\textbf{+0.275}
    & \cellcolor{white}---
    & \cellcolor[RGB]{105,170,108}\textbf{+0.450}
    & \cellcolor[RGB]{121,179,124}\textbf{+0.401}
    & \cellcolor[RGB]{123,180,126}\textbf{+0.396} \\
  Particle Filter
    & \cellcolor{white}---
    & \cellcolor{white}---
    & \cellcolor{white}---
    & \cellcolor[RGB]{219,127,127}\textbf{-0.392}
    & \cellcolor{white}---
    & \cellcolor{white}---
    & \cellcolor{white}---
    & \cellcolor[RGB]{200,60,60}\textbf{-1.167}
    & \cellcolor{white}---
    & \cellcolor{white}---
    & \cellcolor{white}---
    & \cellcolor[RGB]{218,124,124}\textbf{-0.403} \\
  Self\mbox{-}Refine
    & \cellcolor[RGB]{137,188,140}\textbf{+0.353}
    & \cellcolor[RGB]{141,190,143}\textbf{+0.342}
    & \cellcolor[RGB]{160,201,162}\textbf{+0.286}
    & \cellcolor[RGB]{154,198,156}\textbf{+0.303}
    & \cellcolor[RGB]{56,142,60}\textbf{+0.869}
    & \cellcolor[RGB]{56,142,60}\textbf{+0.767}
    & \cellcolor[RGB]{56,142,60}\textbf{+0.792}
    & \cellcolor[RGB]{56,142,60}\textbf{+0.703}
    & \cellcolor[RGB]{244,218,218}-0.114
    & \cellcolor[RGB]{251,244,244}-0.033
    & \cellcolor[RGB]{251,252,251}+0.012
    & \cellcolor[RGB]{238,245,239}+0.049 \\
  \bottomrule
\end{tabular}%
}
\end{table}

\subsection{Qwen3.5-9B results}

\begin{table}[H]
\centering
\setlength{\tabcolsep}{2.5pt}
\tiny
\caption{Headroom capture for \textbf{Qwen3.5-9B}: $h = (\text{realized} - \mu) / (\text{oracle}_\text{ideal} - \mu)$, where $\mu$ is the BoN $N{=}1$ baseline (single-inference mean). $h = 1$ means full exploitation of the oracle gap; $h = 0$ matches $\mu$; $h < 0$ underperforms $\mu$. \colorbox[RGB]{56,142,60}{\textcolor{white}{green}} = positive (gain captured), \colorbox[RGB]{200,60,60}{\textcolor{white}{red}} = negative (below baseline). Bold when $|h| \geq 0.20$.}
\label{tab:headroom_qwen35_9b}
\resizebox{\textwidth}{!}{%
\begin{tabular}{l cccc cccc cccc}
  \toprule
  & \multicolumn{4}{c}{\textbf{HealthBench}} & \multicolumn{4}{c}{\textbf{PRBench}} & \multicolumn{4}{c}{\textbf{LEXam}} \\
  \cmidrule(lr){2-5} \cmidrule(lr){6-9} \cmidrule(lr){10-13}
  & \footnotesize XLow & \footnotesize Low & \footnotesize Mid & \footnotesize XHigh & \footnotesize XLow & \footnotesize Low & \footnotesize Mid & \footnotesize XHigh & \footnotesize XLow & \footnotesize Low & \footnotesize Mid & \footnotesize XHigh \\
  \midrule
  BoN+Skywork
    & \cellcolor[RGB]{227,239,227}+0.084
    & \cellcolor[RGB]{228,239,229}+0.080
    & \cellcolor[RGB]{227,239,227}+0.084
    & \cellcolor[RGB]{225,238,225}+0.089
    & \cellcolor[RGB]{229,240,230}+0.077
    & \cellcolor[RGB]{231,241,231}+0.071
    & \cellcolor[RGB]{234,243,235}+0.061
    & \cellcolor[RGB]{238,245,238}+0.050
    & \cellcolor[RGB]{230,241,231}+0.073
    & \cellcolor[RGB]{235,244,236}+0.058
    & \cellcolor[RGB]{235,243,235}+0.059
    & \cellcolor[RGB]{236,244,236}+0.056 \\
  Fusion
    & \cellcolor{white}---
    & \cellcolor[RGB]{56,142,60}\textbf{+0.615}
    & \cellcolor[RGB]{85,158,88}\textbf{+0.512}
    & \cellcolor[RGB]{98,165,101}\textbf{+0.473}
    & \cellcolor{white}---
    & \cellcolor[RGB]{155,198,157}\textbf{+0.301}
    & \cellcolor[RGB]{140,189,142}\textbf{+0.346}
    & \cellcolor[RGB]{151,196,153}\textbf{+0.313}
    & \cellcolor{white}---
    & \cellcolor[RGB]{116,176,118}\textbf{+0.419}
    & \cellcolor[RGB]{107,171,110}\textbf{+0.445}
    & \cellcolor[RGB]{142,191,144}\textbf{+0.339} \\
  Self\mbox{-}Refine
    & \cellcolor[RGB]{200,60,60}\textbf{-3.615}
    & \cellcolor[RGB]{200,60,60}\textbf{-1.681}
    & \cellcolor[RGB]{200,60,60}\textbf{-1.175}
    & \cellcolor[RGB]{200,60,60}\textbf{-0.918}
    & \cellcolor[RGB]{171,207,173}\textbf{+0.251}
    & \cellcolor[RGB]{147,194,149}\textbf{+0.323}
    & \cellcolor[RGB]{121,178,123}\textbf{+0.404}
    & \cellcolor[RGB]{145,193,148}\textbf{+0.329}
    & \cellcolor[RGB]{200,60,60}\textbf{-1.104}
    & \cellcolor[RGB]{200,60,60}\textbf{-0.970}
    & \cellcolor[RGB]{200,60,60}\textbf{-0.952}
    & \cellcolor[RGB]{200,60,60}\textbf{-1.147} \\
  \bottomrule
\end{tabular}%
}
\vspace{0.4em}
\resizebox{\textwidth}{!}{%
\begin{tabular}{l cccc cccc cccc}
  \toprule
  & \multicolumn{4}{c}{\textbf{WildBench}} & \multicolumn{4}{c}{\textbf{WritingBench}} & \multicolumn{4}{c}{\textbf{Overall}} \\
  \cmidrule(lr){2-5} \cmidrule(lr){6-9} \cmidrule(lr){10-13}
  & \footnotesize XLow & \footnotesize Low & \footnotesize Mid & \footnotesize XHigh & \footnotesize XLow & \footnotesize Low & \footnotesize Mid & \footnotesize XHigh & \footnotesize XLow & \footnotesize Low & \footnotesize Mid & \footnotesize XHigh \\
  \midrule
  BoN+Skywork
    & \cellcolor[RGB]{133,186,136}\textbf{+0.365}
    & \cellcolor[RGB]{136,187,138}\textbf{+0.358}
    & \cellcolor[RGB]{137,188,139}\textbf{+0.355}
    & \cellcolor[RGB]{140,189,142}\textbf{+0.345}
    & \cellcolor[RGB]{188,217,190}+0.200
    & \cellcolor[RGB]{187,216,188}\textbf{+0.204}
    & \cellcolor[RGB]{185,215,186}\textbf{+0.211}
    & \cellcolor[RGB]{186,216,188}\textbf{+0.206}
    & \cellcolor[RGB]{201,224,203}+0.160
    & \cellcolor[RGB]{203,225,204}+0.154
    & \cellcolor[RGB]{203,226,204}+0.154
    & \cellcolor[RGB]{205,226,206}+0.149 \\
  Fusion
    & \cellcolor{white}---
    & \cellcolor[RGB]{91,162,95}\textbf{+0.492}
    & \cellcolor[RGB]{87,159,90}\textbf{+0.506}
    & \cellcolor[RGB]{69,149,73}\textbf{+0.560}
    & \cellcolor{white}---
    & \cellcolor[RGB]{173,208,174}\textbf{+0.247}
    & \cellcolor[RGB]{195,221,196}+0.179
    & \cellcolor[RGB]{214,231,215}+0.123
    & \cellcolor{white}---
    & \cellcolor[RGB]{117,176,120}\textbf{+0.415}
    & \cellcolor[RGB]{123,180,125}\textbf{+0.398}
    & \cellcolor[RGB]{135,186,137}\textbf{+0.362} \\
  Self\mbox{-}Refine
    & \cellcolor[RGB]{235,244,236}+0.058
    & \cellcolor[RGB]{252,246,246}-0.027
    & \cellcolor[RGB]{239,246,239}+0.047
    & \cellcolor[RGB]{246,250,246}+0.026
    & \cellcolor[RGB]{56,142,60}\textbf{+0.833}
    & \cellcolor[RGB]{56,142,60}\textbf{+0.770}
    & \cellcolor[RGB]{56,142,60}\textbf{+0.700}
    & \cellcolor[RGB]{56,142,60}\textbf{+0.630}
    & \cellcolor[RGB]{200,60,60}\textbf{-0.715}
    & \cellcolor[RGB]{225,152,152}\textbf{-0.317}
    & \cellcolor[RGB]{237,191,191}-0.195
    & \cellcolor[RGB]{235,184,184}\textbf{-0.216} \\
  \bottomrule
\end{tabular}%
}
\end{table}

\subsection{Olmo3-32B results}

\begin{table}[H]
\centering
\setlength{\tabcolsep}{2.5pt}
\tiny
\caption{Headroom capture for \textbf{OLMo3-32B}: $h = (\text{realized} - \mu) / (\text{oracle}_\text{ideal} - \mu)$, where $\mu$ is the BoN $N{=}1$ baseline (single-inference mean). $h = 1$ means full exploitation of the oracle gap; $h = 0$ matches $\mu$; $h < 0$ underperforms $\mu$. \colorbox[RGB]{56,142,60}{\textcolor{white}{green}} = positive (gain captured), \colorbox[RGB]{200,60,60}{\textcolor{white}{red}} = negative (below baseline). Bold when $|h| \geq 0.20$.}
\label{tab:headroom_olmo3_32b}
\resizebox{\textwidth}{!}{%
\begin{tabular}{l cccc cccc cccc}
  \toprule
  & \multicolumn{4}{c}{\textbf{HealthBench}} & \multicolumn{4}{c}{\textbf{PRBench}} & \multicolumn{4}{c}{\textbf{LEXam}} \\
  \cmidrule(lr){2-5} \cmidrule(lr){6-9} \cmidrule(lr){10-13}
  & \footnotesize XLow & \footnotesize Low & \footnotesize Mid & \footnotesize XHigh & \footnotesize XLow & \footnotesize Low & \footnotesize Mid & \footnotesize XHigh & \footnotesize XLow & \footnotesize Low & \footnotesize Mid & \footnotesize XHigh \\
  \midrule
  BoN+Skywork
    & \cellcolor[RGB]{222,236,222}+0.099
    & \cellcolor[RGB]{223,236,223}+0.096
    & \cellcolor[RGB]{225,238,225}+0.090
    & \cellcolor[RGB]{226,238,227}+0.086
    & \cellcolor[RGB]{221,235,221}+0.102
    & \cellcolor[RGB]{227,239,228}+0.082
    & \cellcolor[RGB]{231,241,232}+0.070
    & \cellcolor[RGB]{229,240,230}+0.076
    & \cellcolor[RGB]{226,238,227}+0.085
    & \cellcolor[RGB]{233,242,233}+0.065
    & \cellcolor[RGB]{238,245,238}+0.049
    & \cellcolor[RGB]{246,250,246}+0.025 \\
  Fusion
    & \cellcolor{white}---
    & \cellcolor[RGB]{215,116,116}\textbf{-0.426}
    & \cellcolor[RGB]{225,149,149}\textbf{-0.326}
    & \cellcolor{white}---
    & \cellcolor{white}---
    & \cellcolor[RGB]{200,60,60}\textbf{-0.979}
    & \cellcolor[RGB]{200,60,60}\textbf{-1.016}
    & \cellcolor{white}---
    & \cellcolor{white}---
    & \cellcolor[RGB]{178,211,180}\textbf{+0.231}
    & \cellcolor[RGB]{179,211,180}\textbf{+0.228}
    & \cellcolor{white}--- \\
  \bottomrule
\end{tabular}%
}
\vspace{0.4em}
\resizebox{\textwidth}{!}{%
\begin{tabular}{l cccc cccc cccc}
  \toprule
  & \multicolumn{4}{c}{\textbf{WildBench}} & \multicolumn{4}{c}{\textbf{WritingBench}} & \multicolumn{4}{c}{\textbf{Overall}} \\
  \cmidrule(lr){2-5} \cmidrule(lr){6-9} \cmidrule(lr){10-13}
  & \footnotesize XLow & \footnotesize Low & \footnotesize Mid & \footnotesize XHigh & \footnotesize XLow & \footnotesize Low & \footnotesize Mid & \footnotesize XHigh & \footnotesize XLow & \footnotesize Low & \footnotesize Mid & \footnotesize XHigh \\
  \midrule
  BoN+Skywork
    & \cellcolor[RGB]{176,210,177}\textbf{+0.237}
    & \cellcolor[RGB]{181,213,183}\textbf{+0.221}
    & \cellcolor[RGB]{183,214,184}\textbf{+0.216}
    & \cellcolor[RGB]{180,212,181}\textbf{+0.225}
    & \cellcolor[RGB]{143,191,145}\textbf{+0.336}
    & \cellcolor[RGB]{151,196,153}\textbf{+0.312}
    & \cellcolor[RGB]{158,199,160}\textbf{+0.292}
    & \cellcolor[RGB]{170,207,172}\textbf{+0.255}
    & \cellcolor[RGB]{197,222,199}+0.172
    & \cellcolor[RGB]{203,225,204}+0.155
    & \cellcolor[RGB]{207,227,208}+0.143
    & \cellcolor[RGB]{210,229,211}+0.133 \\
  Fusion
    & \cellcolor{white}---
    & \cellcolor[RGB]{200,60,60}\textbf{-0.892}
    & \cellcolor{white}---
    & \cellcolor{white}---
    & \cellcolor{white}---
    & \cellcolor[RGB]{200,60,60}\textbf{-0.833}
    & \cellcolor[RGB]{236,188,188}\textbf{-0.204}
    & \cellcolor{white}---
    & \cellcolor{white}---
    & \cellcolor[RGB]{201,66,66}\textbf{-0.580}
    & \cellcolor[RGB]{224,147,147}\textbf{-0.330}
    & \cellcolor{white}--- \\
  \bottomrule
\end{tabular}%
}
\end{table}

\subsection{Olmo3-7B results}
\begin{table}[H]
\centering
\setlength{\tabcolsep}{2.5pt}
\tiny
\caption{Headroom capture for \textbf{OLMo3-7B}: $h = (\text{realized} - \mu) / (\text{oracle}_\text{ideal} - \mu)$, where $\mu$ is the BoN $N{=}1$ baseline (single-inference mean). $h = 1$ means full exploitation of the oracle gap; $h = 0$ matches $\mu$; $h < 0$ underperforms $\mu$. \colorbox[RGB]{56,142,60}{\textcolor{white}{green}} = positive (gain captured), \colorbox[RGB]{200,60,60}{\textcolor{white}{red}} = negative (below baseline). Bold when $|h| \geq 0.20$.}
\label{tab:headroom_olmo3_7b}
\resizebox{\textwidth}{!}{%
\begin{tabular}{l cccc cccc cccc}
  \toprule
  & \multicolumn{4}{c}{\textbf{HealthBench}} & \multicolumn{4}{c}{\textbf{PRBench}} & \multicolumn{4}{c}{\textbf{LEXam}} \\
  \cmidrule(lr){2-5} \cmidrule(lr){6-9} \cmidrule(lr){10-13}
  & \footnotesize XLow & \footnotesize Low & \footnotesize Mid & \footnotesize XHigh & \footnotesize XLow & \footnotesize Low & \footnotesize Mid & \footnotesize XHigh & \footnotesize XLow & \footnotesize Low & \footnotesize Mid & \footnotesize XHigh \\
  \midrule
  BoN+Skywork
    & \cellcolor[RGB]{196,221,197}+0.177
    & \cellcolor[RGB]{200,223,201}+0.165
    & \cellcolor[RGB]{201,224,202}+0.160
    & \cellcolor[RGB]{200,223,201}+0.165
    & \cellcolor[RGB]{198,222,199}+0.170
    & \cellcolor[RGB]{209,229,210}+0.137
    & \cellcolor[RGB]{215,232,216}+0.119
    & \cellcolor[RGB]{218,234,219}+0.110
    & \cellcolor[RGB]{183,214,184}\textbf{+0.216}
    & \cellcolor[RGB]{198,222,199}+0.171
    & \cellcolor[RGB]{204,226,205}+0.152
    & \cellcolor[RGB]{202,225,203}+0.157 \\
  Fusion
    & \cellcolor{white}---
    & \cellcolor[RGB]{186,216,187}\textbf{+0.207}
    & \cellcolor[RGB]{158,200,160}\textbf{+0.292}
    & \cellcolor{white}---
    & \cellcolor{white}---
    & \cellcolor[RGB]{235,243,235}+0.059
    & \cellcolor[RGB]{180,212,182}\textbf{+0.224}
    & \cellcolor{white}---
    & \cellcolor{white}---
    & \cellcolor[RGB]{137,188,140}\textbf{+0.353}
    & \cellcolor[RGB]{159,201,161}\textbf{+0.286}
    & \cellcolor{white}--- \\
  \bottomrule
\end{tabular}%
}
\vspace{0.4em}
\resizebox{\textwidth}{!}{%
\begin{tabular}{l cccc cccc cccc}
  \toprule
  & \multicolumn{4}{c}{\textbf{WildBench}} & \multicolumn{4}{c}{\textbf{WritingBench}} & \multicolumn{4}{c}{\textbf{Overall}} \\
  \cmidrule(lr){2-5} \cmidrule(lr){6-9} \cmidrule(lr){10-13}
  & \footnotesize XLow & \footnotesize Low & \footnotesize Mid & \footnotesize XHigh & \footnotesize XLow & \footnotesize Low & \footnotesize Mid & \footnotesize XHigh & \footnotesize XLow & \footnotesize Low & \footnotesize Mid & \footnotesize XHigh \\
  \midrule
  BoN+Skywork
    & \cellcolor[RGB]{178,211,180}\textbf{+0.230}
    & \cellcolor[RGB]{189,218,191}+0.196
    & \cellcolor[RGB]{194,220,195}+0.184
    & \cellcolor[RGB]{194,220,195}+0.182
    & \cellcolor[RGB]{202,225,203}+0.158
    & \cellcolor[RGB]{209,228,210}+0.138
    & \cellcolor[RGB]{213,231,214}+0.126
    & \cellcolor[RGB]{213,231,214}+0.125
    & \cellcolor[RGB]{191,219,193}+0.190
    & \cellcolor[RGB]{201,224,202}+0.161
    & \cellcolor[RGB]{205,227,206}+0.148
    & \cellcolor[RGB]{205,227,206}+0.148 \\
  Fusion
    & \cellcolor{white}---
    & \cellcolor[RGB]{237,193,193}-0.190
    & \cellcolor{white}---
    & \cellcolor{white}---
    & \cellcolor{white}---
    & \cellcolor[RGB]{227,239,228}+0.083
    & \cellcolor[RGB]{206,227,207}+0.145
    & \cellcolor{white}---
    & \cellcolor{white}---
    & \cellcolor[RGB]{221,235,221}+0.102
    & \cellcolor[RGB]{176,210,178}\textbf{+0.237}
    & \cellcolor{white}--- \\
  \bottomrule
\end{tabular}%
}
\end{table}


\section{Verifier Correlation and Headroom Capture by RM}
\label{app:rho_v_tables}

Tables~\ref{tab:rho_v_matrix} and~\ref{tab:h_matrix} report
$\hat{\rho}_v$ (denoised) and headroom capture $\hat{h}$ (ideal oracle) for every (generator, RM, benchmark) combination ($N=152$, 4 generators $\times$ 2 RMs $\times$ 19 tasks aggregated to benchmark level). Comparing the two tables cell-by-cell validates $\hat{h} \approx \hat{\rho}_v$ (Equation~\ref{eq:headroom_rho}) across all combinations
(regression: $\hat{h} = 1.198\,\hat{\rho}_v - 0.011$, $R^2=0.66$,
$\rho=0.81$, $p < 10^{-36}$).

\begin{table}[H]
\centering\small\setlength{\tabcolsep}{5pt}
\caption{Verifier correlation $\hat{\rho}_v$ (Spearman between RM scores and judge scores), averaged over tasks within each benchmark ($N=152$, 4 generators $\times$ 2 RMs $\times$ 19 tasks). Both RMs produce near-identical profiles (Skywork mean $\hat{\rho}_v=0.120$, Llama mean $\hat{\rho}_v=0.107$, computed at task level before benchmark aggregation), confirming the failure is structural rather than model-specific.}
\label{tab:rho_v_matrix}
\begin{tabular}{ll rrrrr}
\toprule
\textbf{Generator} & \textbf{RM} & \textbf{Health.} & \textbf{LEXam} & \textbf{PR} & \textbf{Wild} & \textbf{Writing} \\
\midrule
    Qwen3.5-9B & Skywork V2 & 0.063 & 0.045 & 0.064 & 0.255 & 0.197 \\
     & Llama-3.1-70B & 0.068 & 0.058 & 0.050 & 0.178 & 0.189 \\
    \addlinespace[3pt]
    Qwen3.5-35B & Skywork V2 & 0.036 & 0.028 & 0.023 & 0.260 & 0.220 \\
     & Llama-3.1-70B & 0.067 & 0.046 & 0.031 & 0.146 & 0.173 \\
    \addlinespace[3pt]
    Olmo-3-7B & Skywork V2 & 0.127 & 0.136 & 0.113 & 0.180 & 0.123 \\
     & Llama-3.1-70B & 0.119 & 0.154 & 0.101 & 0.166 & 0.095 \\
    \addlinespace[3pt]
    Olmo-3.1-32B & Skywork V2 & 0.078 & 0.053 & 0.082 & 0.177 & 0.251 \\
     & Llama-3.1-70B & 0.064 & 0.051 & 0.073 & 0.180 & 0.189 \\
    \addlinespace[3pt]
\bottomrule
\end{tabular}
\end{table}

\begin{table}[H]
\centering\small\setlength{\tabcolsep}{5pt}
\caption{Headroom capture $\hat{h}=(\text{BoN@16}-\mu)\,/\,(\text{Oracle@16}-\mu)$ for the same generator--RM--benchmark combinations. \textbf{Bold} negative values indicate the RM selects \emph{worse} than random. Comparing cell-by-cell with Table~\ref{tab:rho_v_matrix} confirms $\hat{h}\approx\hat{\rho}_v$ (regression: $\hat{h}=1.20\,\hat{\rho}_v-0.011$, $R^2=0.66$, $\rho=0.81$, $p<10^{-36}$), validating Equation~\ref{eq:headroom_rho}.}
\label{tab:h_matrix}
\begin{tabular}{ll rrrrr}
\toprule
\textbf{Generator} & \textbf{RM} & \textbf{Health.} & \textbf{LEXam} & \textbf{PR} & \textbf{Wild} & \textbf{Writing} \\
\midrule
    Qwen3.5-9B & Skywork V2 & 0.080 & 0.056 & 0.041 & 0.345 & 0.208 \\
     & Llama-3.1-70B & 0.092 & 0.102 & 0.015 & 0.182 & 0.182 \\
    \addlinespace[3pt]
    Qwen3.5-35B & Skywork V2 & 0.010 & 0.051 & 0.002 & 0.413 & 0.275 \\
     & Llama-3.1-70B & 0.056 & 0.090 & 0.046 & 0.243 & 0.253 \\
    \addlinespace[3pt]
    Olmo-3-7B & Skywork V2 & 0.139 & 0.157 & 0.119 & 0.182 & 0.123 \\
     & Llama-3.1-70B & 0.127 & 0.157 & 0.110 & 0.130 & 0.126 \\
    \addlinespace[3pt]
    Olmo-3.1-32B & Skywork V2 & 0.066 & 0.025 & 0.083 & 0.225 & 0.243 \\
     & Llama-3.1-70B & 0.059 & 0.123 & 0.090 & 0.246 & 0.232 \\
    \addlinespace[3pt]
\bottomrule
\end{tabular}
\end{table}

\section{Deterministic Benchmark Results: MATH-500 and GPQA Diamond}
\label{app:deterministic}

To contextualise our open-ended QA findings, we report results on two
standard deterministic benchmarks: \textbf{MATH-500}~\cite{hendrycks2021measuringmathematicalproblemsolving}
(competition mathematics, exact-match verified) and \textbf{GPQA
Diamond}~\cite{rein2023gpqagraduatelevelgoogleproofqa} (graduate-level science, multiple-choice).
These benchmarks admit binary verification, making them the canonical
setting where TTS was developed and where RM-based methods are expected
to perform well.

We evaluate the same generator (Qwen3.5-35B-A3B) and the same TTS
methods at matched compute levels, using the same token-budget
normalisation as the main experiments. This allows a direct comparison
of method rankings and headroom capture between deterministic and
open-ended settings.

\begin{table}[H]
\centering
\setlength{\tabcolsep}{2.5pt}
\tiny
\caption{Realised quality on \textbf{deterministic benchmarks} for
\textbf{Qwen3.5-35B-A3B}. Cell colour encodes delta from the
single-sample baseline (BoN@1):
\colorbox[RGB]{145,193,148}{\textcolor{white}{green}} = improvement,
\colorbox[RGB]{232,152,152}{\textcolor{white}{red}} = regression.
Bold = $|\Delta| \geq 0.03$. Italic Pass@K row reports the BoN-pool
oracle. Particle Filter and Budget Forcing on GPQA were not run.}
\label{tab:deterministic_qwen35b}
\resizebox{\textwidth}{!}{%
\begin{tabular}{l ccccc ccccc}
\toprule
  & \multicolumn{5}{c}{\textbf{MATH-500}} & \multicolumn{5}{c}{\textbf{GPQA Diamond}} \\
  \cmidrule(lr){2-6} \cmidrule(lr){7-11}
  & \footnotesize N=1 & \footnotesize Low & \footnotesize Mid & \footnotesize High & \footnotesize XHigh
  & \footnotesize N=1 & \footnotesize Low & \footnotesize Mid & \footnotesize High & \footnotesize XHigh \\
\midrule
  BoN+Skywork & \cellbase{0.922} & \cellcolor[RGB]{237,245,237}0.930 & \cellcolor[RGB]{230,241,231}0.933 & \cellcolor[RGB]{226,238,227}0.935 & \cellcolor[RGB]{224,237,225}0.936 & \cellbase{0.823} & \cellcolor[RGB]{215,232,216}0.841 & \cellcolor[RGB]{193,220,195}0.851 & \cellcolor[RGB]{191,219,192}0.852 & \cellcolor[RGB]{210,230,212}0.843 \\
  BoN+Llama70B & \cellbase{0.922} & \cellcolor[RGB]{239,246,240}0.929 & \cellcolor[RGB]{232,242,233}0.932 & \cellcolor[RGB]{228,240,229}0.934 & \cellcolor[RGB]{219,235,220}0.938 & \cellbase{0.823} & \cellcolor[RGB]{226,238,227}0.836 & \cellcolor[RGB]{213,231,214}0.842 & \cellcolor[RGB]{215,232,216}0.841 & \cellcolor[RGB]{221,236,222}0.838 \\
  Fusion & \cellcolor{white}--- & \cellcolor{white}--- & \cellcolor[RGB]{224,237,225}0.936 & \cellcolor[RGB]{228,240,229}0.934 & \cellcolor[RGB]{232,242,233}0.932 & \cellcolor{white}--- & \cellcolor{white}--- & \cellcolor[RGB]{175,210,177}\textbf{0.859} & \cellcolor[RGB]{232,242,233}0.833 & \cellcolor[RGB]{255,255,255}0.823 \\
  Beam Search & \cellcolor{white}--- & \cellcolor{white}--- & \cellcolor[RGB]{211,230,212}0.942 & \cellcolor[RGB]{215,232,216}0.940 & \cellcolor[RGB]{228,240,229}0.934 & \cellcolor{white}--- & \cellcolor{white}--- & \cellcolor[RGB]{252,244,244}0.818 & \cellcolor[RGB]{245,213,213}0.803 & \cellcolor[RGB]{252,244,244}0.818 \\
  Particle Filter & \cellcolor{white}--- & \cellcolor{white}--- & \cellcolor[RGB]{250,234,234}0.912 & \cellcolor[RGB]{241,193,193}\textbf{0.892} & \cellcolor[RGB]{240,189,189}\textbf{0.890} & \cellcolor{white}--- & \cellcolor{white}--- & \cellcolor{white}--- & \cellcolor{white}--- & \cellcolor{white}--- \\
  SR & \cellcolor{white}--- & \cellcolor[RGB]{241,247,242}0.928 & \cellcolor[RGB]{237,245,237}0.930 & \cellcolor[RGB]{241,247,242}0.928 & \cellcolor[RGB]{224,237,225}0.936 & \cellcolor{white}--- & \cellcolor[RGB]{244,248,244}0.828 & \cellcolor[RGB]{255,255,255}0.823 & \cellcolor[RGB]{243,203,203}0.798 & \cellcolor[RGB]{221,236,222}0.838 \\
  Budget Forcing & \cellcolor{white}--- & \cellcolor[RGB]{237,245,237}0.930 & \cellcolor[RGB]{219,235,220}0.938 & \cellcolor{white}--- & \cellcolor{white}--- & \cellcolor{white}--- & \cellcolor{white}--- & \cellcolor{white}--- & \cellcolor{white}--- & \cellcolor{white}--- \\
\midrule
  \textit{BoN Oracle (Pass@K)} & \textit{0.922} & \textit{0.935} & \textit{0.942} & \textit{0.948} & \textit{0.954} & \textit{0.823} & \textit{0.870} & \textit{0.909} & \textit{0.938} & \textit{0.960} \\
\bottomrule
\end{tabular}%
}
\end{table}


\paragraph{Headroom is benchmark-fixed.}
Qwen3.5-35B-A3B already reaches the single-sample pass@1 of $0.922$ on MATH-500 and $0.823$ on GPQA Diamond, leaving only
$3.2$\,pp and $13.7$\,pp of headroom up to oracle Pass@16. RM-based BoN at XHigh captures $50\%$ of available headroom
on MATH-500 and $15\%$ on GPQA, versus ${\sim}15\%$ performed on open-ended QA
(Section~\ref{sec:rm_analysis}). Appendix~\ref{fig:skywork-test} reproduces this scaling with the Skywork-RM paper's
much weaker reference generator (single-sample ${\sim}0.42$ on both benchmarks), where BoN+Skywork captures
${\sim}43\%$ of headroom on MATH and ${\sim}22\%$ on GPQA --- within a few points of our $50\%/15\%$. RM efficacy is
therefore approximately benchmark-fixed: stronger generators shrink the absolute headroom rather than improve the rate
at which RMs can exploit it.
\section{Skywork-Reward-V2 Reproduction on Deterministic Benchmarks}
\label{app:skywork_reproduction}

We verified
that our BoN pipeline reproduces
the scaling behaviour reported in the original Skywork-Reward-V2
paper~\cite{skywork2024reward} with the ORM model \textbf{Skywork-Reward-V2-Llama-3.1-8B}. We run Best-of-$N$ sampling with
PPE Correctness scoring on \textbf{MATH-500} and \textbf{GPQA}, matching the authors' evaluation protocol.

\begin{figure}[H]
    \centering
    \includegraphics[width=\textwidth]{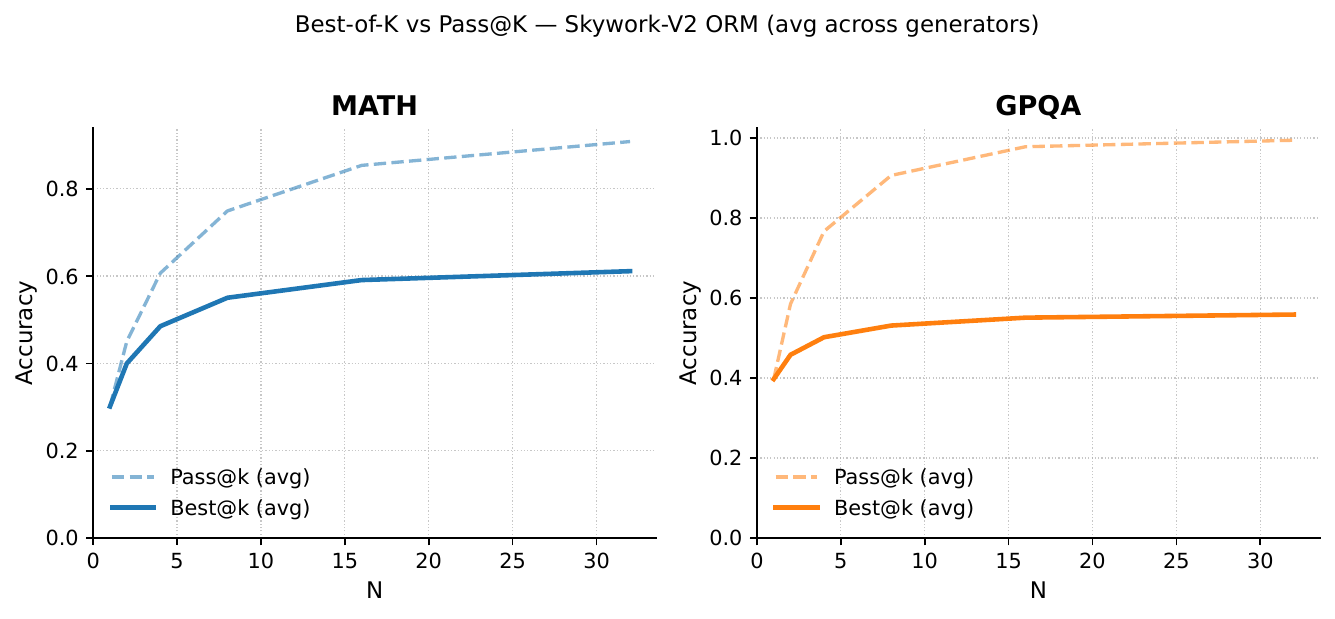}
    \caption{Reproduction of BoN scaling from the Skywork-Reward-V2
    paper~\cite{skywork2024reward} on MATH-500 (left) and GPQA
    Diamond (right). Dashed grey: oracle accuracy. Solid coloured:
    realised accuracy under Skywork-Reward-V2-Llama-3.1-8B
    selection. Results match the authors' reported curves, validating
    our BoN pipeline before deployment on open-ended benchmarks.}
    \label{fig:skywork-test}
\end{figure}

Figure~\ref{fig:skywork-test} shows realised accuracy (solid lines)
and oracle accuracy (dashed) as a function of $N$. Our results
replicate the paper's findings: realised accuracy scales positively
with $N$ on both benchmarks, reaching ${\sim}0.66$ on MATH-500 and
${\sim}0.55$ on GPQA at $N{=}32$, consistent with the
curves reported for Skywork-Reward-V2-Llama-3.1-8B in Figure~4
of~\citet{skywork2024reward}.

The large gap between oracle and realised quality visible in both
panels, the oracle approaching $1.0$ while realised quality
plateaus well below $0.7$, already hints at the exploitation
bottleneck we characterise in the main text. Here, however, the
RM is at least positively correlated with correctness ($\hat\rho_v
> 0$), so realised quality does improve with compute, unlike the
near-zero correlations we observe on open-ended QA
(Section~\ref{sec:rm_analysis}).

\section{WildBench Per-Category Decomposition}
\label{app:wildbench_coding}

Sequential Refinement's aggregate gains on WildBench mask extreme heterogeneity
across task categories. We decompose the total realised lift (sum of
per-item score changes from the initial to the final draft) by
\texttt{primary\_tag}.

Coding~\&~Debugging accounts for 16.6\% of items (170/1{,}024) but
contributes 105.5\% of the total realised gain: a mean lift of
$+27.06$ per item, compared with $-0.28$ per item across the
remaining 854 items. Excluding Coding~\&~Debugging, Sequential Refinement
produces a net regression on WildBench. The aggregate improvement
is therefore not a general property of iterative refinement on
open-ended chat tasks, but is driven almost entirely by a single
near-deterministic subtask where successive drafts converge toward
correct code.

\begin{figure}[H]
    \centering
    \includegraphics[width=0.48\textwidth]{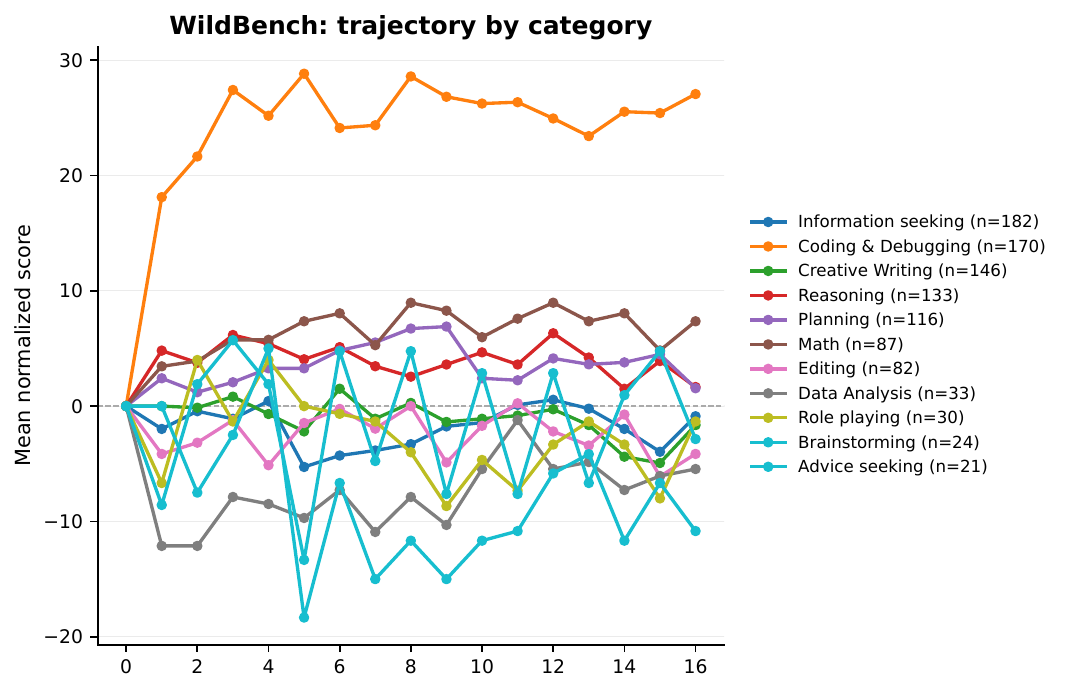}
    \hfill
    \includegraphics[width=0.48\textwidth]{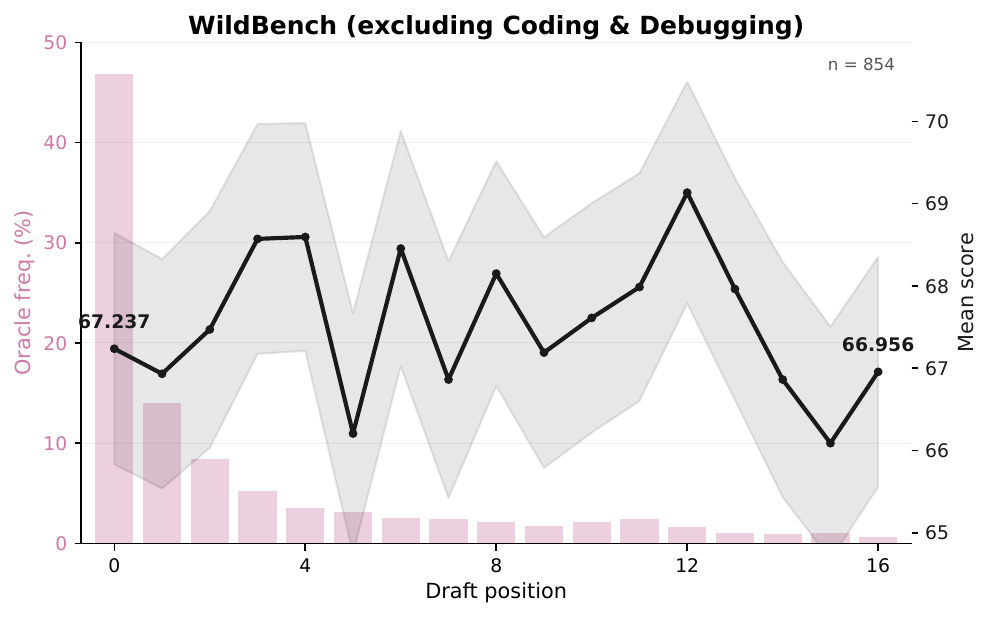}
    \caption{\textbf{Left:} Mean score trajectory across Sequential Refinement
    iterations per WildBench task category. Coding~\&~Debugging rises
    steeply; most other categories are flat or decline.
    \textbf{Right:} Oracle and mean score trajectories on WildBench
    with Coding~\&~Debugging excluded. Without this subtask, mean
    quality regresses across iterations while the oracle continues
    to rise---confirming the exploitation failure pattern observed
    on other benchmarks.}
    \label{fig:wildbench_coding}
\end{figure}

\section{Verbosity Analysis}
\label{app:verbosity}

Response length is not a neutral quantity in open-ended evaluation:
if the judge rewards longer responses, methods that produce progressively
longer outputs will appear to improve even when content quality is flat or
declining. This appendix documents the verbosity bias in two steps.
Section~\ref{app:verbosity_unified} quantifies the length--score
relationship under our unified judge (Qwen3.5-397B-A17B).
Section~\ref{app:verbosity_native} shows that the WritingBench native
judge amplifies this bias substantially, confirming it is a property of
the benchmark's evaluation design rather than of any particular judge.

\subsection{Unified Judge: Length--Score Correlation}
\label{app:verbosity_unified}

\begin{figure}[H]
    \centering
    \includegraphics[width=\textwidth]{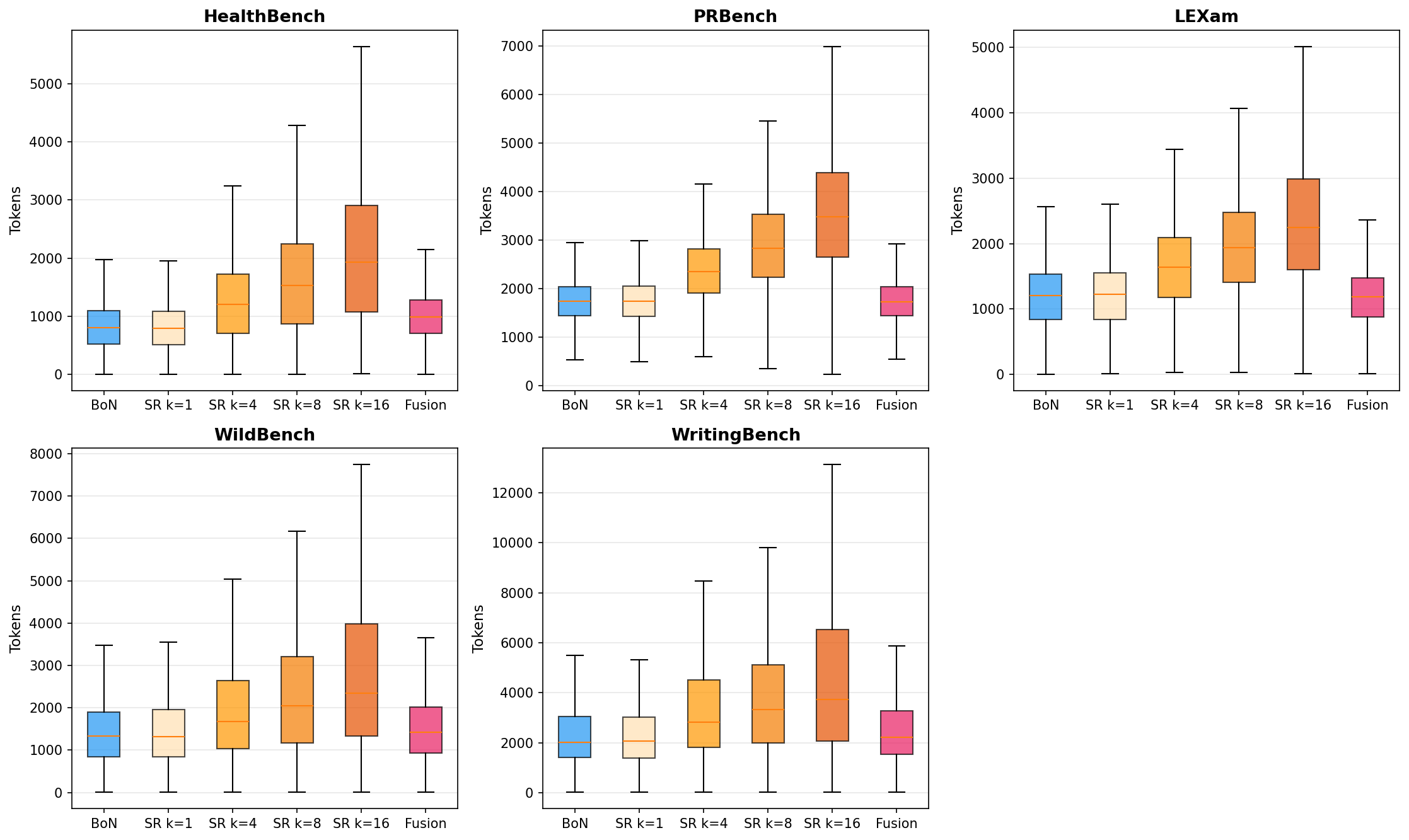}
    \caption{Token length distributions per method and benchmark for
    Qwen3.5-35B-A3B. \textbf{BoN} (blue) and \textbf{Fusion} (pink)
    show stable length distributions across compute levels.
    \textbf{Sequential Refinement} (orange gradient, SR $k{=}1$ through $k{=}16$)
    shows monotonically increasing median length on every benchmark,
    with the steepest growth on WritingBench and PRBench.}
    \label{fig:verbosity-token-length}
\end{figure}

The token length distributions in Figure~\ref{fig:verbosity-token-length}
reveal a structural difference between Sequential-Refinement and all other methods:
response length grows monotonically across iterations on every benchmark.
This matters because if the judge rewards length, Sequential Refinement's apparent
gains on WritingBench reflect elaboration rather than quality improvement.

Tables~\ref{tab:verbosity_spearman_bon}--\ref{tab:verbosity_cvratio_sr}
quantify the relationship between length and judge score within prompts,
separately for BoN and Sequential Refinement. The key findings are:

\begin{itemize}
    \item \textbf{WritingBench} shows the strongest within-prompt
    length--score correlation under both methods ($\hat{\rho} = +0.198$
    for BoN, $+0.352$ for Sequential Refinement), and is the only benchmark where
    CV\textsubscript{len} exceeds CV\textsubscript{score} under
    Sequential Refinement (ratio $= 0.36$). Length growth across iterations is
    the primary driver of score gains on this benchmark.

    \item \textbf{HealthBench and PRBench} are content-driven:
    CV\textsubscript{score} substantially exceeds CV\textsubscript{len}
    under BoN (ratios of $1.57$ and $3.17$ respectively). Sequential Refinement's
    regressions on these benchmarks reflect genuine quality degradation,
    not a scoring artifact.

    \item \textbf{LEXam} is the only benchmark where Sequential Refinement shows
    a \emph{negative} length--score correlation
    ($\hat{\rho} = -0.060$, $p < 10^{-3}$), consistent with iterative
    conditioning producing longer but worse responses on
    knowledge-intensive legal tasks.
\end{itemize}

\begin{table}[H]
\centering
\small
\caption{Within-prompt Spearman rank correlation between response length
and judge score for \textbf{BoN} candidates (16 per prompt),
Qwen3.5-35B-A3B, unified judge.
$\hat{\rho}$ = mean per-prompt Spearman correlation;
95\% CI is bootstrapped;
\%$\rho{>}0$ = fraction of prompts with a positive correlation;
$p$-values from a one-sample $t$-test and Wilcoxon signed-rank test
against $H_0\colon\rho=0$.}
\label{tab:verbosity_spearman_bon}
\begin{tabular}{lrr@{\;}lrcr}
\toprule
\textbf{Benchmark} & \textbf{$n$} &
\multicolumn{2}{c}{$\hat{\rho}$ [95\% CI]} &
\textbf{\%$\rho{>}0$} &
\textbf{$t$-test $p$} &
\textbf{Wilcoxon $p$} \\
\midrule
HealthBench  & 4{,}619 & $+$0.133 & [$+$0.124, $+$0.142] & 67.5\%
             & $7.76\times10^{-180}$ & $8.20\times10^{-167}$ \\
PRBench      & 1{,}634 & $+$0.112 & [$+$0.099, $+$0.126] & 64.7\%
             & $6.07\times10^{-57}$  & $1.87\times10^{-51}$  \\
LEXam        &   502   & $+$0.014 & [$-$0.011, $+$0.039] & 52.0\%
             & $2.73\times10^{-1}$   & $3.81\times10^{-1}$   \\
WildBench    &   917   & $+$0.044 & [$+$0.024, $+$0.063] & 57.6\%
             & $1.93\times10^{-5}$   & $4.72\times10^{-6}$   \\
WritingBench &   554   & $+$0.198 & [$+$0.174, $+$0.222] & 74.5\%
             & $4.97\times10^{-46}$  & $4.10\times10^{-39}$  \\
\bottomrule
\end{tabular}
\end{table}

\begin{table}[H]
\centering
\small
\caption{Within-prompt length variance for \textbf{BoN} candidates
(16 per prompt), Qwen3.5-35B-A3B.
CV = coefficient of variation (std/mean) of token lengths within a
prompt; \%CV${<}0.1$ and \%CV${<}0.2$ report the fraction of prompts
with low length dispersion; mean and median std are in tokens.
PRBench shows the lowest length variance (mean CV $= 0.083$), consistent
with its structured rubric format constraining response length.
WritingBench and WildBench show the highest variance, reflecting
open-ended prompts that admit responses of widely varying length.}
\label{tab:verbosity_cv_bon}
\begin{tabular}{lrrrrrrr}
\toprule
\textbf{Benchmark} & \textbf{$n$} &
\textbf{mean CV} & \textbf{med CV} &
\textbf{\%CV${<}0.1$} & \textbf{\%CV${<}0.2$} &
\textbf{mean std} & \textbf{med std} \\
\midrule
HealthBench  & 5{,}000 & 0.150 & 0.127 & 29.7\% & 83.2\% & 101.3 &  88.0 \\
PRBench      & 1{,}650 & 0.083 & 0.076 & 81.2\% & 98.6\% & 141.6 & 128.8 \\
LEXam        &   516   & 0.119 & 0.096 & 54.3\% & 90.9\% & 113.5 & 110.7 \\
WildBench    & 1{,}023 & 0.186 & 0.131 & 30.2\% & 73.4\% & 295.2 & 160.7 \\
WritingBench &   555   & 0.145 & 0.107 & 45.2\% & 76.2\% & 392.7 & 244.0 \\
\bottomrule
\end{tabular}
\end{table}

\begin{table}[H]
\centering
\small
\caption{Score variability versus length variability for \textbf{BoN}
candidates (16 per prompt), Qwen3.5-35B-A3B, unified judge.
CV\textsubscript{score} and CV\textsubscript{len} are the per-prompt
coefficient of variation of judge scores and token lengths respectively;
mean ratio $=$ mean(CV\textsubscript{score}/CV\textsubscript{len}).
A ratio ${>}1$ indicates score variation is driven by content;
a ratio ${<}1$ indicates length is the dominant axis of variation.
PRBench (ratio $= 3.17$) and HealthBench (ratio $= 1.57$) are
content-driven. WritingBench (ratio $= 0.55$) and WildBench
(ratio $= 0.69$) are length-driven even under BoN, before any
iterative elaboration.}
\label{tab:verbosity_cvratio_bon}
\begin{tabular}{lrrrrrr}
\toprule
\textbf{Benchmark} & \textbf{$n$} &
\textbf{CV\textsubscript{score}} &
\textbf{CV\textsubscript{len}} &
\textbf{mean ratio} &
\textbf{\%ratio${>}2$} &
\textbf{\%both${<}0.05$} \\
\midrule
HealthBench  & 5{,}000 & 0.329 & 0.150 & 1.57 & 37.8\% & 0.1\% \\
PRBench      & 1{,}650 & 0.356 & 0.083 & 3.17 & 76.5\% & 0.2\% \\
LEXam        &   516   & 0.217 & 0.119 & 1.65 & 39.1\% & 0.0\% \\
WildBench    & 1{,}023 & 0.137 & 0.186 & 0.69 &  8.0\% & 0.4\% \\
WritingBench &   555   & 0.084 & 0.145 & 0.55 &  5.2\% & 1.4\% \\
\bottomrule
\end{tabular}
\end{table}

\begin{table}[H]
\centering
\small
\caption{Within-prompt Spearman rank correlation between response length
and judge score for \textbf{Sequential Refinement} drafts ($k=1,\ldots,16$ per
prompt), Qwen3.5-35B-A3B, unified judge. Columns as in
Table~\ref{tab:verbosity_spearman_bon}. Compared with BoN, Sequential Refinement
shows a substantially stronger length--score correlation on WritingBench
($+0.352$ vs.\ $+0.198$) and PRBench ($+0.252$ vs.\ $+0.112$),
confirming that iterative elaboration amplifies the length--score
relationship. LEXam is the only benchmark with a significantly
\emph{negative} correlation ($\hat{\rho}=-0.060$), consistent with
refinement producing longer but worse responses on knowledge-intensive
tasks.}
\label{tab:verbosity_spearman_sr}
\begin{tabular}{lrr@{\;}lrcr}
\toprule
\textbf{Benchmark} & \textbf{$n$} &
\multicolumn{2}{c}{$\hat{\rho}$ [95\% CI]} &
\textbf{\%$\rho{>}0$} &
\textbf{$t$-test $p$} &
\textbf{Wilcoxon $p$} \\
\midrule
HealthBench  & 4{,}703 & $+$0.059 & [$+$0.047, $+$0.069] & 56.2\%
             & $1.26\times10^{-23}$  & $2.94\times10^{-23}$  \\
PRBench      & 1{,}634 & $+$0.252 & [$+$0.232, $+$0.272] & 72.2\%
             & $6.68\times10^{-116}$ & $1.75\times10^{-99}$  \\
LEXam        &   505   & $-$0.060 & [$-$0.092, $-$0.026] & 41.2\%
             & $2.46\times10^{-4}$   & $2.29\times10^{-4}$   \\
WildBench    &   925   & $+$0.023 & [$-$0.001, $+$0.046] & 52.0\%
             & $5.23\times10^{-2}$   & $3.82\times10^{-2}$   \\
WritingBench &   555   & $+$0.352 & [$+$0.319, $+$0.385] & 80.7\%
             & $1.12\times10^{-73}$  & $1.21\times10^{-55}$  \\
\bottomrule
\end{tabular}
\end{table}

\begin{table}[H]
\centering
\small
\caption{Score variability versus length variability for
\textbf{Sequential Refinement} drafts (16 iterations per prompt),
Qwen3.5-35B-A3B, unified judge. Columns as in
Table~\ref{tab:verbosity_cvratio_bon}. Compared with BoN, Sequential Refinement
shows higher CV\textsubscript{len} on every benchmark, reflecting
monotonic length growth across iterations. On WritingBench,
CV\textsubscript{len} substantially exceeds CV\textsubscript{score}
(ratio $= 0.36$), confirming that length growth is the primary driver
of score gains. On HealthBench, the ratio drops from $1.57$ (BoN) to
$0.91$ (Sequential Refinement): iterative conditioning increases length variance
more than score variance, a sign of elaboration without quality
improvement.}
\label{tab:verbosity_cvratio_sr}
\begin{tabular}{lrrrrrr}
\toprule
\textbf{Benchmark} & \textbf{$n$} &
\textbf{CV\textsubscript{score}} &
\textbf{CV\textsubscript{len}} &
\textbf{mean ratio} &
\textbf{\%ratio${>}2$} &
\textbf{\%both${<}0.05$} \\
\midrule
HealthBench  & 5{,}000 & 0.358 & 0.257 & 0.91 & 19.3\% & 0.0\% \\
PRBench      & 1{,}650 & 0.297 & 0.256 & 0.84 & 15.6\% & 0.0\% \\
LEXam        &   516   & 0.237 & 0.218 & 0.86 & 13.2\% & 0.0\% \\
WildBench    & 1{,}024 & 0.126 & 0.256 & 0.46 &  5.8\% & 0.2\% \\
WritingBench &   555   & 0.083 & 0.198 & 0.36 &  2.3\% & 0.5\% \\
\bottomrule
\end{tabular}
\end{table}

\subsection{Native WritingBench Judge: Bias Amplification}
\label{app:verbosity_native}

The unified judge already reveals a strong verbosity bias on WritingBench.
To confirm this is a property of the benchmark's evaluation design and not
of our judge choice, we repeat the analysis using the WritingBench native
judge, \textbf{WritingBench-Critic-Model-Qwen-7B}~\cite{wu2025writingbench},
a Qwen2.5-7B-Instruct model fine-tuned on 50K writing evaluation examples.
Wu et al.\ report $83\%$ human agreement for this critic model on a 300-query, 5-annotator pairwise evaluation~\cite{wu2025writingbench}.

The native judge amplifies the verbosity signal substantially on
WritingBench while leaving all other benchmarks unchanged:
the BoN Spearman correlation rises from $+0.198$ to $+0.370$;
the Sequential Refinement correlation rises from $+0.352$ to $+0.539$;
CV\textsubscript{score} on WritingBench collapses from $0.084$ to $0.034$
(BoN) and from $0.083$ to $0.038$ (Sequential Refinement), meaning the native judge
discriminates on length. These results suggest strongly that
the verbosity bias is intrinsic to WritingBench's evaluation design,
not an artefact of our unified judge.

\begin{table}[H]
\centering
\small
\caption{Within-prompt Spearman rank correlation between response length
and judge score for \textbf{BoN} candidates (16 per prompt),
Qwen3.5-35B-A3B, \textbf{native WritingBench judge}.
WritingBench $\hat{\rho}$ rises from $+0.198$ (unified judge,
Table~\ref{tab:verbosity_spearman_bon}) to $+0.370$; all other
benchmarks are stable, confirming the amplification is
WritingBench-specific.}
\label{tab:wb_judge_spearman_bon}
\begin{tabular}{lrr@{\;}lrcr}
\toprule
\textbf{Benchmark} & \textbf{$n$} &
\multicolumn{2}{c}{$\hat{\rho}$ [95\% CI]} &
\textbf{\%$\rho{>}0$} &
\textbf{$t$-test $p$} &
\textbf{Wilcoxon $p$} \\
\midrule
HealthBench  & 4{,}619 & $+$0.133 & [$+$0.124, $+$0.142] & 67.5\%
             & $7.76\times10^{-180}$ & $8.20\times10^{-167}$ \\
PRBench      & 1{,}634 & $+$0.112 & [$+$0.099, $+$0.125] & 64.7\%
             & $6.07\times10^{-57}$  & $1.87\times10^{-51}$  \\
LEXam        &   502   & $+$0.014 & [$-$0.010, $+$0.038] & 52.0\%
             & $2.73\times10^{-1}$   & $3.81\times10^{-1}$   \\
WildBench    &   917   & $+$0.044 & [$+$0.023, $+$0.064] & 57.6\%
             & $1.93\times10^{-5}$   & $4.72\times10^{-6}$   \\
WritingBench &   555   & $+$0.370 & [$+$0.347, $+$0.394] & 87.9\%
             & $6.24\times10^{-121}$ & $7.28\times10^{-77}$  \\
\bottomrule
\end{tabular}
\end{table}

\begin{table}[H]
\centering
\small
\caption{Score variability versus length variability for \textbf{BoN}
candidates (16 per prompt), Qwen3.5-35B-A3B, \textbf{native
WritingBench judge}. Under the native judge, WritingBench
CV\textsubscript{score} collapses to $0.034$ (vs.\ $0.084$ under the
unified judge, Table~\ref{tab:verbosity_cvratio_bon}), yielding a ratio
of $0.25$: score variation is almost entirely explained by length.
Length statistics are identical to Table~\ref{tab:verbosity_cv_bon}
since they are judge-independent.}
\label{tab:wb_judge_cvratio_bon}
\begin{tabular}{lrrrrrr}
\toprule
\textbf{Benchmark} & \textbf{$n$} &
\textbf{CV\textsubscript{score}} &
\textbf{CV\textsubscript{len}} &
\textbf{mean ratio} &
\textbf{\%ratio${>}2$} &
\textbf{\%both${<}0.05$} \\
\midrule
HealthBench  & 5{,}000 & 0.329 & 0.150 & 1.57 & 37.8\% & 0.1\% \\
PRBench      & 1{,}650 & 0.356 & 0.083 & 3.17 & 76.5\% & 0.2\% \\
LEXam        &   516   & 0.217 & 0.119 & 1.65 & 39.1\% & 0.0\% \\
WildBench    & 1{,}023 & 0.137 & 0.186 & 0.69 &  8.0\% & 0.4\% \\
WritingBench &   555   & 0.034 & 0.145 & 0.25 &  0.0\% & 2.0\% \\
\bottomrule
\end{tabular}
\end{table}

\begin{table}[H]
\centering
\small
\caption{Within-prompt Spearman rank correlation between response length
and judge score for \textbf{Sequential Refinement} drafts ($k=1,\ldots,16$ per
prompt), Qwen3.5-35B-A3B, \textbf{native WritingBench judge}.
WritingBench $\hat{\rho}$ rises to $+0.539$ (vs.\ $+0.352$ under the
unified judge, Table~\ref{tab:verbosity_spearman_sr}), with 92.2\% of
prompts showing a positive correlation. All other benchmarks are stable.}
\label{tab:wb_judge_spearman_sr}
\begin{tabular}{lrr@{\;}lrcr}
\toprule
\textbf{Benchmark} & \textbf{$n$} &
\multicolumn{2}{c}{$\hat{\rho}$ [95\% CI]} &
\textbf{\%$\rho{>}0$} &
\textbf{$t$-test $p$} &
\textbf{Wilcoxon $p$} \\
\midrule
HealthBench  & 4{,}703 & $+$0.059 & [$+$0.047, $+$0.070] & 56.2\%
             & $1.26\times10^{-23}$  & $2.94\times10^{-23}$  \\
PRBench      & 1{,}634 & $+$0.252 & [$+$0.232, $+$0.272] & 72.2\%
             & $6.68\times10^{-116}$ & $1.75\times10^{-99}$  \\
LEXam        &   505   & $-$0.060 & [$-$0.091, $-$0.029] & 41.2\%
             & $2.46\times10^{-4}$   & $2.29\times10^{-4}$   \\
WildBench    &   925   & $+$0.023 & [$+$0.000, $+$0.045] & 52.0\%
             & $5.23\times10^{-2}$   & $3.82\times10^{-2}$   \\
WritingBench &   551   & $+$0.539 & [$+$0.513, $+$0.566] & 92.2\%
             & $1.07\times10^{-159}$ & $1.82\times10^{-82}$  \\
\bottomrule
\end{tabular}
\end{table}

\begin{table}[H]
\centering
\small
\caption{Score variability versus length variability for
\textbf{Sequential Refinement} drafts (16 iterations per prompt),
Qwen3.5-35B-A3B, \textbf{native WritingBench judge}.
WritingBench CV\textsubscript{score} $= 0.038$ (vs.\ $0.083$ under the
unified judge, Table~\ref{tab:verbosity_cvratio_sr}), with a ratio of
$0.17$: under the native judge, virtually all score variation in
Sequential Refinement on WritingBench is attributable to response length.}
\label{tab:wb_judge_cvratio_sr}
\begin{tabular}{lrrrrrr}
\toprule
\textbf{Benchmark} & \textbf{$n$} &
\textbf{CV\textsubscript{score}} &
\textbf{CV\textsubscript{len}} &
\textbf{mean ratio} &
\textbf{\%ratio${>}2$} &
\textbf{\%both${<}0.05$} \\
\midrule
HealthBench  & 5{,}000 & 0.358 & 0.257 & 0.91 & 19.3\% & 0.0\% \\
PRBench      & 1{,}650 & 0.297 & 0.256 & 0.84 & 15.6\% & 0.0\% \\
LEXam        &   516   & 0.237 & 0.218 & 0.86 & 13.2\% & 0.0\% \\
WildBench    & 1{,}024 & 0.126 & 0.256 & 0.46 &  5.8\% & 0.2\% \\
WritingBench &   555   & 0.038 & 0.198 & 0.17 &  0.5\% & 0.4\% \\
\bottomrule
\end{tabular}
\end{table}

\subsection{WritingBench: Judge Robustness Analysis}
\label{app:writingbench_judge}

\begin{figure}[h]
    \centering
    \includegraphics[width=\linewidth]{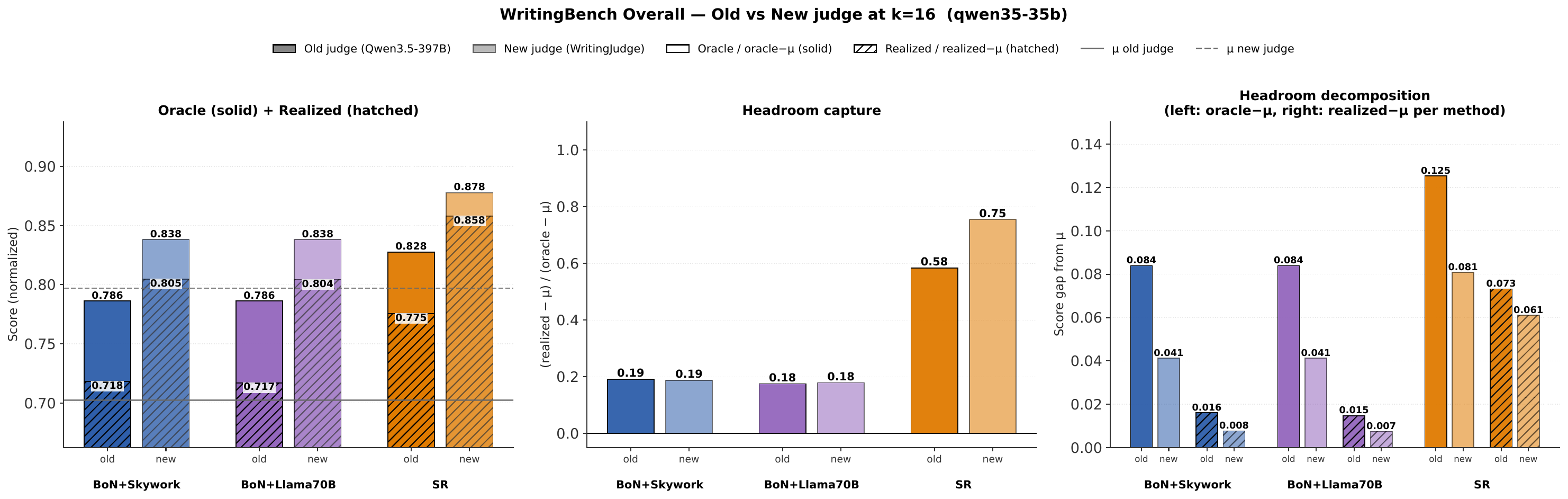}
    \caption{\textbf{WritingBench old vs.\ new judge at $k{=}16$ (Qwen3.5-35B, micro-averaged across 555 samples).}
    \emph{Left:} Oracle (solid) and realized (hatched) scores for BoN+Skywork, BoN+Llama70B,
    and Self-Refine under our judge (Qwen3.5-397B, dark) and the authors' fine-tuned
    WritingBench-Critic-Model-Qwen-7B (light); horizontal lines show the candidate-mean $\mu$ for each judge.
    \emph{Center:} Headroom capture $(\text{realized}-\mu)/(\text{oracle}-\mu)$.
    \emph{Right:} Decomposition into denominator (oracle$-\mu$, solid) and numerator
    (realized$-\mu$, hatched).}
    \label{fig:writingbench_judge_robustness}
\end{figure}

To assess whether our WritingBench judge agreement (QWK $0.408$, Section~\ref{sec:setup})
materially affects our conclusions, we re-score the same TTS outputs with the official fine-tuned Critic model released by the WritingBench
authors~\cite{wu2025writingbench} and recompute realized and oracle scores under both judges (Figure~\ref{fig:writingbench_judge_robustness}). For BoN with either RM (Skywork, Llama70B), headroom capture is judge-invariant ($0.19$ vs.\ $0.19$), indicating that conclusions about parallel selection methods transfer cleanly across
judges. For Sequential Refinement the capture ratio rises from $0.58$ to $0.75$ under Critic model, but the decomposition shows this is driven by a shrinking denominator
(oracle$-\mu$ drops from $0.125$ to $0.081$) rather than an improved numerator (realized$-\mu$ actually \emph{decreases} slightly, from $0.073$ to $0.061$). This
denominator compression is the mechanical signature of Critic model's stronger length bias (Appendix~\ref{app:verbosity}) interacting with SR's
length-growing trajectory: length-favored late iterations (both realized and oracle)
are pulled closer together, inflating the ratio without genuine selection
improvement. The result confirms that our methods-level conclusions are stable
across judges where it matters, and where they diverge the divergence is explained
by a known structural bias of the benchmark rather than by miscalibration of our
judge.

\section{Effect of Model Capacity on Self-Verifier Algorithms}
\label{sec:model_capacity_study}

\begin{figure}[H]
    \centering
    \includegraphics[width=\textwidth]{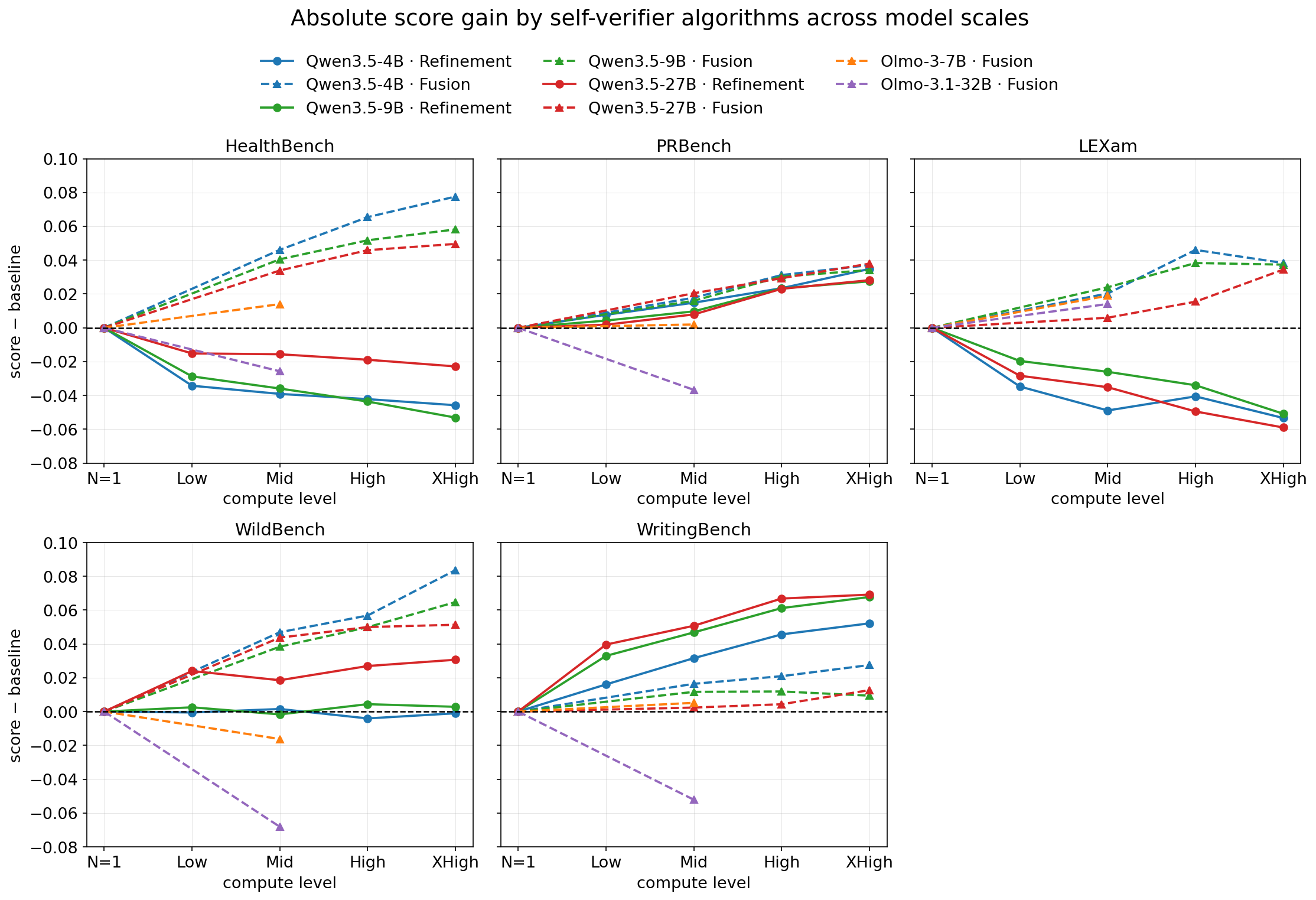}
    \caption{Absolute score gain over the BoN@1 baseline across compute
    levels, by model family and benchmark.}
    \label{fig:self-verifier-abs-gain-per-bench}
\end{figure}

\begin{figure}[H]
    \centering
    \includegraphics[width=\textwidth]{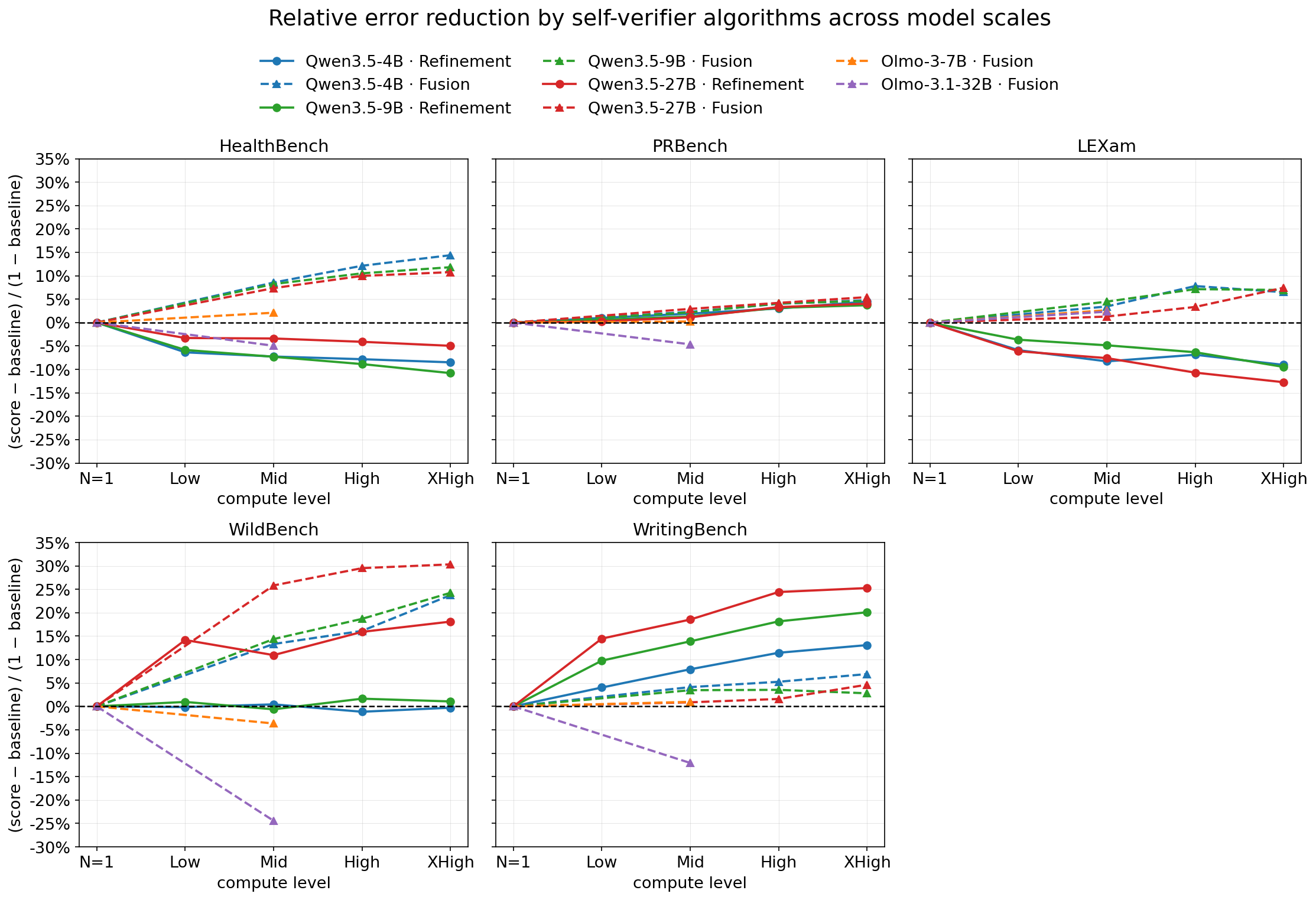}
    \caption{Fraction of remaining gap to a perfect score,
    $(p_k - p_1)/(1 - p_1)$, closed at compute level $k$, faceted by
    benchmark. $p_1$ is the BoN@1 baseline.}
    \label{fig:self-verifier-headroom-gain-per-bench}
\end{figure}

We study self-verifier methods along two axes: compute and model capacity. The model-capacity axis spans the Qwen3.5 dense instruct family at three sizes --- 4B, 9B, and 27B ---, along with Olmo3-7b and Olmo3.1-32b where only fusion was run, holding all
   other settings fixed. We first report the mean absolute score gain on each benchmark in Figure~\ref{fig:self-verifier-abs-gain-per-bench}. Because     
  moving from $0.80 \to 0.85$ is arguably harder than $0.60 \to 0.65$, we also report the relative error reduction --- the fraction of the gap between a BoN@1 baseline and a perfect score that
  is closed --- in Figure~\ref{fig:self-verifier-headroom-gain-per-bench}. Both metrics point to the same conclusions:           
  \begin{enumerate}
      \item \textbf{Fusion ability is model-family-dependent}: Qwen3.5 shows fusion consistently helping across benchmarks, with gains that are roughly stable across model scales. Increasing compute further improves performance monotonically. However, Olmo3.1-32b fails badly with Olmo3-7b still struggling to hit the BoN@1 baseline. 
      \item \textbf{Sequential Refinement performance is benchmark-dependent:} it yields clear gains on 2/5 benchmarks but strictly hurts performance on 2/5. There are no consistent model capacity scaling trends.                                 
  \end{enumerate}

\begin{figure}[H]
    \centering
    \includegraphics[width=\textwidth]{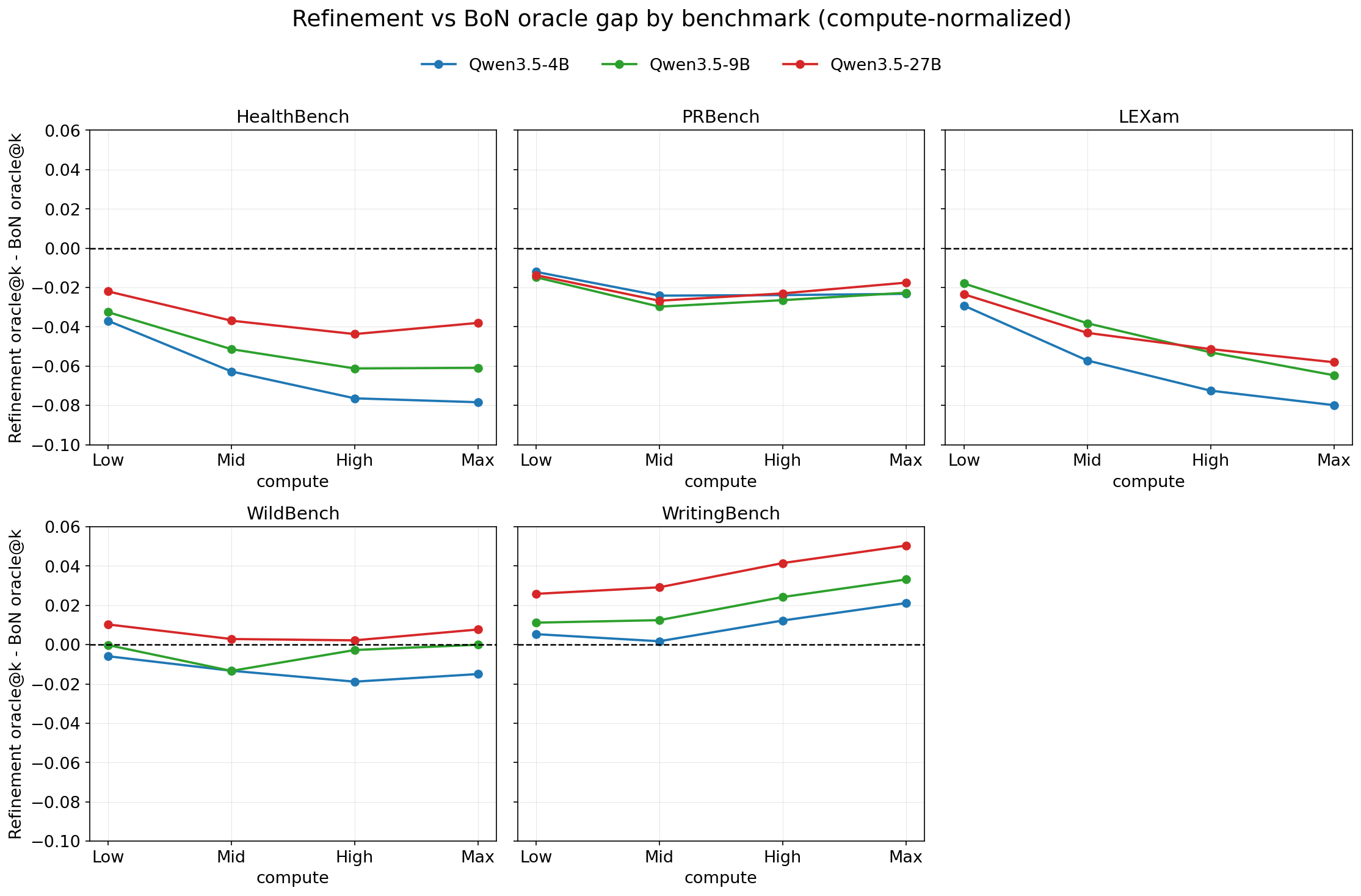}
    \caption{Sequential Refinement oracle vs BoN oracle across compute
    levels, by model family and benchmark.}
    \label{fig:self-refine-vs-bon-oracle-compute-norm}
\end{figure}

  To isolate the \emph{exploration} quality of sequential refinement from its exploitation cost, we compare its oracle score (the maximum score within the pool) against the BoN oracle. When comparing the oracles with matched compute in Figure \ref{fig:self-refine-vs-bon-oracle-compute-norm}, BoN oracle outperforms on 3/5 benchmarks, has roughly equivalent performance on WildBench, and is outperformed on WritingBench although this is known to have a length bias that benefits Refinement (Appendix \ref{app:verbosity_unified}).  
  
\begin{figure}[H]
    \centering
    \includegraphics[width=\textwidth]{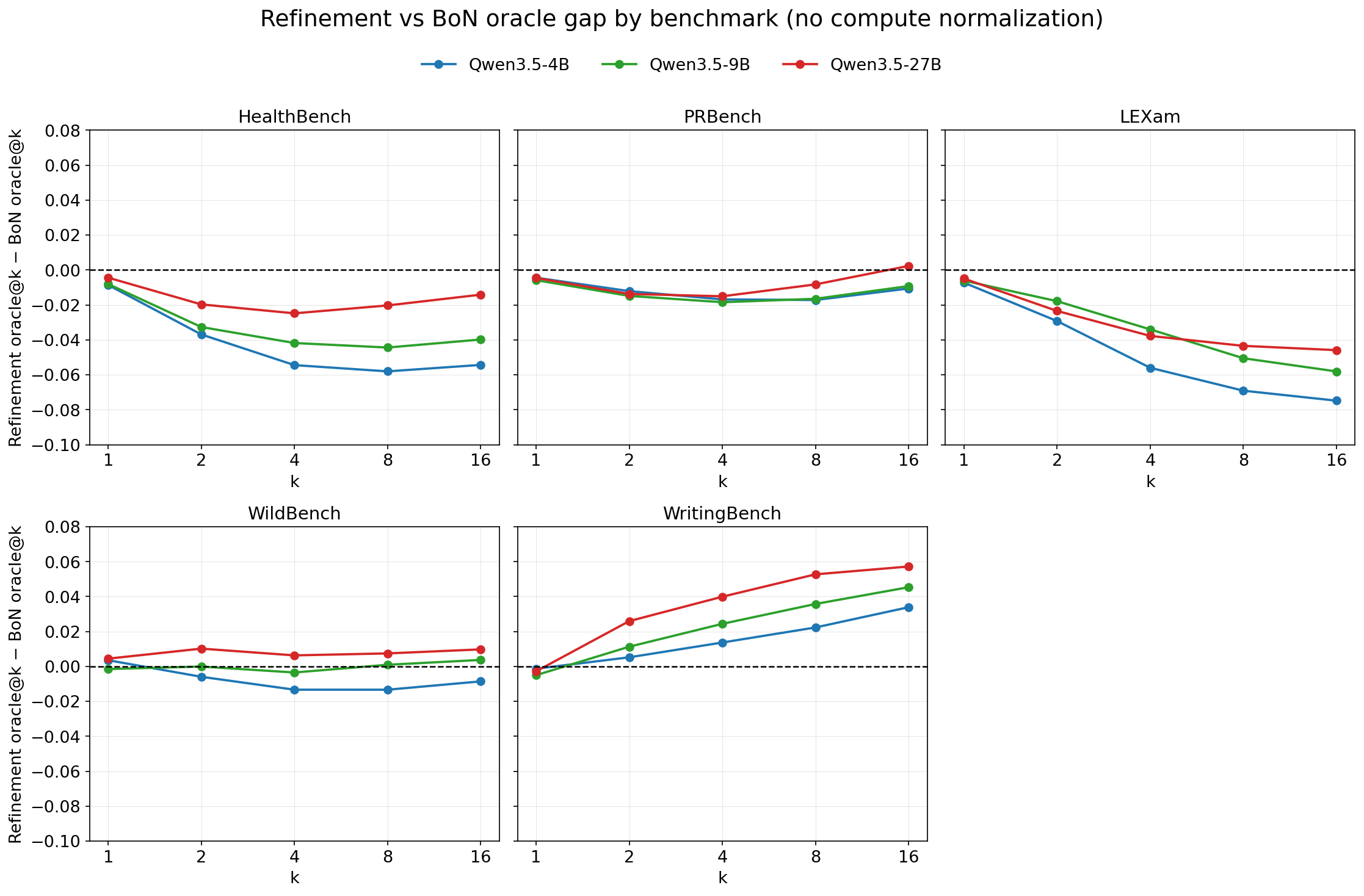}
    \caption{Sequential Refinement oracle vs BoN oracle across compute
    levels, by model family and benchmark. Note that x-axis refers to number of candidates in the pool so compare k self-revisions to k i.i.d. samples here.}
    \label{fig:self-refine-vs-bon-oracle}
\end{figure}

  Next, we ask whether the oracle underperformance is due to (1) fewer candidates from the additional exploitation cost, or (2) lower variance in the candidate pool relative to i.i.d.\ sampling. To test this, we match $k$ self-revisions against $k$ i.i.d.\ samples without compute normalization, asking whether iterative refinement produces a higher-quality pool (in terms of max score) even if exploitation were free.
   Figure~\ref{fig:self-refine-vs-bon-oracle} shows that any oracle advantage appears tied to model size, but in many cases sequential refinement drafts produce a \emph{worse} 
  oracle pool than i.i.d.\ sampling at matched $k$: two benchmarks show strictly negative effects, two show small effects, and WritingBench is the only benchmark with a large positive effect. As before, we attribute
  the WritingBench result to a length bias in the judge combined with sequential refinement's tendency to increase verbosity; see Appendix~\ref{app:verbosity_unified} for the supporting analysis.

\begin{figure}[H]
    \centering
    \includegraphics[width=\textwidth]{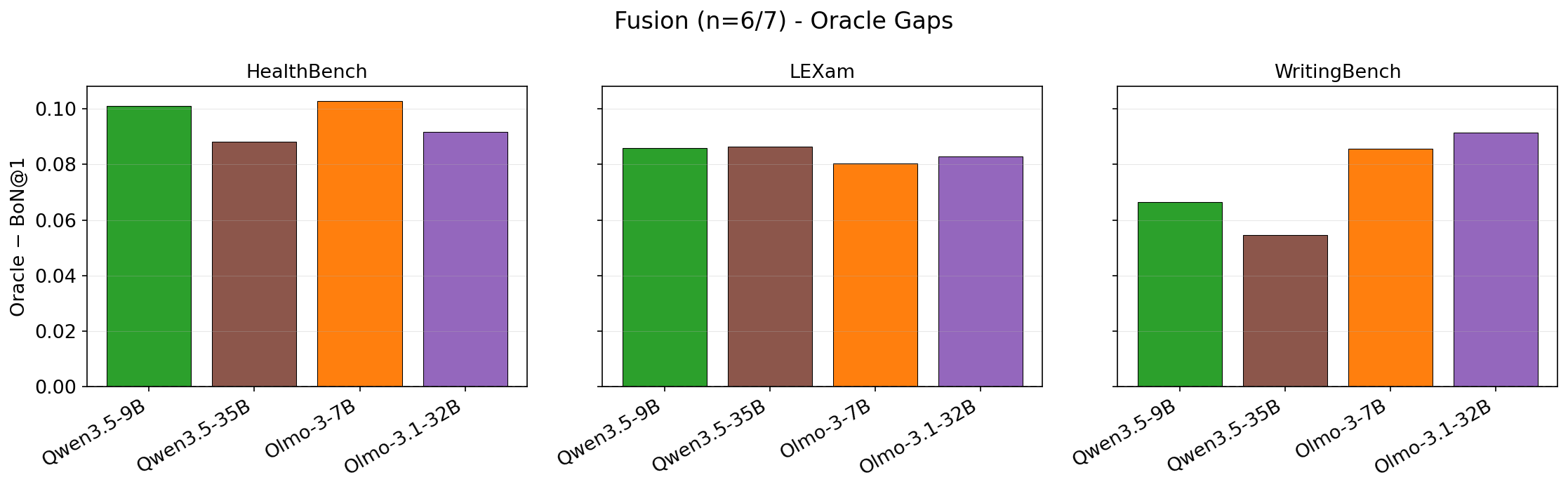}
    \caption{Oracle gaps over a BoN@1 baseline across models for fusion at medium compute (k=8). Note that Olmo models used 6 candidates instead of 7 due to context length limits. Additionally, context-length issues persisted for Olmo on PrBench and WildBench so they are not included here.}
    \label{fig:fusion-oracle-gaps}
\end{figure}

\begin{figure}[H]
    \centering
    \includegraphics[width=\textwidth]{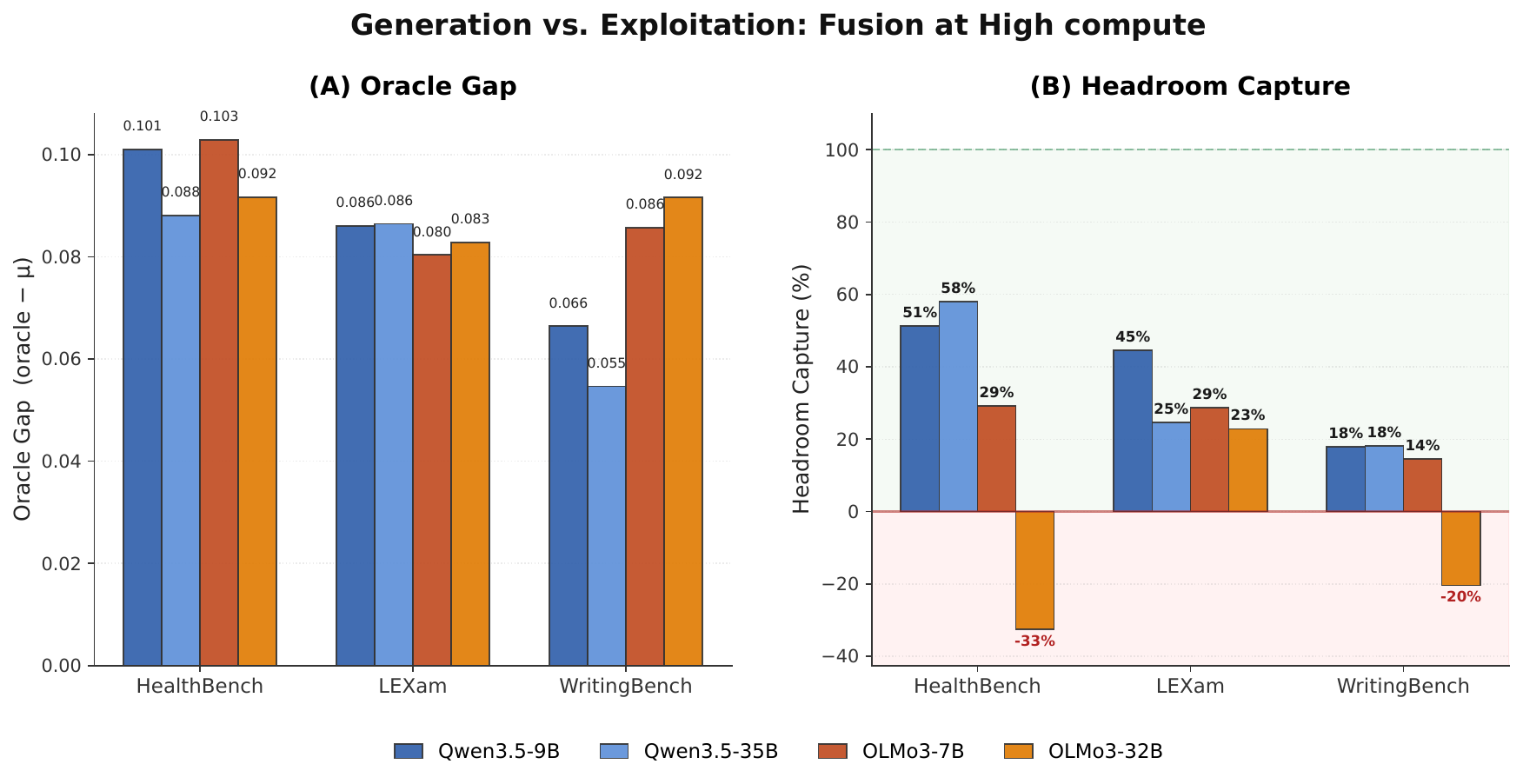}
    \caption{Oracle gap and headroom capture comparison between different families}
    \label{fig:model-families-headroom}
\end{figure}

\section{Task Demand Profiles}
\label{sec:general_scales}

To characterise what each benchmark truly measures, we applied the DeLeAn 
rubric from the ADeLe framework~\cite{zhou2026generalscalesunlockai} to 
annotate instances across our five benchmarks along 18 demand dimensions 
spanning primordial capabilities, domain knowledge, and extraneous factors. 
The resulting demand profiles are visualised as radar plots, where the inner 
ring reflects low-scoring instances (score 0) and the outer ring reflects 
high-scoring instances (score 5), with colour intensity encoding instance 
frequency at each demand level.

This characterisation serves a dual purpose. Beyond describing what each 
benchmark measures, it allows us to shed light on \textit{which primordial 
capabilities} each TTS method is principled to engage, grounding our 
empirical findings in the cognitive skill demands of each benchmark. We 
note that this mapping is interpretive rather than predictive: the demand 
profile identifies the skills a benchmark stresses, but whether a given TTS 
method benefits those skills depends on the exploitation bottleneck 
identified in Section~\ref{sec:analysis}. Crucially, the profiles do not 
yield a clean mapping from demand dimensions to method success---the same 
dimension can be associated with regression on one benchmark and gains on 
another, depending on how the benchmark operationalises that demand and how 
the judge evaluates it.

\begin{figure}[H]
    \centering
    \includegraphics[width=0.45\textwidth]{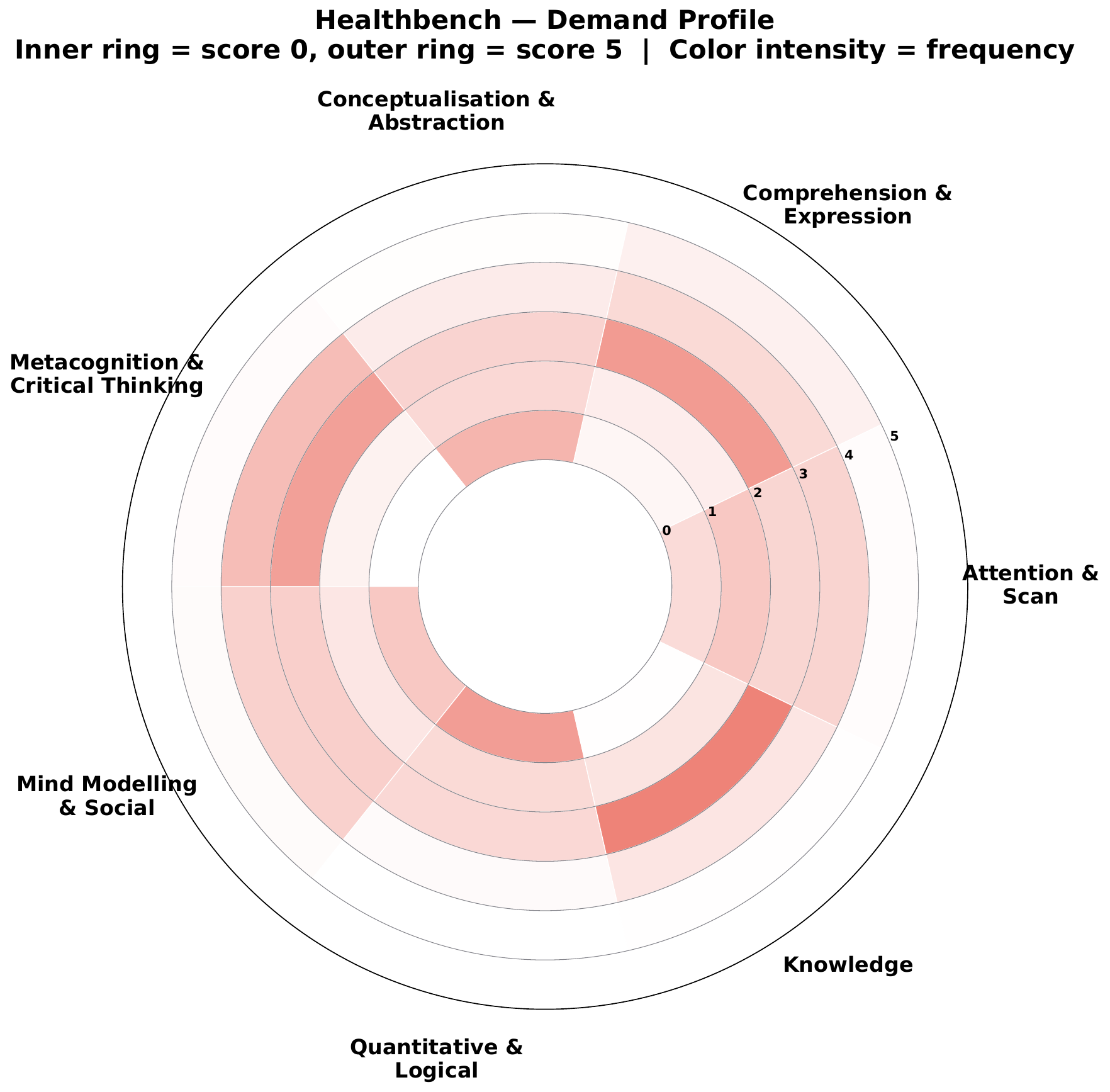}
    \hfill
    \includegraphics[width=0.45\textwidth]{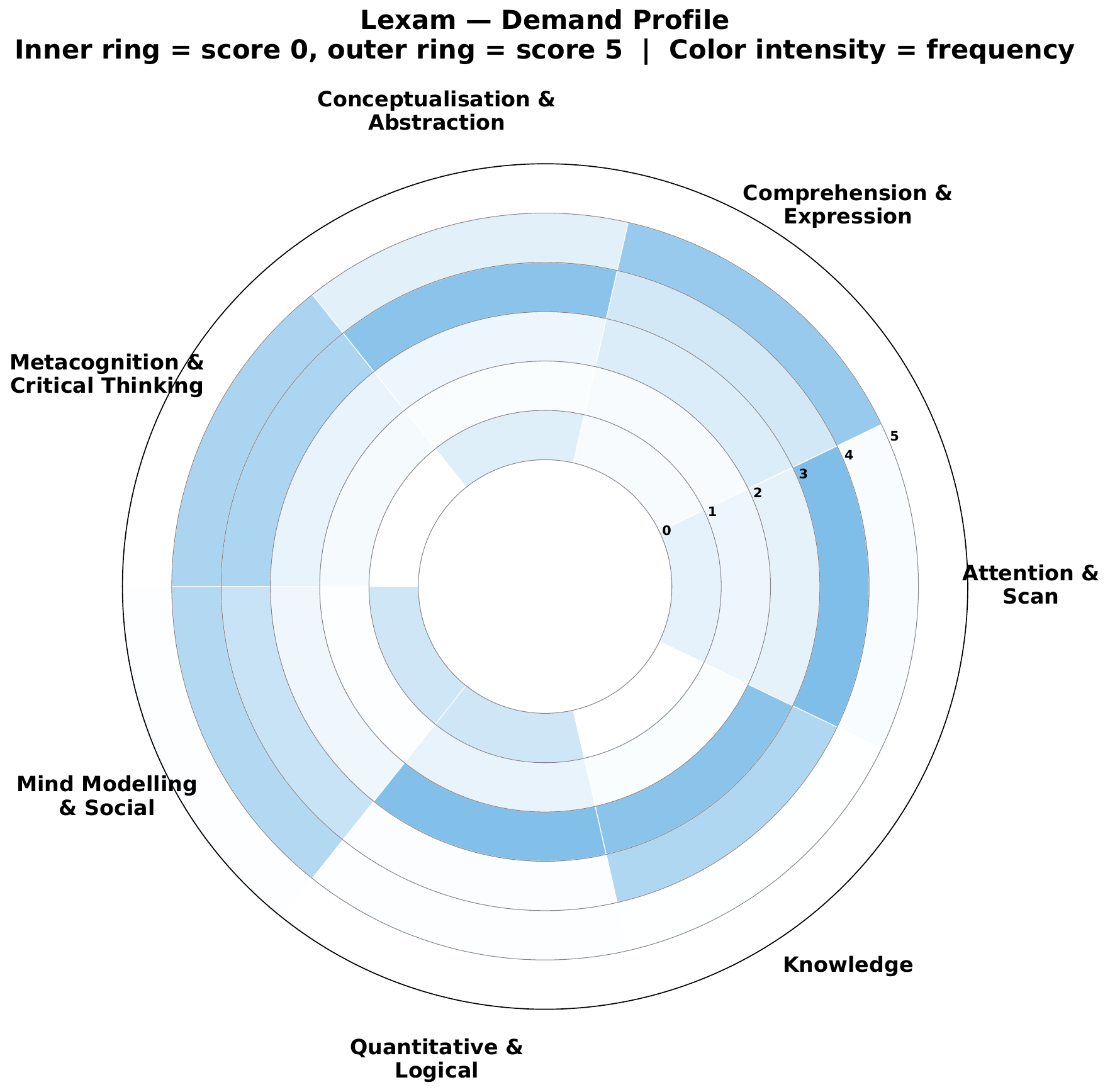}
    \caption{Demand profiles for HealthBench (left) and LEXam (right)}
    \label{fig:profiles_1}
\end{figure}

HealthBench is dominated by \textit{Metacognition \& Critical Thinking} 
(MC), with a strong secondary load on \textit{Comprehension \& Expression} 
(CE) (Figure~\ref{fig:profiles_1}, left). \textit{Mind Modelling \& Social} 
(MS) contributes moderately, while \textit{Conceptualisation \& Abstraction} 
(CL), \textit{Knowledge} (KN), \textit{Attention \& Scan} (AS), and 
\textit{Quantitative \& Logical} (QL) are comparatively absent. Notably, 
the low KN load is counterintuitive for a medical benchmark, but reflects 
that HealthBench instances stress the \textit{process} of reasoning over 
medical queries---self-monitoring and structured response generation 
against physician rubrics---rather than raw factual recall. The dominance 
of MC is consistent with Sequential Refinement's strong regression here ($-2.3$pp 
at XHigh): the first draft already engages the primary metacognitive demand, 
and iterative critique-and-rewrite saturates rather than improves it.

LEXam is dominated by \textit{Comprehension \& Expression} (CE), with 
\textit{Attention \& Scan} (AS) as the second most prominent 
dimension---markedly higher than in HealthBench 
(Figure~\ref{fig:profiles_1}, right). \textit{Metacognition \& Critical 
Thinking} (MC) and \textit{Knowledge} (KN) contribute moderately, while 
\textit{Conceptualisation \& Abstraction} (CL), \textit{Mind Modelling \& 
Social} (MS), and \textit{Quantitative \& Logical} (QL) are largely absent. 
The prominence of AS reflects the need to locate and integrate specific 
information across legal texts, while CE dominance is consistent with the 
interpretive demands of legal reasoning. LEXam exhibits the strongest 
Sequential Refinement regression of all benchmarks ($-4.5$pp at XHigh), despite 
having a different demand profile from HealthBench. This suggests the 
failure mode is not uniquely tied to MC dominance: tasks requiring precise 
comprehension and targeted information extraction (CE+AS) are equally 
ill-suited to iterative refinement, since conditioning on prior drafts 
compounds misinterpretations rather than correcting them.

\begin{figure}[H]
    \centering
    \includegraphics[width=0.45\textwidth]{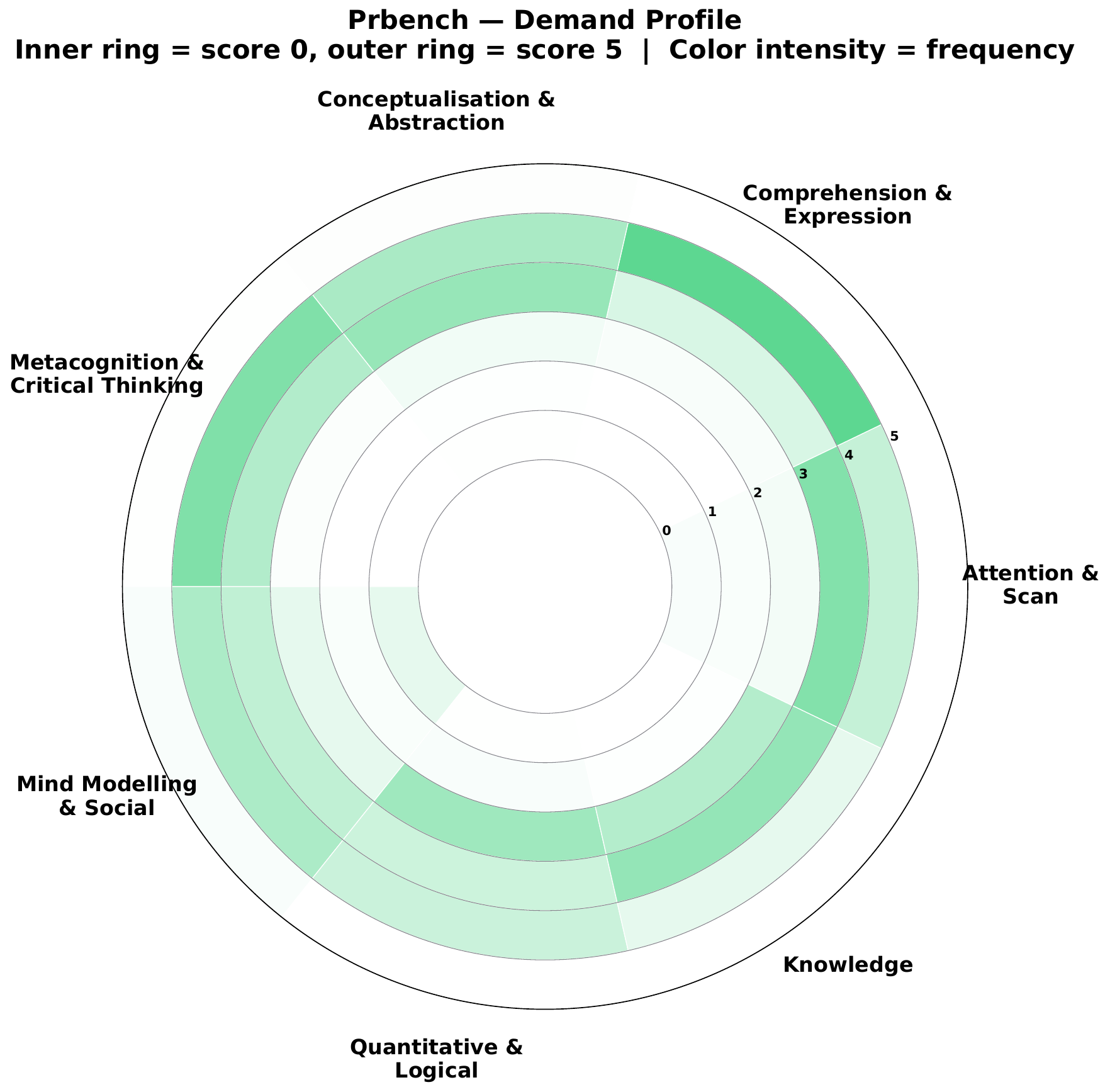}
    \hfill
    \includegraphics[width=0.45\textwidth]{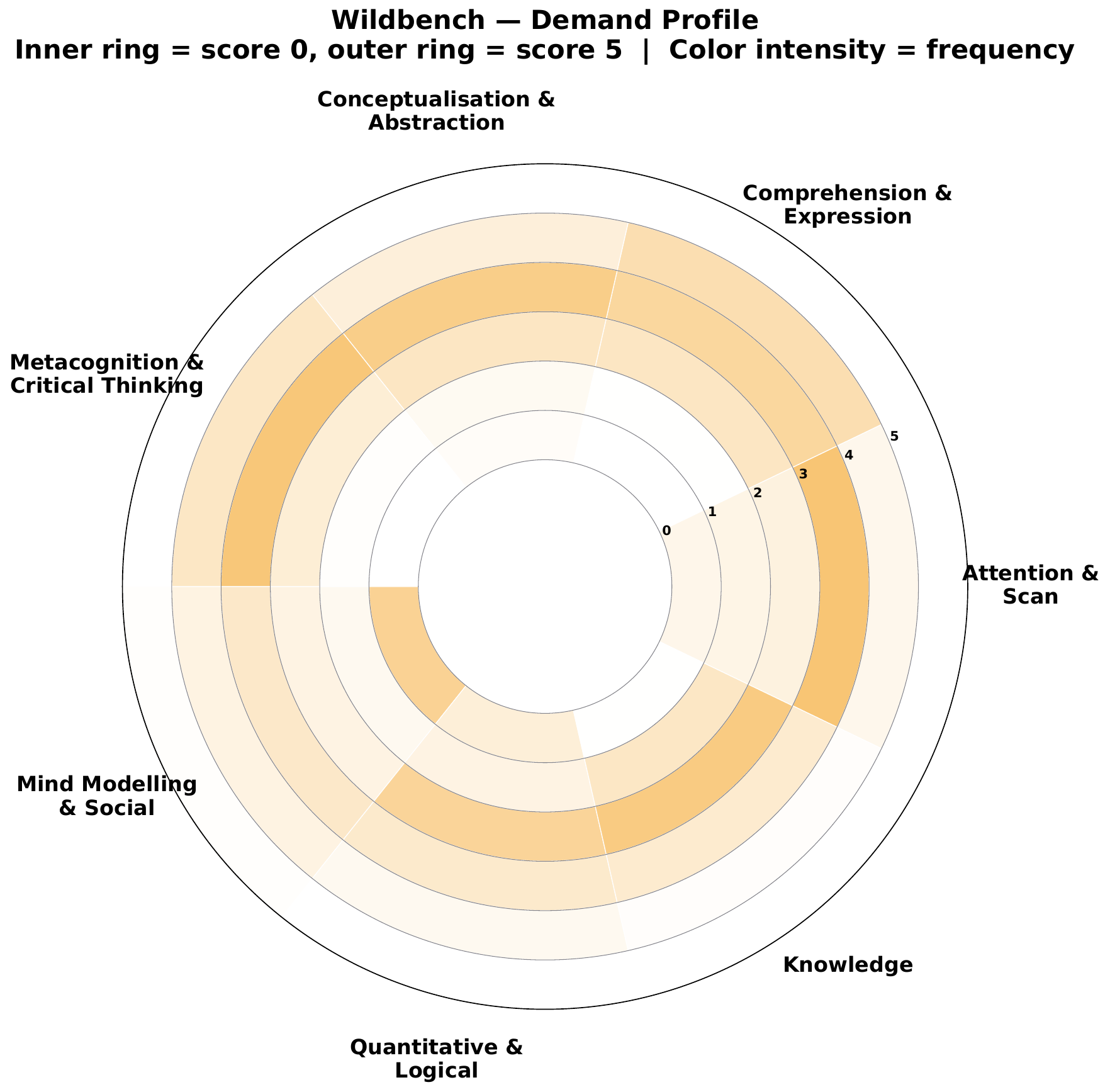}
    \caption{Demand profiles for PRBench (left) and WildBench (right)}
    \label{fig:profiles_2}
\end{figure}

PRBench is co-dominated by \textit{Comprehension \& Expression} (CE) and 
\textit{Metacognition \& Critical Thinking} (MC), with a moderate 
contribution from \textit{Conceptualisation \& Abstraction} (CL) and 
\textit{Mind Modelling \& Social} (MS) (Figure~\ref{fig:profiles_2}, left). 
\textit{Attention \& Scan} (AS), \textit{Knowledge} (KN), and 
\textit{Quantitative \& Logical} (QL) are comparatively absent. 
Sequential Refinement achieves its largest non-WritingBench gains on PRBench 
($+3.5$pp at XHigh), which is noteworthy given that PRBench shares its 
two dominant dimensions (CE, MC) with HealthBench, where Sequential Refinement 
strongly regresses. This apparent contradiction suggests that the demand 
profile alone does not determine method success: the key difference is 
likely how these dimensions are operationalised and evaluated. PRBench's 
rubric-based scoring rewards elaboration and structured reasoning across 
multiple criteria, whereas HealthBench's physician rubrics penalise 
deviation from precise clinical content. The CE+MC demand can therefore 
support either regression or gain depending on whether the judge rewards 
precision or elaboration.

WildBench is dominated by \textit{Metacognition \& Critical Thinking} 
(MC), with \textit{Comprehension \& Expression} (CE) as a strong second 
dimension (Figure~\ref{fig:profiles_2}, right). \textit{Attention \& Scan} 
(AS) is notably the third most prominent dimension, more so than in 
PRBench. \textit{Conceptualisation \& Abstraction} (CL) contributes 
moderately, while \textit{Mind Modelling \& Social} (MS), \textit{Knowledge} 
(KN), and \textit{Quantitative \& Logical} (QL) are comparatively absent. 
WildBench's MC dominance makes it superficially similar to HealthBench, 
yet Sequential Refinement gains modestly here ($+2.4$pp at XHigh) rather than 
regressing. This again points to evaluation operationalisation as the 
mediating factor: WildBench uses an LLM-generated holistic checklist 
judge that is more tolerant of elaborated responses, whereas HealthBench's 
physician rubrics are precision-anchored. The MC demand is present in both, 
but its interaction with the evaluation mechanism produces opposite outcomes.

\begin{figure}[H]
    \centering
    \includegraphics[width=0.45\textwidth]{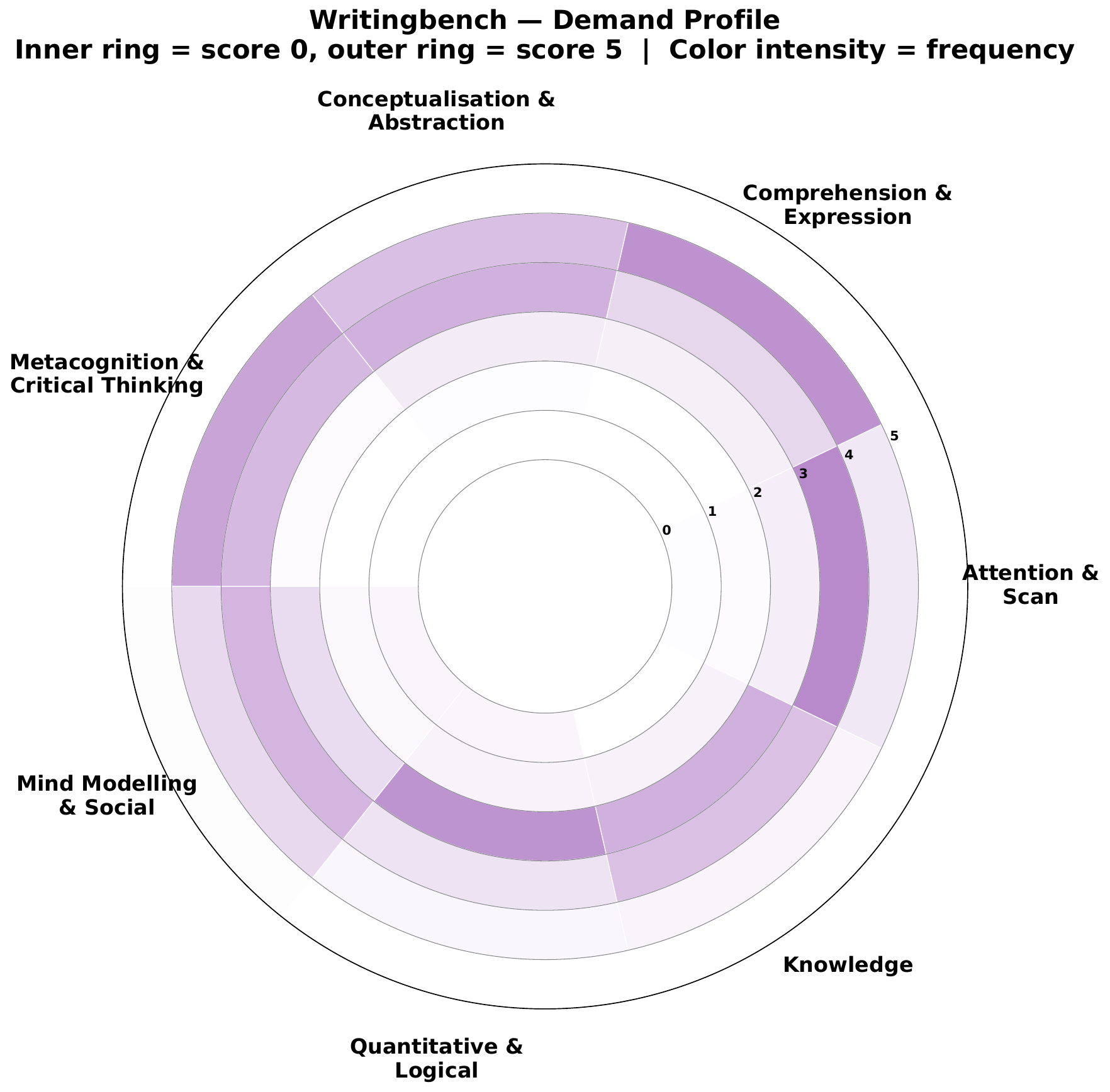}
    \caption{WritingBench demand profile}
    \label{fig:writingbench}
\end{figure}

WritingBench is distinctively loaded on \textit{Comprehension \& Expression} 
(CE) and \textit{Conceptualisation \& Abstraction} (CL), which together 
dominate the profile with the highest frequency at outer demand levels 
(Figure~\ref{fig:writingbench}). \textit{Metacognition \& Critical Thinking} 
(MC) contributes moderately, while \textit{Mind Modelling \& Social} (MS) 
and \textit{Attention \& Scan} (AS) are present but secondary. 
\textit{Knowledge} (KN) and \textit{Quantitative \& Logical} (QL) are 
comparatively absent. The CE+CL dominance, with near-absent knowledge and 
precision-retrieval demands, makes WritingBench the most distinctive profile 
among the five benchmarks. Tasks primarily requiring expression and 
conceptualisation have no fixed correct answer, which could plausibly reward 
iterative elaboration. However, as we show in 
Section~\ref{sec:generative_analysis} and Appendix~\ref{app:verbosity}, 
Sequential Refinement's large gains here ($+7.3$pp) are substantially confounded by 
the WritingBench judge's verbosity bias. The demand profile alone is 
insufficient to explain the magnitude: CE+CL makes WritingBench a 
plausible candidate for refinement benefit, but the verbosity confound 
means we cannot attribute the observed gains to genuine skill improvement 
without verbosity-controlled evaluation.

\paragraph{Cross-benchmark demand structure and method fit.}
The five profiles reveal that demand dimensions alone do not predict method 
success. The same dimensions---MC and CE---appear as dominant in 
HealthBench (Sequential Refinement regresses $-2.3$pp), PRBench (Sequential Refinement gains 
$+3.5$pp), and WildBench (Sequential Refinement gains $+2.4$pp). The differentiating 
factor is not the demand profile but how each benchmark operationalises and 
evaluates those demands.
Precision-sensitive judges — whether rubric-based (HealthBench) or reference-guided (LEXam) — penalise the elaboration that iterative refinement produces: HealthBench's physician rubrics reward only predefined clinical criteria, while LEXam's reference-guided LLM judge penalises the compounded misinterpretations that conditioning on prior drafts introduces. Checklist and holistic judges (PRBench, WildBench, WritingBench) are, by contrast, more tolerant of or actively reward elaborated responses. This implies that the demand profile is useful for characterising \textit{what} a benchmark tests, but predicting TTS method success requires additionally knowing \textit{how}the benchmark evaluates outputs. Fusion, by contrast, improves uniformly across all five profiles, confirming that its gains are structural — bypassing verifier quality entirely — and not contingent on any particular demand configuration or evaluation style.

These profiles make explicit what the per-method analyses implied: demand
dimensions identify \textit{what} a benchmark stresses, but they do not
determine method success on their own. The same MC+CE profile supports
opposite Sequential Refinement outcomes depending on whether the judge rewards
precision or elaboration. This evaluation-operationalisation interaction is
orthogonal to model capability.



\section{Token usage}
\label{app:tokens}


\begin{table}[H]
  \centering
  \small
  \caption{BoN $n{=}16$ total completion tokens and $\pm27\%$ compute band (min--max) per benchmark.}
  \label{tab:bon-tokens}
  \resizebox{\textwidth}{!}{
  \begin{tabular}{lrrrrrrrr}
  \toprule
  Benchmark & \multicolumn{2}{c}{Qwen3.5-9B} & \multicolumn{2}{c}{Qwen3.5-35B-A3B} & \multicolumn{2}{c}{OLMo-3-7B} & \multicolumn{2}{c}{OLMo-3.1-32B} \\
  \cmidrule(lr){2-3} \cmidrule(lr){4-5} \cmidrule(lr){6-7} \cmidrule(lr){8-9}
  & Tokens & Band [min--max] & Tokens & Band [min--max] & Tokens & Band [min--max] & Tokens & Band [min--max] \\
  \midrule
  healthbench\_communication & 39.6M & [28.9M--50.3M] & 35.7M & [26.1M--45.4M] & 32.9M & [24.0M--41.8M] & 34.0M & [24.8M--43.2M] \\
  healthbench\_complex\_responses & 17.3M & [12.6M--22.0M] & 16.2M & [11.8M--20.6M] & 14.1M & [10.3M--17.9M] & 16.0M & [11.7M--20.3M] \\
  healthbench\_context\_seeking & 19.3M & [14.1M--24.5M] & 17.8M & [13.0M--22.6M] & 15.6M & [11.4M--19.8M] & 15.6M & [11.4M--19.8M] \\
  healthbench\_emergency\_referrals & 15.9M & [11.6M--20.1M] & 14.5M & [10.6M--18.4M] & 12.6M & [9.2M--16.0M] & 12.1M & [8.8M--15.3M] \\
  healthbench\_global\_health & 41.3M & [30.1M--52.4M] & 37.5M & [27.4M--47.7M] & 34.5M & [25.2M--43.9M] & 34.7M & [25.3M--44.0M] \\
  healthbench\_health\_data\_tasks & 20.8M & [15.2M--26.4M] & 19.7M & [14.4M--25.1M] & 17.3M & [12.7M--22.0M] & 20.1M & [14.7M--25.5M] \\
  healthbench\_hedging & 39.2M & [28.6M--49.8M] & 35.7M & [26.1M--45.3M] & 29.9M & [21.8M--38.0M] & 31.5M & [23.0M--40.0M] \\
  lexam\_open & 31.1M & [22.7M--39.5M] & 27.8M & [20.3M--35.4M] & 24.0M & [17.5M--30.4M] & 25.8M & [18.8M--32.8M] \\
  prbench\_finance & 48.2M & [35.2M--61.2M] & 44.6M & [32.5M--56.6M] & 45.6M & [33.3M--58.0M] & 44.3M & [32.3M--56.2M] \\
  prbench\_finance\_hard & 24.5M & [17.9M--31.1M] & 22.8M & [16.6M--28.9M] & 22.9M & [16.7M--29.1M] & 22.2M & [16.2M--28.2M] \\
  prbench\_legal & 32.6M & [23.8M--41.4M] & 30.7M & [22.4M--38.9M] & 24.7M & [18.0M--31.3M] & 26.3M & [19.2M--33.4M] \\
  prbench\_legal\_hard & 17.1M & [12.5M--21.7M] & 16.0M & [11.7M--20.4M] & 12.5M & [9.2M--15.9M] & 13.4M & [9.8M--17.1M] \\
  WildBench & 88.7M & [64.8M--112.7M] & 80.0M & [58.4M--101.6M] & 82.0M & [59.9M--104.2M] & 85.7M & [62.6M--108.9M] \\
  writingbench\_academic\_engineering & 7.9M & [5.8M--10.0M] & 8.1M & [5.9M--10.2M] & 8.1M & [5.9M--10.4M] & 9.9M & [7.2M--12.6M] \\
  writingbench\_advertising\_marketing & 5.0M & [3.7M--6.4M] & 5.0M & [3.6M--6.3M] & 5.0M & [3.6M--6.3M] & 6.4M & [4.7M--8.2M] \\
  writingbench\_education & 4.4M & [3.2M--5.6M] & 4.3M & [3.2M--5.5M] & 4.8M & [3.5M--6.0M] & 5.5M & [4.0M--7.0M] \\
  writingbench\_finance\_business & 9.3M & [6.8M--11.8M] & 9.3M & [6.8M--11.8M] & 9.2M & [6.7M--11.7M] & 10.5M & [7.7M--13.4M] \\
  writingbench\_literature\_arts & 9.2M & [6.7M--11.6M] & 9.0M & [6.6M--11.5M] & 7.1M & [5.2M--9.0M] & 8.6M & [6.3M--10.9M] \\
  writingbench\_politics\_law & 7.3M & [5.3M--9.2M] & 7.0M & [5.1M--8.9M] & 6.9M & [5.1M--8.8M] & 8.7M & [6.4M--11.1M] \\
  \bottomrule
  \end{tabular}
}
\end{table}

\begin{table}[H]
  \centering
  \small
  \caption{Particle Filter (Mid ($n{=}4$)) total completion tokens per benchmark.}
  \label{tab:pf-low}
  \resizebox{\textwidth}{!}{
  \begin{tabular}{lrrrr}
  \toprule
  Benchmark & Qwen3.5-9B & Qwen3.5-35B-A3B & OLMo-3-7B & OLMo-3.1-32B \\
  \midrule
  healthbench\_communication & 10.2M & 9.0M & 8.3M & 8.6M \\
  healthbench\_complex\_responses & 4.5M & 4.3M & 3.4M & 3.9M \\
  healthbench\_context\_seeking & 5.0M & 4.6M & 3.9M & 3.9M \\
  healthbench\_emergency\_referrals & 4.1M & 3.7M & 3.1M & 3.0M \\
  healthbench\_global\_health & 10.5M & 9.4M & 8.7M & 8.9M \\
  healthbench\_health\_data\_tasks & 5.5M & 5.0M & 4.3M & 5.1M \\
  healthbench\_hedging & 10.3M & 9.2M & 7.5M & 7.9M \\
  lexam\_open & 8.0M & 7.0M & 6.0M & 6.4M \\
  prbench\_finance & 12.3M & 11.3M & 12.3M & 11.6M \\
  prbench\_finance\_hard & 6.3M & 5.8M & 6.1M & 5.9M \\
  prbench\_legal & 8.2M & 7.7M & 6.1M & 6.6M \\
  prbench\_legal\_hard & 4.4M & 4.0M & 3.0M & 3.3M \\
  WildBench & 22.6M & 20.6M & 22.5M & 23.7M \\
  writingbench\_academic\_engineering & 2.0M & 2.0M & 2.1M & 2.5M \\
  writingbench\_advertising\_marketing & 1.4M & 1.3M & 1.2M & 1.7M \\
  writingbench\_education & 1.1M & 1.1M & 1.2M & 1.4M \\
  writingbench\_finance\_business & 2.4M & 2.3M & 2.3M & 2.7M \\
  writingbench\_literature\_arts & 2.3M & 2.2M & 1.8M & 2.3M \\
  writingbench\_politics\_law & 1.8M & 1.7M & 1.8M & 2.3M \\
  \bottomrule
  \end{tabular}
}
\end{table}

\begin{table}[H]
  \centering
  \small
  \caption{Particle Filter (High ($n{=}8$)) total completion tokens per benchmark.}
  \label{tab:pf-mid}
  \resizebox{\textwidth}{!}{
  \begin{tabular}{lrrrr}
  \toprule
  Benchmark & Qwen3.5-9B & Qwen3.5-35B-A3B & OLMo-3-7B & OLMo-3.1-32B \\
  \midrule
  healthbench\_communication & 20.3M & 18.2M & 16.5M & 17.1M \\
  healthbench\_complex\_responses & 9.1M & 8.6M & 6.8M & 8.0M \\
  healthbench\_context\_seeking & 10.0M & 9.1M & 7.8M & 8.0M \\
  healthbench\_emergency\_referrals & 8.2M & 7.3M & 6.4M & 6.1M \\
  healthbench\_global\_health & 21.2M & 18.9M & 17.5M & 17.7M \\
  healthbench\_health\_data\_tasks & 10.9M & 10.2M & 8.3M & 9.9M \\
  healthbench\_hedging & 20.2M & 18.3M & 15.0M & 16.1M \\
  lexam\_open & 15.9M & 14.0M & 11.8M & 12.8M \\
  prbench\_finance & 24.3M & 22.4M & 24.7M & 23.7M \\
  prbench\_finance\_hard & 12.5M & 11.4M & 12.6M & 11.9M \\
  prbench\_legal & 16.4M & 15.3M & 12.3M & 13.2M \\
  prbench\_legal\_hard & 8.7M & 8.0M & 6.1M & 6.9M \\
  WildBench & 45.4M & 40.8M & 44.7M & 47.7M \\
  writingbench\_academic\_engineering & 3.9M & 4.1M & 3.9M & 5.1M \\
  writingbench\_advertising\_marketing & 2.7M & 2.6M & 2.5M & 3.4M \\
  writingbench\_education & 2.3M & 2.2M & 2.3M & 2.9M \\
  writingbench\_finance\_business & 4.6M & 4.6M & 4.6M & 5.2M \\
  writingbench\_literature\_arts & 4.4M & 4.6M & 3.7M & 4.6M \\
  writingbench\_politics\_law & 3.6M & 3.6M & 3.5M & 4.3M \\
  \bottomrule
  \end{tabular}
}
\end{table}

\begin{table}[H]
  \centering
  \small
  \caption{Particle Filter (XHigh ($n{=}16$)) total completion tokens per benchmark.}
  \label{tab:pf-high}
  \resizebox{\textwidth}{!}{
  \begin{tabular}{lrrrr}
  \toprule
  Benchmark & Qwen3.5-9B & Qwen3.5-35B-A3B & OLMo-3-7B & OLMo-3.1-32B \\
  \midrule
  healthbench\_communication & 40.5M & 36.1M & 33.0M & 35.0M \\
  healthbench\_complex\_responses & 18.1M & 17.0M & 13.7M & 15.9M \\
  healthbench\_context\_seeking & 20.0M & 18.2M & 16.0M & 16.2M \\
  healthbench\_emergency\_referrals & 16.5M & 14.7M & 12.9M & 12.5M \\
  healthbench\_global\_health & 42.3M & 37.8M & 35.2M & 36.3M \\
  healthbench\_health\_data\_tasks & 21.9M & 20.5M & 16.8M & 20.4M \\
  healthbench\_hedging & 40.7M & 36.5M & 30.3M & 32.8M \\
  lexam\_open & 31.5M & 27.9M & 23.6M & 25.5M \\
  prbench\_finance & 48.1M & 44.5M & 49.4M & 47.5M \\
  prbench\_finance\_hard & 24.7M & 22.6M & 25.0M & 23.5M \\
  prbench\_legal & 33.0M & 30.7M & 24.5M & 26.3M \\
  prbench\_legal\_hard & 17.0M & 16.0M & 12.2M & 13.5M \\
  WildBench & 89.6M & 81.5M & 89.8M & 95.2M \\
  writingbench\_academic\_engineering & 7.9M & 8.2M & 8.6M & 10.4M \\
  writingbench\_advertising\_marketing & 5.1M & 5.3M & 5.1M & 6.9M \\
  writingbench\_education & 4.4M & 4.4M & 4.8M & 5.6M \\
  writingbench\_finance\_business & 9.3M & 9.4M & 9.1M & 10.8M \\
  writingbench\_literature\_arts & 8.9M & 9.0M & 6.8M & 9.0M \\
  writingbench\_politics\_law & 7.5M & 7.2M & 6.9M & 8.9M \\
  \bottomrule
  \end{tabular}
}
\end{table}

\begin{table}[H]
  \centering
  \small
  \caption{Beam Search (Mid ($\mathrm{{bw}}{=}2,\,n{=}2$)) total completion tokens per benchmark.}
  \label{tab:bs-low}
  \resizebox{\textwidth}{!}{
  \begin{tabular}{lrrrr}
  \toprule
  Benchmark & Qwen3.5-9B & Qwen3.5-35B-A3B & OLMo-3-7B & OLMo-3.1-32B \\
  \midrule
  healthbench\_communication & 10.2M & 9.0M & 8.0M & 8.4M \\
  healthbench\_complex\_responses & 4.7M & 4.2M & 3.4M & 4.1M \\
  healthbench\_context\_seeking & 5.0M & 4.6M & 3.8M & 3.8M \\
  healthbench\_emergency\_referrals & 4.1M & 3.7M & 2.9M & 2.9M \\
  healthbench\_global\_health & 10.5M & 9.3M & 8.4M & 8.5M \\
  healthbench\_health\_data\_tasks & 5.5M & 5.1M & 3.9M & 5.0M \\
  healthbench\_hedging & 10.3M & 9.2M & 7.3M & 7.7M \\
  lexam\_open & 7.9M & 7.0M & 5.9M & 6.4M \\
  prbench\_finance & 12.3M & 11.2M & 12.4M & 11.9M \\
  prbench\_finance\_hard & 6.3M & 5.7M & 6.2M & 5.9M \\
  prbench\_legal & 8.1M & 7.6M & 6.0M & 6.4M \\
  prbench\_legal\_hard & 4.3M & 4.0M & 3.1M & 3.3M \\
  WildBench & 24.1M & 20.5M & 22.5M & 23.9M \\
  writingbench\_academic\_engineering & 2.1M & 2.1M & 2.1M & 2.5M \\
  writingbench\_advertising\_marketing & 1.3M & 1.3M & 1.2M & 1.7M \\
  writingbench\_education & 1.2M & 1.2M & 1.2M & 1.4M \\
  writingbench\_finance\_business & 2.4M & 2.4M & 2.3M & 2.7M \\
  writingbench\_literature\_arts & 2.4M & 2.3M & 1.9M & 2.2M \\
  writingbench\_politics\_law & 1.9M & 1.7M & 1.7M & 2.2M \\
  \bottomrule
  \end{tabular}
}
\end{table}

\begin{table}[H]
  \centering
  \small
  \caption{Beam Search (High ($\mathrm{{bw}}{=}2,\,n{=}4$)) total completion tokens per benchmark.}
  \label{tab:bs-mid}
  \resizebox{\textwidth}{!}{
  \begin{tabular}{lrrrr}
  \toprule
  Benchmark & Qwen3.5-9B & Qwen3.5-35B-A3B & OLMo-3-7B & OLMo-3.1-32B \\
  \midrule
  healthbench\_communication & 19.8M & 17.7M & 15.7M & 16.4M \\
  healthbench\_complex\_responses & 9.4M & 8.8M & 6.6M & 7.9M \\
  healthbench\_context\_seeking & 9.7M & 8.8M & 7.4M & 7.5M \\
  healthbench\_emergency\_referrals & 8.1M & 7.2M & 5.7M & 5.5M \\
  healthbench\_global\_health & 20.6M & 18.4M & 16.4M & 16.6M \\
  healthbench\_health\_data\_tasks & 10.7M & 10.2M & 8.0M & 9.6M \\
  healthbench\_hedging & 20.4M & 18.2M & 14.1M & 15.1M \\
  lexam\_open & 16.0M & 13.7M & 11.8M & 12.8M \\
  prbench\_finance & 24.2M & 22.1M & 25.0M & 23.5M \\
  prbench\_finance\_hard & 12.5M & 11.2M & 12.3M & 12.0M \\
  prbench\_legal & 16.1M & 15.0M & 11.9M & 12.9M \\
  prbench\_legal\_hard & 8.5M & 7.8M & 6.0M & 6.4M \\
  WildBench & 46.6M & 41.6M & 45.0M & 47.2M \\
  writingbench\_academic\_engineering & 4.0M & 4.1M & 4.0M & 5.2M \\
  writingbench\_advertising\_marketing & 2.7M & 2.6M & 2.4M & 3.3M \\
  writingbench\_education & 2.3M & 2.2M & 2.3M & 2.8M \\
  writingbench\_finance\_business & 4.7M & 4.7M & 4.6M & 5.2M \\
  writingbench\_literature\_arts & 5.1M & 4.6M & 3.6M & 4.4M \\
  writingbench\_politics\_law & 3.8M & 3.5M & 3.3M & 4.5M \\
  \bottomrule
  \end{tabular}
}
\end{table}

\begin{table}[H]
  \centering
  \small
  \caption{Beam Search (XHigh ($\mathrm{{bw}}{=}4,\,n{=}4$)) total completion tokens per benchmark. $^{\dagger}$Run at $\mathrm{{bw}}{=}3,\,n{=}4$.}
  \label{tab:bs-high}
  \resizebox{\textwidth}{!}{
  \begin{tabular}{lrrrr}
  \toprule
  Benchmark & Qwen3.5-9B & Qwen3.5-35B-A3B & OLMo-3-7B & OLMo-3.1-32B \\
  \midrule
  healthbench\_communication & 40.4M & 34.9M & 32.0M & 34.0M \\
  healthbench\_complex\_responses & 21.6M & 19.8M & 14.0M & 16.7M \\
  healthbench\_context\_seeking & 21.0M & 18.1M & 15.1M & 15.5M \\
  healthbench\_emergency\_referrals & 16.6M & 14.4M & 11.3M & 11.8M \\
  healthbench\_global\_health & 42.6M & 36.6M & 33.3M & 34.9M \\
  healthbench\_health\_data\_tasks & 23.3M & 21.4M & 16.6M & 20.4M \\
  healthbench\_hedging & 45.3M & 38.4M & 29.6M & 32.2M \\
  lexam\_open & 32.1M & 27.9M & 23.6M & 26.8M \\
  prbench\_finance & 48.6M & 44.3M & 50.1M & 47.7M \\
  prbench\_finance\_hard & 24.5M & 22.5M & 25.7M & 23.8M \\
  prbench\_legal & 32.8M & 30.0M & 24.3M & 25.9M \\
  prbench\_legal\_hard & 16.8M & 15.7M & 12.1M & 13.2M \\
  WildBench & 100.0M & 89.6M & 92.8M & 99.2M \\
  writingbench\_academic\_engineering & 9.1M & 9.2M & 8.8M & 11.0M \\
  writingbench\_advertising\_marketing & 4.1M$^{\dagger}$ & 5.6M & 5.0M & 6.6M \\
  writingbench\_education & 5.1M & 5.1M & 4.6M & 6.2M \\
  writingbench\_finance\_business & 10.4M & 10.1M & 9.3M & 11.1M \\
writingbench\_literature\_arts & 10.8M & 7.3M$^{\dagger}$ & 7.0M & 9.0M \\
writingbench\_politics\_law & 8.1M & 8.1M & 7.0M & 9.3M \\
  \bottomrule
  \end{tabular}
}
\end{table}
\begin{table}[H]
  \centering
  \small
  \caption{Sequential Refinement ($\text{iter}{=}16$) total completion tokens per benchmark.}
  \label{tab:sr-iter16}
  \resizebox{\textwidth}{!}{
  \begin{tabular}{lrrrr}
  \toprule
  Benchmark & Qwen3.5-9B & Qwen3.5-35B-A3B & OLMo-3-7B & OLMo-3.1-32B \\
  \midrule
  healthbench\_communication & 83.0M & 82.6M & -- & -- \\
  healthbench\_complex\_responses & 33.0M & 32.4M & -- & -- \\
  healthbench\_context\_seeking & 45.6M & 44.7M & -- & -- \\
  healthbench\_emergency\_referrals & 36.4M & 36.0M & -- & -- \\
  healthbench\_global\_health & 91.6M & 88.3M & -- & -- \\
  healthbench\_health\_data\_tasks & 40.1M & 40.2M & -- & -- \\
  healthbench\_hedging & 85.2M & 83.1M & -- & -- \\
  lexam\_open & 48.7M & 48.1M & -- & -- \\
  prbench\_finance & 71.3M & 74.8M & -- & -- \\
  prbench\_finance\_hard & 35.8M & 38.0M & -- & -- \\
  prbench\_legal & 54.5M & 58.2M & -- & -- \\
  prbench\_legal\_hard & 27.5M & 29.5M & -- & -- \\
  WildBench & 120.6M & 121.5M & -- & -- \\
  writingbench\_academic\_engineering & 14.3M & 15.3M & -- & -- \\
  writingbench\_advertising\_marketing & 7.7M & 7.8M & -- & -- \\
  writingbench\_education & 7.5M & 7.8M & -- & -- \\
  writingbench\_finance\_business & 17.0M & 17.7M & -- & -- \\
  writingbench\_literature\_arts & 13.6M & 14.1M & -- & -- \\
  writingbench\_politics\_law & 11.7M & 12.0M & -- & -- \\
  \bottomrule
  \end{tabular}
}
\end{table}

\begin{table}[H]
  \centering
  \small
  \caption{Fusion (Mid ($s{=}3$)) total completion tokens per benchmark.}
  \label{tab:fusion-low}
  \resizebox{\textwidth}{!}{
  \begin{tabular}{lrrrr}
  \toprule
  Benchmark & Qwen3.5-9B & Qwen3.5-35B-A3B & OLMo-3-7B & OLMo-3.1-32B \\
  \midrule
  healthbench\_communication & 7.4M & 6.7M & 6.2M & 6.4M \\
  healthbench\_complex\_responses & 3.2M & 3.0M & 2.6M & 3.0M \\
  healthbench\_context\_seeking & 3.6M & 3.3M & 2.9M & 2.9M \\
  healthbench\_emergency\_referrals & 3.0M & 2.7M & 2.4M & 2.3M \\
  healthbench\_global\_health & 7.7M & 7.0M & 6.5M & 6.5M \\
  healthbench\_health\_data\_tasks & 3.9M & 3.7M & 3.2M & 3.8M \\
  healthbench\_hedging & 7.3M & 6.7M & 5.6M & 5.9M \\
  lexam\_open & 5.8M & 5.2M & 4.5M & 4.8M \\
  prbench\_finance & 9.0M & 8.4M & 8.6M & 8.3M \\
  prbench\_finance\_hard & 4.6M & 4.3M & 4.3M & 4.2M \\
  prbench\_legal & 6.1M & 5.7M & 4.6M & 4.9M \\
  prbench\_legal\_hard & 3.2M & 3.0M & 2.4M & 2.5M \\
  WildBench & 16.6M & 15.0M & 15.4M & 16.1M \\
  writingbench\_academic\_engineering & 1.5M & 1.5M & 1.5M & 1.9M \\
  writingbench\_advertising\_marketing & 940K & 934K & 930K & 1.2M \\
  writingbench\_education & 825K & 809K & 892K & 1.0M \\
  writingbench\_finance\_business & 1.7M & 1.7M & 1.7M & 2.0M \\
  writingbench\_literature\_arts & 1.7M & 1.7M & 1.3M & 1.6M \\
  writingbench\_politics\_law & 1.4M & 1.3M & 1.3M & 1.6M \\
  \bottomrule
  \end{tabular}
}
\end{table}

\begin{table}[H]
  \centering
  \small
  \caption{Fusion (High, $s{=}6 or 7$) total completion tokens per benchmark.}
  \label{tab:fusion-mid}
  \resizebox{\textwidth}{!}{
  \begin{tabular}{lrrrr}
  \toprule
  Benchmark & Qwen3.5-9B & Qwen3.5-35B-A3B & OLMo-3-7B & OLMo-3.1-32B \\
  \midrule
  healthbench\_communication & 14.9M & 13.4M & 12.3M & 12.8M \\
  healthbench\_complex\_responses & 6.5M & 6.1M & 5.3M & 6.0M \\
  healthbench\_context\_seeking & 7.2M & 6.7M & 5.8M & 5.8M \\
  healthbench\_emergency\_referrals & 5.9M & 5.4M & 4.7M & 4.5M \\
  healthbench\_global\_health & 15.5M & 14.1M & 13.0M & 13.0M \\
  healthbench\_health\_data\_tasks & 7.8M & 7.4M & 6.5M & 7.5M \\
  healthbench\_hedging & 14.7M & 13.4M & 11.2M & 11.8M \\
  lexam\_open & 11.7M & 10.4M & 9.0M & 9.7M \\
  prbench\_finance & 18.1M & 16.7M & 17.1M & 16.6M \\
  prbench\_finance\_hard & 9.2M & 8.5M & 8.6M & 8.3M \\
  prbench\_legal & 12.2M & 11.5M & 9.3M & 9.9M \\
  prbench\_legal\_hard & 6.4M & 6.0M & 4.7M & 5.0M \\
  WildBench & 33.3M & 30.0M & 30.8M & 32.1M \\
  writingbench\_academic\_engineering & 3.0M & 3.0M & 3.1M & 3.7M \\
  writingbench\_advertising\_marketing & 1.9M & 1.9M & 1.9M & 2.4M \\
  writingbench\_education & 1.7M & 1.6M & 1.8M & 2.1M \\
  writingbench\_finance\_business & 3.5M & 3.5M & 3.4M & 4.0M \\
  writingbench\_literature\_arts & 3.4M & 3.4M & 2.7M & 3.2M \\
  writingbench\_politics\_law & 2.7M & 2.6M & 2.6M & 3.3M \\
  \bottomrule
  \end{tabular}
}
\end{table}

\begin{table}[H]
  \centering
  \small
  \caption{Fusion (XHigh ($s{=}15$)) total completion tokens per benchmark.}
  \label{tab:fusion-high}
  \resizebox{\textwidth}{!}{
  \begin{tabular}{lrrrr}
  \toprule
  Benchmark & Qwen3.5-9B & Qwen3.5-35B-A3B & OLMo-3-7B & OLMo-3.1-32B \\
  \midrule
  healthbench\_communication & 37.1M & 33.5M & -- & -- \\
  healthbench\_complex\_responses & 16.2M & 15.2M & -- & -- \\
  healthbench\_context\_seeking & 18.1M & 16.7M & -- & -- \\
  healthbench\_emergency\_referrals & 14.9M & 13.6M & -- & -- \\
  healthbench\_global\_health & 38.7M & 35.2M & -- & -- \\
  healthbench\_health\_data\_tasks & 19.5M & 18.5M & -- & -- \\
  healthbench\_hedging & 36.7M & 33.5M & -- & -- \\
  lexam\_open & 29.2M & 26.1M & -- & -- \\
  prbench\_finance & 45.2M & 41.8M & -- & -- \\
  prbench\_finance\_hard & 23.0M & 21.3M & -- & -- \\
  prbench\_legal & 30.6M & 28.7M & -- & -- \\
  prbench\_legal\_hard & 16.0M & 15.0M & -- & -- \\
  WildBench & 83.2M & 75.0M & -- & -- \\
  writingbench\_academic\_engineering & 7.4M & 7.5M & -- & -- \\
  writingbench\_advertising\_marketing & 4.7M & 4.7M & -- & -- \\
  writingbench\_education & 4.1M & 4.0M & -- & -- \\
  writingbench\_finance\_business & 8.7M & 8.7M & -- & -- \\
  writingbench\_literature\_arts & 8.6M & 8.5M & -- & -- \\
  writingbench\_politics\_law & 6.8M & 6.6M & -- & -- \\
  \bottomrule
  \end{tabular}
}
\end{table}

\end{document}